\documentclass[twoside,11pt]{article}

\usepackage{amsmath}
\usepackage{amsfonts}
\usepackage{mathtools}
\usepackage{verbatim} 
\usepackage{multirow}

\usepackage{wrapfig}
\usepackage{booktabs}

\usepackage{pgfplots}
\usepackage{tikz} 

\usepackage{subfloat}
\usepackage{booktabs}
\usepackage{caption}
\usepackage{subcaption}
\usepackage{xcolor}

\pgfplotsset{compat=newest}
\usetikzlibrary{decorations.pathmorphing} 
\usetikzlibrary{fit} 
\usetikzlibrary{backgrounds} 
\usetikzlibrary{external}
\usetikzlibrary{pgfplots.groupplots}
\usetikzlibrary{positioning}
\usetikzlibrary{shapes.misc}
\usepackage{thmtools,thm-restate}

\usepackage{jmlr2e}

\tikzset{every axis/.append style = {
    major tick length=2.5,
    every tick/.style={
            black,
    }
    }
}

\tikzset{font={\fontsize{8pt}{9.6pt}\selectfont}}

\tikzset{cross/.style={cross out, draw=black, minimum size=2*(#1-\pgflinewidth), inner sep=0pt, outer sep=0pt, very thick}, cross/.default={1pt}}

\usepackage{algorithm}
\usepackage{algorithmic}

\DeclareMathOperator*{\argmax}{arg\,max}
\DeclareMathOperator*{\argmin}{arg\,min}
\newcommand{\norm}[1]{\left\lVert#1\right\rVert}
\newcommand{\cvec}[1]{\boldsymbol{#1}}
\newcommand{\svec}[1]{\mathbf{#1}}
\newcommand{\kldiv}[2]{D_{\text{KL}}\left(#1\ \middle\|\ #2 \right)}
\newcommand{\dif}{\mathop{}\!\mathrm{d}}

\usepackage{lastpage}

\ShortHeadings{Self-Paced Learning in RL}{Klink, Abdulsamad, Belousov, D'Eramo, Peters and Pajarinen}
\firstpageno{1}

\begin{document}

\title{A Probabilistic Interpretation of Self-Paced Learning with Applications to Reinforcement Learning}

\author{\name Pascal Klink \email pascal.klink@tu-darmstadt.de \\
       \addr Intelligent Autonomous Systems, TU Darmstadt, Germany
\AND
\name Hany Abdulsamad \email hany.abdulsamad@tu-darmstadt.de \\
       \addr Intelligent Autonomous Systems, TU Darmstadt, Germany
\AND
\name Boris Belousov \email boris.belousov@tu-darmstadt.de \\
       \addr Intelligent Autonomous Systems, TU Darmstadt, Germany 
\AND
\name Carlo D'Eramo\email carlo.deramo@tu-darmstadt.de \\
       \addr Intelligent Autonomous Systems, TU Darmstadt, Germany
\AND
\name Jan Peters \email jan.peters@tu-darmstadt.de \\
       \addr Intelligent Autonomous Systems, TU Darmstadt, Germany
\AND
\name Joni Pajarinen \email joni.pajarinen@aalto.fi \\
       \addr Intelligent Autonomous Systems, TU Darmstadt, Germany \\ Department of Electrical Engineering and Automation, Aalto University, Finland}

\editor{George Konidaris}

\maketitle

\begin{abstract}%
Across machine learning, the use of curricula has shown strong
empirical potential to improve learning from data by avoiding local optima of training objectives. For reinforcement learning (RL), curricula are especially interesting, as the underlying optimization has a strong tendency to get stuck in local optima due to the exploration-exploitation trade-off. Recently, a number of approaches for an automatic generation of curricula for RL have been shown to increase performance while requiring less expert knowledge compared to manually designed curricula. However, these
approaches are seldomly investigated from a theoretical perspective,
preventing a deeper understanding of their mechanics. In this paper, we present an approach for automated curriculum generation in RL with a clear theoretical underpinning. More precisely, we 
formalize the well-known self-paced learning paradigm as inducing a distribution over training tasks, which trades off between task complexity and the objective to match a desired
task distribution. Experiments show that training on this induced distribution helps to avoid poor local optima across RL algorithms in different tasks with uninformative rewards and challenging exploration requirements.
\end{abstract}


\section{Introduction}

Research on reinforcement learning (RL)~\citep{sutton1998introduction} has led to recent successes in long-horizon planning \citep{mnih2015human, silver2017mastering}
and robot control \citep{kober2009policy, levine2016end}. A driving factor of these successes has been the combination of RL paradigms with powerful function approximators, commonly referred to as deep RL (DRL). While DRL has considerably pushed the boundary w.r.t. the type and size of tasks that can be tackled, its algorithms suffer from high sample complexity. This can lead to poor performance in scenarios where the demand for samples is not satisfied. Furthermore, crucial challenges such as poor exploratory behavior of RL agents are still far from being solved, resulting in a large body of research that aims to reduce sample complexity by improving this exploratory behavior of RL agents \citep{machado2018count,tang2017exploration,bellemare2016unifying,houthooft2016vime,pmlr-v100-schultheis20a}.
\\
Another approach to making more efficient use of samples is to leverage similarities between learning environments and tasks in the framework of contextual- or multi-task RL. In these frameworks, a shared task structure permits simultaneous optimization of a policy for multiple tasks via inter- and extrapolation \citep{kupcsik2013data, schaul2015universal,jaderberg2016reinforcement}, resulting in tangible speed ups in learning across tasks. Such approaches expose the agent to tasks drawn from a distribution under which the agent should optimize its behavior. Training on such a fixed distribution, however, does not fully leverage the contextual RL setting in case there is a difference in difficulty among tasks. In such a scenario, first training on ``easier'' tasks and exploiting the generalizing behavior of the agent to gradually progress to ``harder'' ones promises to make more efficient use of environment interaction. This idea is at the heart of curriculum learning (CL), a term introduced by \cite{bengio2009curriculum} for supervised learning problems. By now, applications of CL have increasingly expanded to RL problems, where the aim is to design task sequences that maximally benefit the learning progress of an RL agent \citep{narvekar2020curriculum}.
\\
Recently, an increasing number of algorithms for an automated generation of curricula have been proposed \citep{baranes2010intrinsically,florensa2017reverse,andrychowicz2017hindsight,riedmiller2018learning}. While empirically demonstrating their beneficial effect on the learning performance of RL agents, the heuristics that guide the generation of the curriculum are, as of now, theoretically not well understood. In contrast, in supervised learning, self-paced learning \citep{kumar2010self} is an approach to curriculum generation that enjoys wide adaptation in practice \citep{supancic2013self,fan2018unsupervised,jiang2014easy} and has a firm theoretical interpretation as a majorize-minimize algorithm applied to a regularized objective \citep{meng2017theoretical}. In this paper, we develop an interpretation of self-paced learning as the process of generating a sequence of distributions over samples. We use this interpretation to transfer the concept of self-paced learning to RL problems, where the resulting approach generates a curriculum based on two quantities: the value function of the agent (reflecting the task complexity) and the KL divergence to a target distribution of tasks (reflecting the incorporation of desired tasks). \\
\textbf{Contribution} We propose an interpretation of the self-paced learning algorithm from a probabilistic perspective, in which the weighting of training samples corresponds to a sampling distribution (Section \ref{sec:prob-self-paced-learning}). Based on this interpretation, we apply self-paced learning to the contextual RL setting, obtaining a curriculum over RL tasks that trades-off agent performance and matching a target distribution of tasks (Section \ref{sec:sprl}). We connect the approach to the RL-as-inference paradigm \citep{toussaint2006probabilistic,levine2018reinforcement}, recovering well-known regularization techniques in the inference literature (Section \ref{sec:inference-view}). We experimentally evaluate algorithmic realizations of the curriculum in both episodic- (Section \ref{sec:episodic-rl}) and step-based RL settings (Section \ref{sec:step-based-rl}). Empirical evidence suggests that the scheme can match and surpass state-of-the-art CL methods for RL in environments of different complexity and with sparse and dense rewards.

\section{Related Work}
 
Simultaneously evolving the learning task with the learner has been investigated in a variety of fields ranging from behavioral psychology~\citep{skinner2019behavior} to evolutionary robotics~\citep{bongard2004once} and RL~\citep{asada1996purposive,erez2008does,wang2019poet}. For supervised learning, this principle was given the name \emph{curriculum learning} by \cite{bengio2009curriculum}. The name has by now also been established in the reinforcement learning (RL) community,  where a variety of algorithms aiming to generate curricula that maximally benefit the learner have been proposed.
\\
A driving principle behind curriculum reinforcement learning (CRL) is the idea of transferring successful behavior from one task to another, deeply connecting it to the problem of transfer learning \citep{pan2009survey,taylor2009transfer,lazaric2012transfer}. In general, transferring knowledge is\textemdash depending on the scenario\textemdash a challenging problem on its own, requiring a careful definition of what is to be transferred and what are the assumptions about the tasks between which to transfer. Aside from this problem, \citet{narvekar2019learning} showed that learning to create an \textit{optimal} curriculum can be computationally harder than learning the solution for a task from scratch. Both of these factors motivate research on tractable approximations to the problem of transfer and curriculum generation.
\\
To ease the problem of transferring behavior between RL tasks, a shared state-action space between tasks as well as an additional variable encoding the task to be solved are commonly assumed. This variable is usually called a goal \citep{schaul2015universal} or a context \citep{modi2018markov,kupcsik2013data}. In this paper, we will adapt the second name, also treating the word ``context'' and ``task'' interchangeably, i.e. treating the additional variable and the task that it represents as the same entity.
\\
It has been shown that function approximators can leverage the shared state-action space and the additional task information to generalize important quantities, such as value functions, across tasks \citep{schaul2015universal}. This approach circumvents the complicated problem of transfer in its generality, does however impose assumptions on the set of Markov decision processes (MDPs) as well as the contextual variable that describes them. Results from \cite{modi2018markov} suggest that one such assumption may be a gradual change in reward and dynamics of the MDP w.r.t. the context, although this requirement would need to be empirically verified. For the remainder of this document, we will disregard this important problem and focus on RL problems with similar characteristics as the ones investigated by \cite{modi2018markov}, as often done for other CRL algorithms. A detailed study of these assumptions and their impact on CRL algorithms is not known to us but is an interesting endeavor. We now continue to highlight some CRL algorithms and refer to the survey by \cite{narvekar2020curriculum} for an extensive overview.
\\
The majority of CRL methods can be divided into three categories w.r.t. the underlying concept. On the one hand, in tasks with binary rewards or success indicators, the idea of keeping the agent's success rate within a certain range has resulted in algorithms with drastically improved sample efficiency \citep{florensa2018automatic,florensa2017reverse,andrychowicz2017hindsight}. On the other hand, many CRL methods \citep{schmidhuber1991curious,baranes2010intrinsically,portelas2019teacher,fournier2018accuracy} are inspired by the idea of `curiosity' or `intrinsic motivation' \citep{oudeyer2007intrinsic,blank2005bringing}---terms that refer to the way humans organize autonomous learning even in the absence of a task to be accomplished. The third category includes algorithms that use the value function to guide the curriculum. While similar to methods based on success indicators in sparse reward settings, these methods can allow to incorporate the richer feedback available in dense rewards settings. To the best of our knowledge, only our work and that of \cite{wohlke2020performance} fall into this category. The work of \cite{wohlke2020performance} defines a curriculum over starting states using the gradient of the value function w.r.t. the starting state. The proposed curriculum prefers starting states with a large gradient norm of the value function, creating similarities to metrics used in intrinsic motivation. In our method, the value function is used as a competence measure to trade-off between easy tasks and tasks that are likely under a target distribution.
\\
Our approach to curriculum generation builds upon the idea of \emph{self-paced learning} (SPL), initially proposed by \cite{kumar2010self} for supervised learning tasks and extended by \cite{jiang2014self,jiang2015self} to allow for user-chosen penalty functions and constraints. SPL generates a curriculum by trading-off between exposing the learner to all available training samples and selecting samples in which the learner performs well. The approach has been employed in a variety of supervised learning problems \citep{supancic2013self,fan2018unsupervised,jiang2014easy}. Furthermore, \citet{meng2017theoretical} proposed a theoretical interpretation of SPL, identifying it as a majorize-minimize algorithm applied to a regularized objective function. Despite its well-understood theoretical standing and empirical success in supervised learning tasks, SPL has only been applied in a limited way to RL problems, restricting its use to the prioritization of replay data from an experience buffer in deep $Q$-networks \citep{ren2018self}. Orthogonal to this approach, we will make use of SPL to adaptively select training tasks during learning of the agent.
\\
Furthermore, we will connect the resulting algorithms to the RL-as-inference perspective during the course of this paper. Therefore, we wish to briefly point to several works employing this perspective~\citep{dayan1997using,toussaint2006probabilistic,deisenroth2013survey,rawlik2013stochastic,levine2018reinforcement}. 
Taking an inference perspective is beneficial when dealing with inverse problems or problems that require tractable approximations \citep{hennig2015probabilistic,prince2012computer}. Viewing RL as an inference problem naturally motivates regularization methods such as maximum- or relative entropy~\citep{ziebart2008maximum,peters2010relative,haarnoja2018soft} that have proven highly beneficial in practice. Further, this view allows to rigorously reason about the problem of optimal exploration in RL \citep{ghavamzadeh2015bayesian}. Finally, it stimulates the development of new, and interpretation of existing, algorithms as different approximations to the intractable integrals that need to be computed in probabilistic inference problems \citep{abdolmaleki2018maximum,fellows2019virel}. This results in a highly principled approach to tackling the challenging problem of RL. 

\section{Preliminaries}

This section introduces the necessary notation for both self-paced and reinforcement learning. Furthermore, the end of Section \ref{sec:prel-reinforcement-learning} details the intuition of curriculum learning for RL and in particular our approach.

\subsection{Self-Paced Learning}
\label{sec:prel-self-paced-learning}

The concept of self-paced learning (SPL), as introduced by \cite{kumar2010self} and extended by \cite{jiang2015self}, is defined for supervised learning settings, in which a function approximator $y = m(\svec{x}, \cvec{\omega})$ with parameters ${\cvec{\omega} \in \mathbb{R}^{d_{\cvec{\omega}}}}$ is trained w.r.t. a given data set $\mathcal{D} = \left\{ (\svec{x}_i, y_i)\ \vert\ \svec{x}_i \in \mathbb{R}^{d_{\svec{x}}},\ y_i \in \mathbb{R},\ i \in [1, N] \right\}$. In this setting, SPL generates a curriculum over the data set $\mathcal{D}$ by introducing a vector $\cvec{\nu} = [\nu_1\ \nu_2\ \ldots \ \nu_N] \in [0, 1]^N$ of weights $\nu_i$ for the entries $(\mathbf{x}_i, y_i)$ in the data set. These weights are automatically adjusted during learning via a `self-paced regularizer' $f(\alpha, \nu_i)$ in the SPL objective
\begin{align}
\cvec{\nu}^*, \cvec{\omega}^* = \argmin_{\cvec{\nu}, \cvec{\omega}}\ r(\cvec{\omega}) + \sum_{i=1}^N \left( \nu_i l(\svec{x}_i, y_i, \cvec{\omega}) + f(\alpha, \nu_i) \right), \qquad \alpha > 0. \label{eq:spl-objective}
\end{align}
The term $r(\cvec{\omega})$ represents potentially employed regularization of the model and $l(\svec{x}_i, y_i, \cvec{\omega})$ represents the error in the model prediction $\tilde{y}_i{=}m(\svec{x}_i, \cvec{\omega})$ for sample $(\svec{x}_i, y_i)$. The motivation for this principle as well as its name are best explained by investigating the solution $\cvec{\nu}^*(\alpha, \cvec{\omega})$ of optimization problem (\ref{eq:spl-objective}) when only optimizing it w.r.t. $\cvec{\nu}$ while keeping $\alpha$ and $\cvec{\omega}$ fixed. Introducing the notation $\nu^*(\alpha, l) = \argmin_{\nu} \nu l + f(\alpha, \nu)$,  we can define the optimal $\cvec{\nu}$ for given $\alpha$ and $\cvec{\omega}$ as
\begin{align*}
\cvec{\nu}^*(\alpha, \cvec{\omega}) = [\nu^*(\alpha, l(\svec{x}_1, y_1, \cvec{\omega}))\ \nu^*(\alpha, l(\svec{x}_2, y_2, \cvec{\omega}))\ \ldots\ \nu^*(\alpha, l(\svec{x}_N, y_N, \cvec{\omega}))].
\end{align*}
For the self-paced function $f_{\text{Bin}}(\alpha, \nu_i) {=} - \alpha \nu_i$ initially proposed by \cite{kumar2010self}, it holds that 
\begin{align}
\nu_{\text{Bin}}^*(\alpha, l) = \begin{cases}
1,\ \text{if}\ l < \alpha \\
0,\ \text{else.} 
\end{cases} \label{eq:nu-opt}
\end{align}
We see that the optimal weights $\cvec{\nu}_{\text{Bin}}^*(\alpha, \boldsymbol{\omega})$ focus on examples on which the model under the current parameters $\boldsymbol{\omega}$ performs better than a chosen threshold $\alpha$. By continuously increasing $\alpha$ and updating $\cvec{\nu}$ and $\cvec{\omega}$ in a block-coordinate manner, SPL creates a curriculum consisting of increasingly ``hard'' training examples w.r.t. the current model. A highly interesting connection between SPL and well-known regularization terms for machine learning has been established by \cite{meng2017theoretical}. Based on certain axioms on the self-paced regularizer $f(\alpha, \nu_i)$  (see appendix), \cite{meng2017theoretical} showed that the SPL scheme of alternatingly optimizing (\ref{eq:spl-objective}) w.r.t. $\cvec{\omega}$ and $\cvec{\nu}$ implicitly optimizes the regularized objective
\begin{align}
\min_{\cvec{\omega}} r(\cvec{\omega}) + \sum_{i=1}^N F_{\alpha}(l(\svec{x}_i, y_i, \cvec{\omega})), \qquad F_{\alpha}(l(\svec{x}_i, y_i, \cvec{\omega})) = \int_0^{l(\svec{x}_i, y_i, \cvec{\omega})} \nu^*(\alpha, \iota) \dif \iota. \label{eq:spl-int-form}
\end{align}
Using the Leibniz integral rule on $F_{\alpha}$, we can see that $\nabla_l F_{\alpha}(l) = \nu^*(\alpha, l)$. Put differently, the weight $\nu_i^*(\alpha, \cvec{\omega})$ encodes how much a decrease in the prediction error $l(\svec{x}_i, y_i, \cvec{\omega})$ for the training example $(\mathbf{x}_i, y_i)$ decreases the regularized objective (\ref{eq:spl-int-form}). In combination with the previously mentioned axioms on the self-paced regularizer $f(\alpha, \nu_i)$, this allowed \cite{meng2017theoretical} to prove the connection between (\ref{eq:spl-objective}) and (\ref{eq:spl-int-form}). Furthermore, they showed that, depending on the chosen self-paced regularizer, the resulting regularizer $F_{\alpha}(l(\svec{x}_i, y_i, \cvec{\omega}))$ corresponds exactly to non-convex regularization terms used in machine learning to e.g. guide feature selection \citep{zhang2008multi,zhang2010nearly}. Opposed to feature selection, SPL makes use of these regularizers to attenuate the influence of training examples which the model cannot explain under the current parameters $\cvec{\omega}$. This is done by reducing their contribution to the gradient w.r.t. $\cvec{\omega}$ via the function $F_{\alpha}$ (see Figure \ref{fig:reg-vis}). This naturally explains tendencies of SPL to improve learning e.g. in the presence of extreme noise, as empirically demonstrated by \cite{jiang2015self}.  \\
To summarize, we have seen that SPL formulates a curriculum over a set of training data as an alternating optimization of weights for the training data $\cvec{\nu}$ given the current model and the model parameters $\cvec{\omega}$ given the current weights. This alternating optimization performs an implicit regularization of the learning objective, suppressing the gradient contribution of samples that the model cannot explain under the current parameters. Empirically, this has been shown to reduce the likelihood of converging to poor local optima. In the subsequent sections, we apply SPL to the problem of reinforcement learning, for which we now introduce the necessary notation. 

\subsection{Reinforcement Learning}
\label{sec:prel-reinforcement-learning}

Reinforcement learning (RL) is defined as an optimization problem on a Markov decision process (MDP), a tuple $\mathcal{M} = \langle \mathcal{S}, \mathcal{A}, p, r, p_0 \rangle$ that defines an environment with states $\svec{s} \in \mathcal{S}$, actions $\svec{a} \in \mathcal{A}$, transition probabilities $p(\svec{s}' \vert \svec{s}, \svec{a})$, reward function $r: \mathcal{S} \times \mathcal{A} \mapsto \mathbb{R}$ and initial state distribution $p_0(\svec{s})$. Typically $\mathcal{S}$ and $\mathcal{A}$ are discrete spaces or subsets of $\mathbb{R}^n$. RL encompasses approaches that maximize a $\gamma$-discounted performance measure
\begin{align}
\max_{\cvec{\omega}} J(\cvec{\omega}) = \max_{\cvec{\omega}} \mathbb{E}_{p_0(\svec{s}_0), p(\svec{s}_{i+1} \vert \svec{s}_i, \svec{a}_i), \pi(\svec{a}_i \vert \svec{s}_i, \cvec{\omega})} \left[\sum_{i=0}^{\infty} \gamma^i r(\svec{s}_i, \svec{a}_i) \right] \label{eq:rl-objective}
\end{align}
by finding optimal parameters $\cvec{\omega}$ for the policy $\pi(\svec{a} \vert \svec{s}, \cvec{\omega})$ through interaction with the environment. A key ingredient to many RL algorithms is the value function
\begin{align}
V_{\cvec{\omega}}(\svec{s}) = \mathbb{E}_{\pi(\svec{a} \vert \svec{s}, \cvec{\omega})}\left[ r(\svec{s}, \svec{a}) + \gamma \mathbb{E}_{p(\svec{s}' \vert \svec{s}, \svec{a})} \left[ V_{\cvec{\omega}}(\svec{s}') \right] \right], \label{eq:value-function}
\end{align}
which encodes the long-term expected discounted reward of following policy $\pi(\cdot \vert \cdot, \cvec{\omega})$ from state $\svec{s}$. The value function (or an estimate of it) is related to the RL objective by ${J(\cvec{\omega}) = \mathbb{E}_{p_0(\svec{s})}\left[ V_{\cvec{\omega}}(\svec{s}) \right]}$. In order to exploit learning in multiple MDPs, we need to give the agent ways of generalizing behavior over them. A common approach to accomplish this is to assume a shared state-action space for the MDPs and parameterize the MDP by a contextual parameter $\svec{c} \in \mathcal{C} \subseteq \mathbb{R}^{m}$, i.e. ${\mathcal{M}(\svec{c}) = \langle \mathcal{S}, \mathcal{A}, p_{\svec{c}}, r_{\svec{c}}, p_{0, \svec{c}} \rangle}$. By conditioning the agent's behavior on this context, i.e. $\pi(\svec{a} \vert \svec{s}, \svec{c}, \cvec{\omega})$, and introducing a distribution $\mu(\svec{c})$ over the contextual parameter, we end up with a \emph{contextual} RL objective
\begin{align}
\max_{\cvec{\omega}} J(\cvec{\omega}, \mu) = \max_{\cvec{\omega}} \mathbb{E}_{\mu(\svec{c})} \left[ J(\cvec{\omega}, \svec{c}) \right] = \max_{\cvec{\omega}}  \mathbb{E}_{\mu(\svec{c}), p_{0, \svec{c}}(\svec{s})} \left[ V_{\cvec{\omega}}(\svec{s}, \svec{c}) \right] \label{eq:con-rl-objective}.
\end{align}
The value function $V_{\cvec{\omega}}(\svec{s}, \svec{c})$ now encodes the expected discounted reward of being in states $\svec{s}$ \emph{in context $\svec{c}$} and following the conditioned policy $\pi(\svec{a} \vert \svec{s}, \svec{c}, \cvec{\omega})$, i.e.
\begin{align}
V_{\cvec{\omega}}(\svec{s}, \svec{c}) = \mathbb{E}_{\pi(\svec{a} \vert \svec{s}, \svec{c}, \cvec{\omega})}\left[ r_{\svec{c}}(\svec{s}, \svec{a}) + \gamma \mathbb{E}_{p_{\svec{c}}(\svec{s}' \vert \svec{s}, \svec{a})} \left[ V_{\cvec{\omega}}(\svec{s}', \svec{c}) \right] \right]. \label{eq:con-value-fun}
\end{align} 
This formulation has been investigated by multiple works from different perspectives~\citep{neumann2011variational,schaul2015universal,modi2018markov}. Despite the generality of the RL paradigm and its power in formulating the problem of inferring optimal behavior from experience as a stochastic optimization, the practical realization of sophisticated RL algorithms poses many challenges in itself. For example, the extensive function approximations that need to be performed often result in the particular RL algorithm converging to a local optimum of the typically non-convex objectives (\ref{eq:rl-objective}) and (\ref{eq:con-rl-objective}), which may or may not encode behavior that is able to solve the task (or the distribution of tasks). In the single-task RL objective (\ref{eq:rl-objective}), the only way to avoid such problems is to improve approximations, e.g. by increasing the number of samples, or develop algorithms with strong exploratory behavior. In the contextual case (\ref{eq:con-rl-objective}), however, there is an appealing different approach. \\
Assume a task $\mathcal{M}(\svec{c})$, in which learning can robustly take place despite aforementioned approximations, e.g. since objective (\ref{eq:rl-objective}) is convex for $\mathcal{M}(\svec{c})$ and hence only one optimum exists. Furthermore, consider a second task $\mathcal{M}(\svec{c}')$ which now admits multiple optima with different expected rewards. If the solution to $\mathcal{M}(\svec{c})$ lies within the basin of attraction of the optimal solution to $\mathcal{M}(\svec{c}')$, first learning in $\svec{c}$ and afterward in $\svec{c}'$ promises to stabilize learning towards the optimal solution for $\svec{c}'$. This intuition is at the core of curriculum learning for RL. Looking at (\ref{eq:con-rl-objective}), a suitable formulation of a curriculum is as a sequence of context distributions $p_i(\svec{c})$. This sequence should converge to a desired target distribution $\mu(\svec{c})$, i.e. $\lim_{i \to \infty} p_i(\svec{c}) = \mu(\svec{c})$. \\
Before we show that self-paced learning induces such a sequence of distributions, we first want
to note an important property that is exploited by contextual- and curriculum reinforcement learning (CRL) algorithms especially in continuous domains: A
small distance $\| \svec{c} - \svec{c}' \|$ implies a certain
similarity between the tasks $\mathcal{M}(\svec{c})$ and
$\mathcal{M}(\svec{c}')$. Note that the imprecision of this
formulation is not by accident but is rather an acknowledgment that
the question of similarity between MDPs is a complicated topic on its
own. Nonetheless, if in a curriculum, a new training task $\svec{c}'$ is generated via
additive noise on a task $\svec{c}$ in which the agent demonstrates
good performance, the property is clearly exploited.
Furthermore, policy representations $\pi(\svec{a} \vert \svec{s}, \svec{c},
\cvec{\omega})$ such as e.g. (deep) neural networks also tend to encode continuity
w.r.t. $\svec{c}$. We wanted to highlight these observations, as they allow to judge
whether a given CRL algorithm, as well as ours, is applicable to a given problem.

\section{A Probabilistic Interpretation of Self-Paced Learning}
\label{sec:prob-self-paced-learning}

\begin{figure}
\centering
\includegraphics{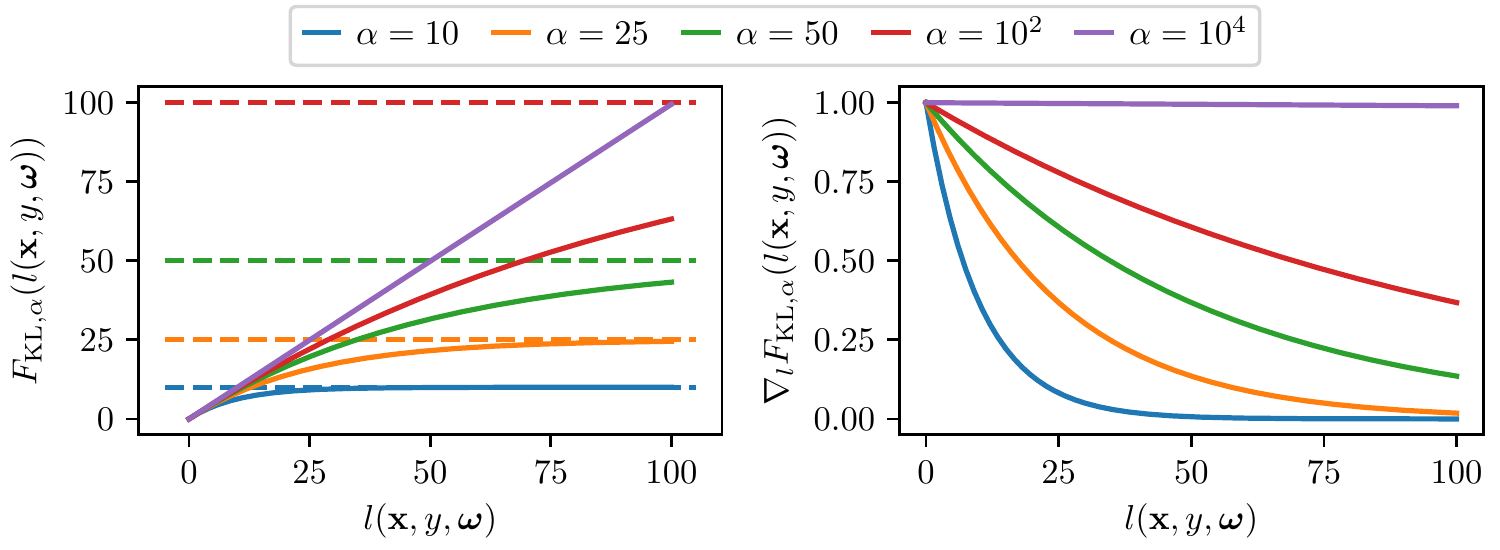}
\caption{A visualization of the effect of $F_{\text{KL}, \alpha}$ (see Equation \ref{eq:kl-non-conv-reg}) for different values of $\alpha$ and a single data-point $(\svec{x}, y)$. The left plot shows the transformation of the model error $l(\svec{x}, y, \cvec{\omega})$ by $F_{\text{KL}, \alpha}$. The right plot shows the gradient of $F_{\text{KL}, \alpha}$ w.r.t. $l(\svec{x}, y, \cvec{\omega})$, i.e. the corresponding weight $\nu^*_{\text{KL}}(\alpha, \cvec{\omega})$.}
\label{fig:reg-vis}
\end{figure}

At this point, we have discussed RL and highlighted the problem of policy optimization converging to a local optimum or only converging slowly. Ensuring to learn globally optimal policies with optimal sample complexity in its whole generality is an open problem. However, in Section \ref{sec:prel-self-paced-learning} we discussed that for supervised learning, the use of regularizing functions $F_{\alpha}$ that transform the loss can smooth out local optima created e.g. by noisy training data. Motivated by this insight, we now apply the aforementioned functions to regularize the contextual RL objective, obtaining
\begin{align}
\min_{\cvec{\omega}} \mathbb{E}_{\mu(\svec{c})} \left[ F_{\alpha}(-J(\cvec{\omega}, \svec{c})) \right]. \label{eq:self-paced-rl}
\end{align}
This objective has two slight differences to objective (\ref{eq:spl-int-form}). First, it misses the regularization term $r(\cvec{\omega})$  from (\ref{eq:spl-int-form}). Second, objective (\ref{eq:self-paced-rl}) is defined as an expectation of the regularized performance $F_{\alpha}(-J(\cvec{\omega}, \svec{c}))$ w.r.t. to the context distribution $\mu(\svec{c})$ instead of a sum over the regularized performances. This can be seen as a generalization of (\ref{eq:spl-int-form}), in which we allow to chose $\mu(\svec{c})$ differently from a uniform distribution over a discrete set of values. Regardless of these technical differences, one could readily optimize objective (\ref{eq:self-paced-rl}) in a supervised learning scenario e.g. via a form of stochastic gradient descent. As argued in Section \ref{sec:prel-self-paced-learning}, this results in an SPL optimization scheme (\ref{eq:spl-objective}) since the regularizer $F_{\alpha}$ performs an implicit weighting of the gradients $\nabla_{\cvec{\omega}} J(\cvec{\omega}, \svec{c})$. In an RL setting, the problem with such a straightforward optimization is that each evaluation of $J(\cvec{\omega}, \svec{c})$ and its gradient is typically expensive. If now for given parameters $\cvec{\omega}$ and context $\svec{c}$, the regularizer $F_{\alpha}$ leads to a negligible influence of $J(\cvec{\omega}, \svec{c})$ to the gradient of the objective (see Figure \ref{fig:reg-vis}), evaluating $J(\cvec{\omega}, \svec{c})$ wastes the precious resources that the learning agent should carefully utilize. In an RL setting, it is hence crucial to make use of a sampling distribution $p(\svec{c})$ that avoids the described wasteful evaluations. At this point, the insight that an SPL weight is equal to the gradient of the regularizing function $F_{\alpha}$ for the corresponding context, i.e. $\nu^*(\alpha, -J(\svec{c}, \cvec{\omega})) = \nabla_l F_{\alpha}(l) \vert_{l=-J(\svec{c}, \cvec{\omega})}$, directly yields a method for efficiently evaluating objective (\ref{eq:self-paced-rl})---that is by sampling a context $\svec{c}$ according to its SPL weight $\nu^*(\alpha, -J(\svec{c}, \cvec{\omega}))$. To make this intuition rigorous, we now introduce a probabilistic view on self-paced learning that views the weights $\cvec{\nu}$ in the SPL objective (\ref{eq:spl-objective}) as probabilities of a distribution over samples. More precisely, we define the categorical probability distribution $p(c{=}i \vert \cvec{\nu}) = \nu_i$ for $i \in [1, N]$. Note that we restrict ourselves to discrete distributions $p(c{=}i \vert \cvec{\nu})$ in this section to both ease the exposition and more easily establish connections to the SPL objective introduced in Section \ref{sec:prel-self-paced-learning}, although the results can be generalized to continuous distributions $\mu(\svec{c})$. For $p(c{=}i\vert \cvec{\nu}) = \nu_i$ to be a valid probability distribution, we only need to introduce the constraint $\sum_{i=1}^N \nu_i = 1$, as $\nu_i \geq 0$ per definition of SPL. Hence, we rewrite the SPL objective (\ref{eq:spl-objective}) as
\begin{align}
\cvec{\nu}^*, \cvec{\omega}^* = \argmin_{\cvec{\nu}, \cvec{\omega}}\ &r(\cvec{\omega}) + \mathbb{E}_{p(c \vert \cvec{\nu})} \left[ l(\svec{x}_c, y_c, \cvec{\omega}) \right] + \sum_{i=1}^N f(\alpha, p(c {=} i \vert \cvec{\nu})), \qquad \alpha > 0 \nonumber \\
\text{s.t.}\ &\sum_{i=1}^N \nu_i = 1. \label{eq:pspl-objective}
\end{align}
Apart from changes in notation, the only difference w.r.t. (\ref{eq:spl-objective}) is the constraint that forces the variables $\nu_i$ to sum to $1$. Interestingly, depending on the self-paced regularizer $f(\alpha, p(c {=} i \vert \cvec{\nu}))$, this constraint does not just lead to a normalization of the SPL weights obtained by optimizing objective (\ref{eq:spl-objective}). This is because the previously independent SPL weights $\nu^*(\alpha, l(\svec{x}_i, y_i, \cvec{\omega}))$ are now coupled via the introduced normalization constraint. An example of such a ``problematic'' regularizer is the seminal one $f_{\text{Bin}}(\alpha, \nu_i) = - \alpha \nu_i$ explored by \cite{kumar2010self}. With the additional constraint, the optimal solution $\cvec{\nu}^*_{\text{Bin}}$ to (\ref{eq:pspl-objective}) simply puts all weight on the sample with the minimum loss instead of sampling uniformly among samples with a loss smaller than $\alpha$. Although there seems to be no general connection between objective (\ref{eq:spl-objective}) and (\ref{eq:pspl-objective}) that holds for arbitrary self-paced regularizers, we can show that for the self-paced regularizer
\begin{align}
f_{\text{KL},i}(\alpha, \nu_i) = \alpha \nu_i \left( \log(\nu_i) - \log(\mu(c{=}i)) \right) - \alpha \nu_i, \label{eq:pspl-kl-regularizer}
\end{align}
the value of $\nu^*_{\text{KL}, i}(\alpha, \cvec{\omega})$ obtained by optimizing (\ref{eq:spl-objective}) and (\ref{eq:pspl-objective}) w.r.t. $\cvec{\nu}$ is identical up to a normalization constant. The user-chosen distribution $\mu(c)$ in (\ref{eq:pspl-kl-regularizer}) represents the likelihood of $(\svec{x}_c, y_c)$ occurring and has the same interpretation as in objective (\ref{eq:self-paced-rl}). The corresponding function $F_{\text{KL},\alpha,i}$ is given by
\begin{align}
F_{\text{KL}, \alpha, i}(l(\svec{x}_i, y_i, \cvec{\omega})) {=} \int_0^{l(\svec{x}_i, y_i, \cvec{\omega})} \nu_{\text{KL},i}^*(\alpha, \iota)  \dif \iota = \mu(c {=} i) \alpha \left( 1 {-} \exp \left( -\frac{1}{a} l(\svec{x}_i, y_i, \cvec{\omega}) \right) \right) \label{eq:kl-non-conv-reg}
\end{align}
and is visualized in Figure \ref{fig:reg-vis}.  Note the additional subscript $i$ in both $f_{\text{KL},i}$ and $F_{\text{KL}, \alpha, i}$. This extra subscript arises due to the  appearance of the likelihood term $\mu(c{=}i)$ in both formulas, resulting in an individual regularizer for each sample $(\svec{x}_i, y_i)$. As can be seen, $F_{\text{KL}, \alpha, i}(l)$ exhibits a squashing effect to limit the attained loss $l$ to a maximum value of $\alpha$. The closer the non-regularized loss $l$ attains this maximum value of $\alpha$, the more it is treated as a constant value by $F_{\text{KL}, \alpha, i}(l)$. For $l$ increasingly smaller than $\alpha$, a change in the non-regularized loss $l$ leads to an increasingly linear change in the regularized loss $F_{\text{KL}, \alpha, i}(l)$. More interestingly, using $f_{\text{KL},i}(\alpha, \nu_i)$ in objective (\ref{eq:pspl-objective}) results in a KL-Divergence penalty to $\mu(c)$. Theorem \ref{theo:1} summarizes these findings. The proof can be found in the appendix.\begin{restatable}{theo}{retheo}
Alternatingly solving
\begin{align*}
\min_{\cvec{\omega}, \cvec{\nu}} \mathbb{E}_{p(c \vert \cvec{\nu})} \left[  l(\svec{x}_c, y_c, \cvec{\omega}) \right] + \alpha \kldiv{p(c \vert \cvec{\nu})}{\mu(c)}
\end{align*}
w.r.t. $\cvec{\omega}$ and $\cvec{\nu}$ is a majorize-minimize scheme applied to the regularized objective
\begin{align*}
\min_{\cvec{\omega}} \mathbb{E}_{\mu(c)} \left[ \alpha \left( 1 - \exp \left( -\frac{1}{\alpha} l(\svec{x}_c, y_c, \cvec{\omega}) \right) \right) \right].
\end{align*}
\label{theo:1}
\end{restatable}
\noindent In the following section, we make use of the insights summarized in Theorem \ref{theo:1} to motivate a curriculum as an effective evaluation of the regularized RL objective (\ref{eq:self-paced-rl}) under the particular choice $F_{\alpha} = F_{\text{KL},\alpha,i}$.

\section{Self-Paced Learning for Reinforcement Learning}
\label{sec:sprl}

Obtaining an efficient way of optimizing objective (\ref{eq:self-paced-rl}) with $F_{\alpha} = F_{\text{KL},\alpha,i}$ is as easy as exploiting Theorem \ref{theo:1} to define the alternative objective 
\begin{align*}
\max_{\cvec{\omega}, \cvec{\nu}} \mathbb{E}_{p(\svec{c} \vert \cvec{\nu})} \left[ J(\cvec{\omega}, \svec{c}) \right] - \alpha \kldiv{p(\svec{c} \vert \cvec{\nu})}{\mu(\svec{c})}.
\end{align*}
As discussed in the previous section, this formulation introduces a way of computing the desired sampling distribution that efficiently evaluates objective (\ref{eq:self-paced-rl}) given the current agent parameters $\cvec{\omega}$ by optimizing the above optimization problem w.r.t. $\cvec{\nu}$. As discussed in Section \ref{sec:prel-self-paced-learning}, $p(\svec{c} \vert \cvec{\nu})$ will assign probability mass to a context $\svec{c}$ based on its contribution to the gradient of objective (\ref{eq:self-paced-rl}). Before we look at the application to RL problems, we will introduce a regularization that is an important ingredient to achieve practicality. More precisely, we introduce a KL divergence constraint between subsequent context distributions $p(\svec{c} \vert \cvec{\nu})$ and $p(\svec{c} \vert \cvec{\nu}')$, yielding
\begin{align}
\max_{\cvec{\omega}, \cvec{\nu}}\ &\mathbb{E}_{p(\svec{c} \vert \nu)} \left[ J(\cvec{\omega}, \svec{c}) \right] - \alpha \kldiv{p(\svec{c} \vert \cvec{\nu})}{\mu(\svec{c})} \nonumber \\
\text{s.t.}\ &\kldiv{p(\svec{c} \vert \cvec{\nu})}{p(\svec{c} \vert \cvec{\nu}')} \leq \epsilon, \label{eq:kl-constrained-sprl}
\end{align}
with $\cvec{\nu}'$ being the parameters of the previously computed context distribution. In a practical algorithm, this secondary regularization is important because the expected performance $J(\cvec{\omega}, \svec{c})$ is approximated by a learned value function, which may not predict accurate values for contexts not likely under $p(\svec{c} \vert \cvec{\nu}')$. The KL divergence constraint helps to avoid exploiting these false estimates too greedily. Furthermore, it forces the distribution over contextual variables, and hence tasks, to gradually change, which we argued to be beneficial for RL algorithms at the end of Section \ref{sec:prel-reinforcement-learning}. From a theoretical perspective on SPL, the constraint changes the form of $\cvec{\nu}^*$ making it not only dependent on $\alpha$ and $\cvec{\omega}$, but also on the previous parameter $\cvec{\nu}'$. Although it may be possible to relate this modification to a novel regularizer $F_{\alpha,i}$, we do not pursue this idea here but rather connect (\ref{eq:kl-constrained-sprl}) to the RL-as-inference perspective in Section \ref{sec:inference-view}, where we can show highly interesting similarities to the well-known concept of tempering in inference. To facilitate the intuition of the proposed curriculum and its usage, we, however, first present applications and evaluations in the following sections. \\
An important design decision for such applications is the schedule for $\alpha$, i.e. the parameter of the regularizing function $F_{\alpha}$. As can be seen in (\ref{eq:kl-constrained-sprl}), $\alpha$ corresponds to the trade-off between reward maximization and progression to $\mu(\svec{c})$. In a supervised learning scenario, it is preferable to increase $\alpha$ as slowly as possible to gradually transform the objective from an easy version towards the target one. In an RL setting, each algorithm iteration requires the collection of data from the (real) system. Since the required amount of system interaction should be minimized, we cannot simply choose very small step sizes for $\alpha$, as this would lead to a slower than necessary progression towards $\mu(\svec{c})$. In the implementations in sections \ref{sec:episodic-rl} and \ref{sec:step-based-rl}, the parameter $\alpha$ is chosen such that the KL divergence penalty w.r.t. the current context distribution $p(\svec{c} \vert \cvec{\nu}_k)$ is in constant proportion $\zeta$ to the expected reward under this current context distribution and current policy parameters $\cvec{\omega}_k$
\begin{align}
\alpha_k = \mathcal{B}(\cvec{\nu}_k, \cvec{\omega}_k) = \zeta \frac{\mathbb{E}_{p(\svec{c} \vert \cvec{\nu}_k)} \left[ J(\cvec{\omega}_k, \svec{c}) \right]}{\kldiv{p(\svec{c} \vert \cvec{\nu}_k)}{\mu(\svec{c})}}. \label{eq:alpha-computation}
\end{align}
For the first $K_{\alpha}$ iterations, we set $\alpha$ to zero, i.e. only focus on maximizing the reward under $p(\svec{c} \vert \cvec{\nu})$. In combination with an initial context distribution $p(\svec{c} \vert \cvec{\nu}_0)$ covering large parts of the context space, this allows to tailor the context distribution to the learner in the first iterations by focusing on tasks in which it performs best under the initial parameters. Note that this schedule is a naive choice, that nonetheless worked sufficiently well in our experiments. In Section \ref{sec:optimal-alpha}, we revisit this design choice and investigate it more carefully.

\section{Application to Episodic Reinforcement Learning}
\label{sec:episodic-rl}

In this section, we implement and evaluate our formulation of SPL for RL in a slightly different way than the ``full'' RL setting, which has been described in Section \ref{sec:prel-reinforcement-learning} and will be evaluated in the next section. Instead, we frame RL as a black-box optimization problem \citep{conn2009introduction,hansen2010comparing}. This setting is interesting for two reasons: Firstly, it has been and still is a core approach to perform RL on real (robotic) systems \citep{kupcsik2013data,parisi2015reinforcement,ploeger2020high}, where ``low-level'' policies such as Dynamic- and Probabilistic Movement Primitives \citep{schaal2006dynamic,paraschos-promps} or PD-control laws \citep{berkenkamp2016safe} are commonly used to ensure smooth and stable trajectories while keeping the dimensionality of the search space reasonably small. Secondly, the different mechanics of the employed episodic RL algorithm and the resulting different implementation of objective (\ref{eq:kl-constrained-sprl}) serve as another validation of our SPL approach to CRL apart from the deep RL experiments in the next section.
Readers not interested in or familiar with the topic of black-box optimization (and episodic RL) can skip this section and continue to the experiments with deep RL algorithms in Section \ref{sec:step-based-rl}. \\
The episodic RL setting arises if we introduce an additional ``low-level'' policy $\pi(\svec{a} \vert \svec{s}, \cvec{\theta})$ with parameters $\cvec{\theta} \in \mathbb{R}^{d_{\cvec{\theta}}}$ and change the policy introduced in Section \ref{sec:prel-reinforcement-learning} to not generate actions given the current state and context, but only generate a parameter $\cvec{\theta}$ for the low-level policy given the current context, i.e. $\pi(\cvec{\theta} \vert \svec{c}, \cvec{\omega})$. Defining the expected reward for a parameter $\cvec{\theta}$ in context~$\svec{c}$ as
\begin{align}
r(\cvec{\theta}, \svec{c}) = \mathbb{E}_{p_{0,\svec{c}}(\svec{s})} \left[ V_{\pi(\svec{a} \vert \svec{s}, \cvec{\theta})}(\svec{s}, \svec{c}) \right], \label{eq:ep-rl-objective}
\end{align}
we see that we can simply interpret $r(\cvec{\theta}, \svec{c})$ as a function that, due to its complicated nature, does only allow for noisy observations of its function value without any gradient information. The noise in function observations arises from the fact that a rollout of policy $\pi(\svec{a} \vert \svec{s}, \cvec{\theta})$ in a context $\svec{c}$ corresponds to approximating the expectations in $r(\cvec{\theta}, \svec{c})$ with a single sample. 
\\
As a black-box optimizer for the experiments, we choose the contextual relative entropy policy search (C-REPS) algorithm \citep{neumann2011variational,kupcsik2013data,parisi2015reinforcement}, which frames the maximization of (\ref{eq:ep-rl-objective}) over a task distribution $\mu(\svec{c})$ as a repeated entropy-regularized optimization
\begin{align*}
\max_{q(\cvec{\theta}, \svec{c})}\ &\mathbb{E}_{q(\cvec{\theta}, \svec{c})} \left[ r( \cvec{\theta}, \svec{c}) \right]\quad \text{s.t.}\ \kldiv{q(\cvec{\theta}, \svec{c})}{p(\cvec{\theta}, \svec{c})} \leq \epsilon \quad  \int q(\cvec{\theta}, \svec{c}) \dif \cvec{\theta} = \mu(\svec{c})\ \forall \svec{c} \in \mathcal{C},
\end{align*}
where $p(\cvec{\theta}, \svec{c}) = p(\cvec{\theta} \vert \svec{c}) \mu(\svec{c})$ is the distribution obtained in the previous iteration. Note that the constraint in the above optimization problem implies that only the policy $q(\cvec{\theta} \vert \svec{c})$ is optimized since the constraint requires that $q(\cvec{\theta}, \svec{c}) = q(\cvec{\theta} \vert \svec{c}) \mu(\svec{c})$. This notation is common for this algorithm as it eases the derivations of the solution via the concept of Lagrangian multipliers. Furthermore, this particular form of the C-REPS algorithm allows for a straightforward incorporation of SPL, simply replacing the constraint $\int q(\cvec{\theta}, \svec{c}) d\cvec{\theta} = \mu(\svec{c})$ by a penalty term on the KL divergence between $q(\svec{c}) = \int q(\cvec{\theta}, \svec{c}) d\cvec{\theta}$ and $\mu(\svec{c})$
\begin{align}
\max_{q(\cvec{\theta}, \svec{c})}\ &\mathbb{E}_{q(\cvec{\theta}, \svec{c})} \left[ r( \cvec{\theta}, \svec{c}) \right] - \alpha \kldiv{q(\svec{c})}{\mu(\svec{c})} \nonumber \\
\text{s.t.}\ &\kldiv{q(\cvec{\theta}, \svec{c})}{p(\cvec{\theta}, \svec{c})} \leq \epsilon. \label{eq:sprl-episodic}
\end{align}
The above objective does not yet include the parameters $\cvec{\omega}$ or $\cvec{\nu}$ of the policy or the context distribution to be optimized. This is because both C-REPS and also our implementation of SPL for episodic RL solve the above optimization problem analytically to obtain a re-weighting scheme for samples $(\cvec{\theta}_i, \svec{c}_i) \sim p(\cvec{\theta} \vert \svec{c}, \cvec{\omega}_k) p(\svec{c} \vert \cvec{\nu}_k)$ based on the observed rewards $r(\cvec{\theta}_i, \svec{c}_i)$. The next parameters $\cvec{\omega}_{k+1}$ and $\cvec{\nu}_{k+1}$ are then found by a maximum-likelihood fit to the set of weighted samples. The following section will detail some of the practical considerations necessary for this.

\begin{algorithm}[t]
	\caption{Self-Paced Episodic Reinforcement Learning (SPRL)}
	\label{alg:episodic-sprl}
	\begin{algorithmic}
		\STATE{\bfseries Input:}  Initial context distribution- and policy parameters $\cvec{\nu}_0$ and $\cvec{\omega}_0$, Target context distribution $\mu(\svec{c})$, KL penalty proportion $\zeta$, Offset $K_{\alpha}$, Number of iterations $K$, Rollouts per policy update $M$, Relative entropy bound $\epsilon$
		
		\FOR{$k=1$ {\bfseries to} $K$}
		\STATE {\bfseries Collect Data:}
		\STATE Sample contexts: $\svec{c}_{i} \sim p(\svec{c} \vert \cvec{\nu}_{k-1}),\ i \in [1, M]$
		\STATE Sample parameters: $\cvec{\theta}_{i} \sim p(\cvec{\theta} \vert \svec{c}_{i}, \cvec{\omega}_{k-1})$
		\STATE Execute $\pi(\cdot \vert \svec{s}, \cvec{\theta}_i)$ in $\svec{c}_i$ and observe reward: $r_i = r\left(\cvec{\theta}_i, \svec{c}_i\right)$
		\STATE Create sample set: $\mathcal{D}_k = \{(\cvec{\theta}_i, \svec{c}_i, r_i) \vert i \in [1, M] \}$
		
		\STATE {\bfseries Update Policy and Context Distributions:}
		
		\STATE Update schedule: $\alpha_k = 0,\ \text{if } k \leq K_{\alpha},\ \text{else } \mathcal{B}(\cvec{\nu}_{k-1}, \cvec{\omega}_{k - 1})$ (\ref{eq:alpha-computation})
		\STATE Optimize dual function:  $\left[\eta_q^*, \eta_{\tilde{q}}^*, V^*\right] \leftarrow \argmin \mathcal{G}(\eta_{q}, \eta_{\tilde{q}}, V)$ (\ref{eq:sprl_dual})
		
		\STATE Calculate sample weights: $\left[w_i, \tilde{w}_i\right] \leftarrow \left[ \exp\left(\frac{A(\cvec{\theta}_i, \svec{c}_i)}{\eta_{q}^*}\right), \exp\left(\frac{\beta(\svec{c}_i)}{\alpha_k + \eta_{\tilde{q}}^*}\right)\right]$ (\ref{eq:episodic-pol-update}), (\ref{eq:episodic-con-update})
		
		\STATE Infer new parameters: $\left[\cvec{\omega}_k, \cvec{\nu}_k\right] \leftarrow \left\{(w_i, \tilde{w}_i, \cvec{\theta}_{i}, \svec{c}_i) \vert i \in \left[1, M\right] \right\}$
		\ENDFOR
	\end{algorithmic}
\end{algorithm}

\subsection{Algorithmic Implementation}

Solving (\ref{eq:sprl-episodic}) analytically using the technique of Lagrangian multipliers, we obtain the following form for the variational distributions
\begin{align}
q(\cvec{\theta}, \svec{c}) &{\propto} p(\cvec{\theta}, \svec{c} \vert \cvec{\omega}_k, \cvec{\nu}_k) \exp \left( \frac{r(\cvec{\theta}, \svec{c}) - V(\svec{c})}{\eta_q} \right) = p(\cvec{\theta}, \svec{c} \vert \cvec{\omega}_k, \cvec{\nu}_k) \exp \left( \frac{A(\cvec{\theta}, \svec{c})}{\eta_q} \right), \label{eq:episodic-pol-update} \\
q(\svec{c}) &{\propto} p(\svec{c} \vert \cvec{\nu}_k) \exp \left( \frac{V(\svec{c}) + \alpha (\log(\mu(\svec{c})) - \log(p(\svec{c} \vert \cvec{\nu}_k)))}{\alpha + \eta_{\tilde{q}}} \right) = p(\svec{c} \vert \cvec{\nu}_k) \exp \left( \frac{\beta(\svec{c})}{\alpha + \eta_{\tilde{q}}} \right), \label{eq:episodic-con-update}
\end{align}
with $\eta_q$, $\eta_{\tilde{q}}$ as well as $V(\svec{c})$ being Lagrangian multipliers that are found by solving the dual objective
\begin{align}
\mathcal{G} & = (\eta_{q} + \eta_{\tilde{q}})\epsilon + \eta_{q}\log\left(\mathbb{E}_{p}\left[\exp\left(\frac{A(\cvec{\theta}, \svec{c})}{\eta_{q}}\right)\right]\right) + (\alpha+\eta_{\tilde{q}})\log\left(\mathbb{E}_{p}\left[\exp\left(\frac{\beta(\svec{c})}{\alpha+\eta_{\tilde{q}}}\right)\right]\right) \label{eq:sprl_dual}.
\end{align}
The derivation of the dual objective, as well as the solution to objective (\ref{eq:sprl-episodic}), are shown in the appendix. As previously mentioned, in practice the algorithm has only access to a set of samples $\mathcal{D} = \left\{(\cvec{\theta}_i, \svec{c}_i, r_i) \vert i \in [1, M] \right\}$ and hence the analytic solutions (\ref{eq:episodic-pol-update}) and (\ref{eq:episodic-con-update}) are approximated by re-weighting the samples via weights $w_i$. To compute the optimal weights $w_i$, the multipliers $V^*$, $\eta_q^*$ and $\eta_{\tilde{q}}^*$ need to be obtained by minimizing the dual (\ref{eq:sprl_dual}), to which two approximations are introduced: First, the expectations w.r.t. $p(\cvec{\theta}, \svec{c} \vert \cvec{\omega}, \cvec{\nu})$ (abbreviated as $p$ in Equation \ref{eq:sprl_dual}) are replaced by a sample-estimate from the collected samples in $\mathcal{D}$. Second, we introduce a parametric form for the value function $V(\svec{c}) = \cvec{\chi}^T \phi(\svec{c})$ with a user-chosen feature function $\phi(\svec{c})$, such that we can optimize (\ref{eq:sprl_dual}) w.r.t. $\cvec{\chi}$ instead of $V$.
\\
After finding the minimizers $\cvec{\chi}^*$, $\eta_q^*$ and $\eta_{\tilde{q}}^*$ of (\ref{eq:sprl_dual}), the weights $w_i$ are then given by the exponential terms in (\ref{eq:episodic-pol-update}) and (\ref{eq:episodic-con-update}). The new policy- and context distribution parameters are fitted via maximum likelihood to the set of weighted samples. In our implementation, we use Gaussian context distributions and policies. To account for the error that originates from the sample-based approximation of the expectations in (\ref{eq:sprl_dual}), we enforce the KL divergence constraint $\kldiv{p(\cvec{\theta}, \svec{c} \vert \cvec{\omega}_{k}, \cvec{\nu}_{k})}{q(\cvec{\theta}, \svec{c} \vert \cvec{\omega}_{k+1}, \cvec{\nu}_{k+1})} \leq \epsilon$ when updating the policy and context distribution. Again, details on this maximum likelihood step can be found in the appendix. To compute the schedule for $\alpha$ according to (\ref{eq:alpha-computation}), we approximate the expected reward under the current policy with the mean of the observed rewards, i.e. $\mathbb{E}_{p(\svec{c} \vert \cvec{\nu}_k)} \left[ J(\cvec{\omega}_k, \svec{c}) \right] \approx \frac{1}{M} \sum_{i=1}^M r_i$. The overall procedure is summarized in Algorithm~\ref{alg:episodic-sprl}. 

\subsection{Experiments}

We now evaluate the benefit of the SPL paradigm in the episodic RL scenario (SPRL). Besides facilitating learning on a diverse set of tasks, we are also interested in the idea of facilitating the learning of a hard target task via a curriculum. This modulation can be achieved by choosing $\mu(\svec{c})$ to be a narrow probability distribution focusing nearly all probability density on the particular target task. To judge the benefit of our SPL adaptation for these endeavors, we compared our implementation to C-REPS, \mbox{CMA-ES}~\citep{hansen-cmaes}, GoalGAN~\citep{florensa2018automatic} and SAGG-RIAC~\citep{baranes2010intrinsically}. With CMA-ES being a non-contextual algorithm, we only use it in experiments with narrow target distributions, where we then train and evaluate only on the mean of the target context distributions. We will start with a simple point-mass problem, where we evaluate the benefit of our algorithm for broad and narrow target distributions. We then turn towards more challenging tasks, such as a modified version of the reaching task implemented in the OpenAI Gym simulation environment \citep{brockman2016openai} and a sparse ball-in-a-cup task. Given that GoalGAN and SAGG-RIAC are algorithm agnostic curriculum generation approaches, we combine them with C-REPS to make the results as comparable as possible.
\\
In all experiments, we use radial basis function (RBF) features to approximate the value function $V(\svec{c})$, while the policy $p(\cvec{\theta} \vert \svec{c}, \cvec{\omega}) = \mathcal{N}(\cvec{\theta} \vert \cvec{A}_{\cvec{\omega}} \phi(\svec{c}), \cvec{\Sigma}_{\cvec{\omega}})$ uses linear features $\phi(\svec{c})$. SPRL and C-REPS always use the same number of RBF features for a given environment. SPRL always starts with a wide initial sampling distribution $p\left(\svec{c} \vert \cvec{\nu}_0\right)$ that, in combination with setting $\alpha = 0$ for the first $K_{\alpha}$ iterations, allows the algorithm to automatically choose the initial tasks on which learning should take place. After the first $K_{\alpha}$ iterations, we then choose $\alpha$ following the scheme outlined in the previous section. Experimental details that are not mentioned here to keep the section short can be found in the appendix. \footnote{Code is publicly available under \url{https://github.com/psclklnk/self-paced-rl}.}

\subsubsection{Point-Mass Environment}
\label{sec:ep-rl-gate-exp}

\begin{figure}[!b]
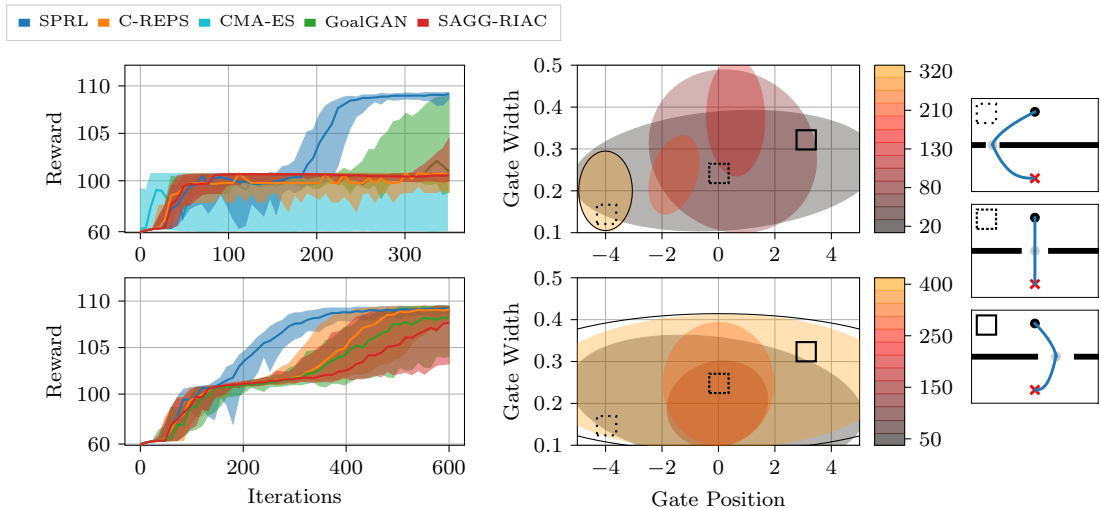

	\centering
	\begin{tikzpicture}
		\node (O0)  {\input{./img/gate-precision+gate-global-results.tex}};
		\node [above right = -80pt and -5pt of O0] (O1) {\scalebox{0.11}{\input{./img/gate-pos-1.pgf}}};
		\node [below = -5.8pt of O1] (O2) {\scalebox{0.11}{\input{./img/gate-pos-3.pgf}}};
		\node [below = -5.8pt of O2] (O3) {\scalebox{0.11}{\input{./img/gate-pos-2.pgf}}};

		\node[draw,thick,dotted,minimum width=0.2cm,minimum height=0.2cm, above left = -15pt and -15pt of O1] (C1) {};
		\node[draw,thick,dotted,minimum width=0.2cm,minimum height=0.2cm, below left = -41pt and -235pt of O0] (C1) {};
		\node[draw,thick,dotted,minimum width=0.2cm,minimum height=0.2cm, below left = -121pt and -235pt of O0] (C1) {};

		\node[draw,thick,dash pattern={on 1pt off 0.75pt},minimum width=0.2cm,minimum height=0.2cm, above left = -15pt and -15pt of O2] (C1) {};
		\node[draw,thick,dash pattern={on 1pt off 0.75pt},minimum width=0.2cm,minimum height=0.2cm, below left = -57pt and -277.5pt of O0] (C1) {};
		\node[draw,thick,dash pattern={on 1pt off 0.75pt},minimum width=0.2cm,minimum height=0.2cm, below left = -136.5pt and -277.5pt of O0] (C1) {};

		\node[draw,thick,minimum width=0.2cm,minimum height=0.2cm, above left = -15pt and -15pt of O3] (C1) {};
		\node[draw,thick,minimum width=0.2cm,minimum height=0.2cm, below left = -69pt and -310.5pt of O0] (C1) {};
		\node[draw,thick,minimum width=0.2cm,minimum height=0.2cm, below left = -149.2pt and -310.5pt of O0] (C1) {};
	\end{tikzpicture}
	\vspace{-5pt}
	\caption[Gate Experiments]{Left: Reward in the ``precision'' (top row) and ``global'' setting (bottom row) on the target context distributions in the gate environment. Thick lines represent the $50\%$-quantiles and shaded areas the intervals from $10\%$- to $90\%$-quantile of $40$ algorithm executions. Middle: Evolution of the sampling distribution $p\left(\svec{c} \vert \cvec{\nu} \right)$ (colored areas) of one SPRL run together with the target distribution $\mu\left(\svec{c}\right)$ (black line). Right: Task visualizations for different gate positions and widths. The boxes mark the corresponding positions in the context space.}
	\label{fig:gate-experiments}
	\vspace{-10pt}
\end{figure}

\begin{figure}[t]
	\centering
	\input{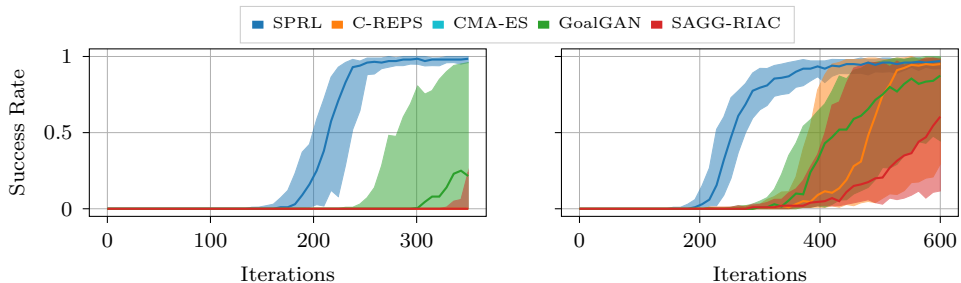}
	\vspace{-5pt}
	\caption[Gate Rewards]{Success rates in the ``precision'' (left) and ``global'' setting (right) of the gate environment. Thick lines represent the $50\%$-quantiles and shaded areas show the intervals from $10\%$- to $90\%$-quantile. Quantiles are computed using $40$ algorithm executions.}
	\label{fig:gate-successes}
	\vspace{-7pt}
\end{figure}

In the first environment, the agent needs to steer a point-mass in a two-dimensional space from the starting position $\left[0\ \ 5\right]$ to the goal position at the origin. The dynamics of the point-mass are described by a simple linear system subject to a small amount of Gaussian noise. Complexity is introduced by a wall at height $y = 2.5$, which can only be traversed through a gate. The $x$-position and width of the gate together define a task $\svec{c}$. If the point-mass crashes into the wall, the experiment is stopped and the reward is computed based on the current position. The reward function is the exponentiated negative distance to the goal position with additional L2-Regularization on the generated actions. The point-mass is controlled by two linear controllers, whose parameters need to be tuned by the agent. The controllers are switched as soon as the point-mass reaches the height of the gate, which is why the desired $y$-position of the controllers are fixed to $2.5$ (the height of the gate) and $0$, while all other parameters are controlled by the policy $\pi$, making $\cvec{\theta}$ a $14$-dimensional vector. We evaluate two setups in this gate environment, which differ in their target context distribution $\mu(\svec{c})$: In the first one, the agent needs to be able to steer through a very narrow gate far from the origin (``precision'') and in the second it is required to steer through gates with a variety of positions and widths (``global''). The two target context distributions are shown in Figure~\ref{fig:gate-experiments}. Figure~\ref{fig:gate-experiments} further visualizes the obtained rewards for the investigated algorithms, the evolution of the sampling distribution $p\left(\svec{c} \vert \cvec{\nu} \right)$ as well as tasks from the environment. In the ``global'' setting, we can see that SPRL converges significantly faster to the optimum than the other algorithms while in the ``precision'' setting, SPRL avoids a local optimum to which C-REPS and CMA-ES converge and which, as can be seen in Figure~\ref{fig:gate-successes}, does not encode desirable behavior. Furthermore, both curriculum learning algorithms SAGG-RIAC and GoalGAN only slowly escape this local optimum in the ``precision'' setting. We hypothesize that this slow convergence to the optimum is caused by SAGG-RIAC and GoalGAN not having a notion of a target distribution. Hence, these algorithms cannot guide the sampling of contexts to sample relevant tasks according to $\mu(\svec{c})$. This is especially problematic if $\mu(\svec{c})$ covers only a small fraction of the context space with a non-negligible probability density. The visualized sampling distributions in Figure~\ref{fig:gate-experiments} indicate that tasks with wide gates positioned at the origin seem to be easier to solve starting from the initially zero-mean Gaussian policy, as in both settings SPRL first focuses on these kinds of tasks and then subsequently changes the sampling distributions to match the target distribution. Interestingly, the search distribution of CMA-ES did not always converge in the ``precision'' setting, as can be seen in Figure~\ref{fig:gate-experiments}. This behavior persisted across various hyperparameters and population sizes.

\begin{figure}[t]
	\centering
	\begin{tikzpicture}
		\node (O0) {\input{./img/reacher-obstacle-default-results.tex}};
		\node [above right = -40pt and 0pt of O0] (O1) {\scalebox{0.055}{\includegraphics{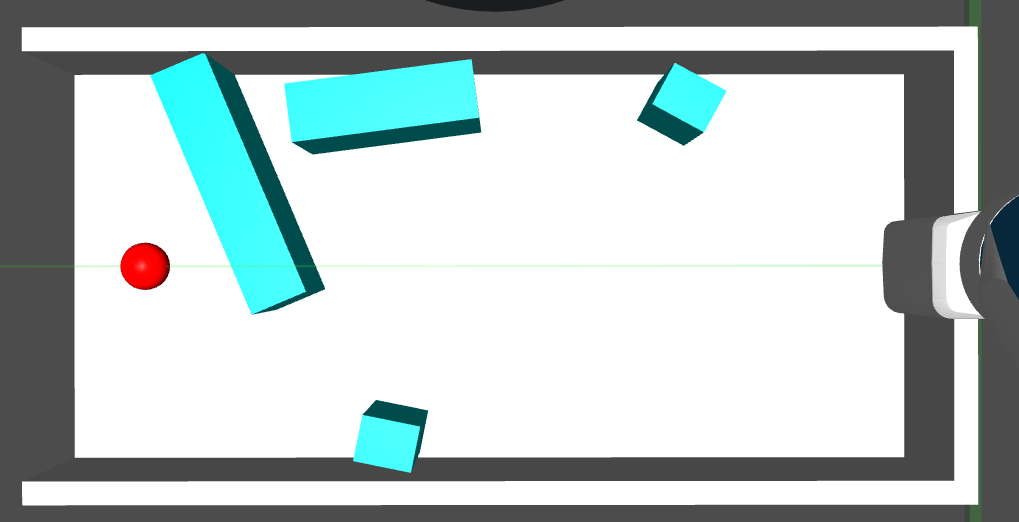}}};
		\node [below = -4pt of O1] (O2) {\scalebox{0.055}{\includegraphics{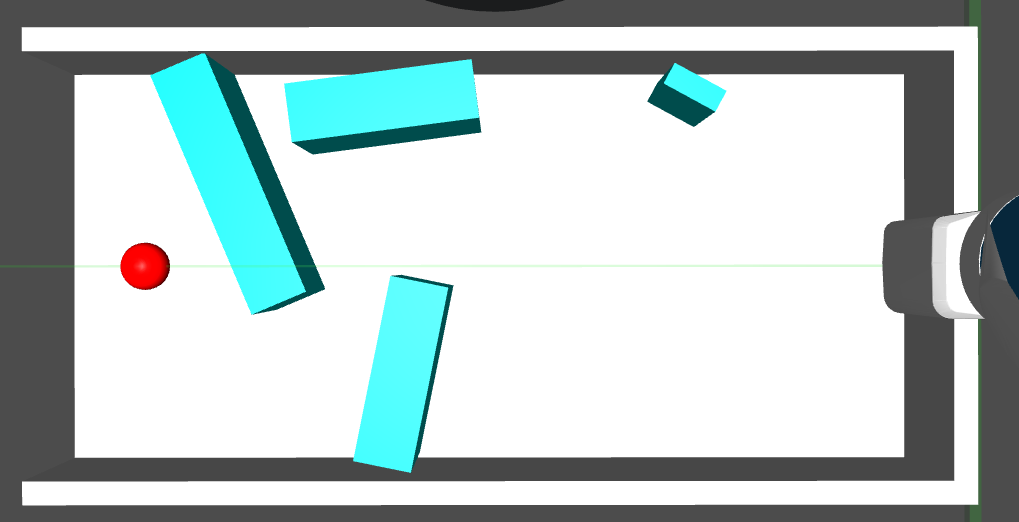}}};
		\node [below = -4pt of O2] (O3) {\scalebox{0.055}{\includegraphics{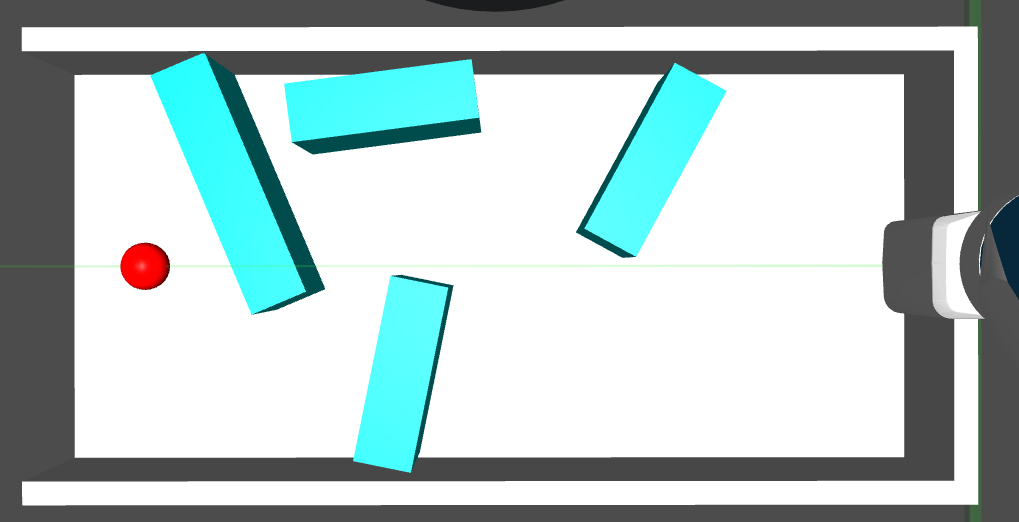}}};

		\node[cross=4pt, dotted, above left = -27pt and -52pt of O1] (C1) {};
		\node[cross=4pt, dotted, below left = -41pt and -247.5pt of O0] (C1) {};

		\node[cross=4pt, dash pattern={on 1pt off 0.75pt}, above left = -27pt and -52pt of O2] (C1) {};
		\node[cross=4pt, dash pattern={on 1pt off 0.75pt}, above left = -31pt and -247pt of O0] (C1) {};

		\node[cross=4pt, black, above left = -27pt and -52pt of O3] (C1) {};
		\node[cross=4pt, black, above left = -31pt and -311.5pt of O0] (C1) {};
	\end{tikzpicture}
	\vspace{-5pt}
		\caption[Avoid Experiments]{Left: $50\%$-quantiles (thick lines) and intervals from $10\%$- to $90\%$-quantile (shaded areas) of the reward in the reacher environment. Quantiles are computed over $40$ algorithm runs. Middle: The sampling distribution $p\left(\svec{c} \vert \cvec{\nu} \right)$ at different iterations (colored areas) of one SPRL run together with the target distribution (black line). Right: Task visualizations for different contexts with black crosses marking the corresponding positions in context space.}
	\label{fig:avoid-experiments}
	\vspace{-7pt}
\end{figure}

\begin{figure}[!b]
	\centering
	\input{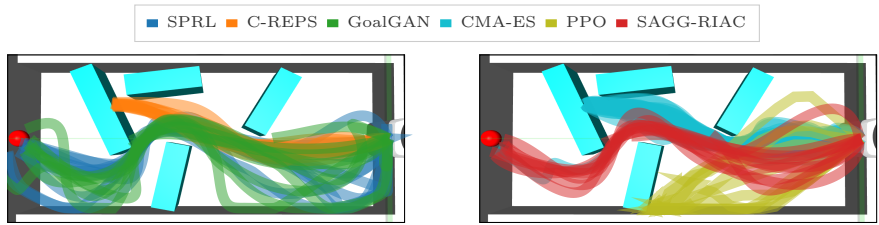}
	\caption[Gate Policies]{Trajectories generated by final policies learned with different algorithms in the reacher environment. The trajectories should reach the red dot while avoiding the cyan boxes. Please note that the visualization is not completely accurate, as we did not account for the viewpoint of the simulation camera when plotting the trajectories.}
	\label{fig:avoid-policies}
\end{figure}

\subsubsection{Reacher Environment}

For the next evaluation, we modify the three-dimensional reacher environment of the \mbox {OpenAI} Gym toolkit. In our version, the goal is to move the end-effector along the surface of a table towards the goal position while avoiding obstacles that are placed on the table. With the obstacles becoming larger, the robot needs to introduce a more pronounced curve movement to reach the goal without collisions. To simplify the visualization of the task distribution, we only allow two of the four obstacles to vary in size. The sizes of those two obstacles make up a task $\svec{c}$ in this environment. Just as in the first environment, the robot should not crash into the obstacles, and hence the movement is stopped if one of the four obstacles is touched. The policy $\pi$ encodes a ProMP ~\citep{paraschos-promps}, from which movements are sampled during training. In this task, $\cvec{\theta}$ is a $40$-dimensional vector.
\\
Looking at Figure~\ref{fig:avoid-experiments}, we can see that C-REPS and CMA-ES find a worse optimum compared to SPRL. This local optimum does---just as in the previous experiment---not encode optimal behavior, as we can see in Figure~\ref{fig:avoid-policies}. GoalGAN and SAGG-RIAC tend to find the same optimum as SPRL, however with slower convergence. This is nonetheless surprising given that---just as for the ``precision'' setting of the previous experiment---the algorithm deals with a narrow target context distribution. Although the $10$\%-$90$\% quantile of SAGG-RIAC and GoalGAN contain policies that do not manage to solve the task (i.e. are below the performance of C-REPS), the performance is in stark contrast to the performance in the previously discussed ``precision'' setting, in which the majority of runs did not solve the task. Nonetheless, the $10$\%-$50$\% quantile of the performance displayed in Figure~\ref{fig:avoid-experiments} still indicates the expected effect that SPRL leverages the knowledge of the target distribution to yield faster convergence to the optimal policy in the median case. 
\\
Another interesting artifact is the initial decrease in performance of SPRL between iterations $50-200$. This can be accounted to the fact that in this phase, the intermediate distribution $p(\svec{c} \vert \cvec{\nu})$ only assigns negligible probability density on areas covered by $\mu(\svec{c})$ (see Figure~\ref{fig:avoid-experiments}). Hence the agent performance on $\mu(\svec{c})$ during this stage is completely dependent on the extrapolation behavior of the agent, which seems to be rather poor in this setting. This once more illustrates the importance of appropriate transfer of behavior between tasks, which is, however, out of the scope of this paper.  
\\
The sampling distributions visualized in Figure~\ref{fig:avoid-experiments} indicate that SPRL focuses on easier tasks with smaller obstacle sizes first and then moves on to the harder, desired tasks. Figure~\ref{fig:avoid-policies} also shows that PPO \citep{schulman2017proximal}, a step-based reinforcement learning algorithm, is not able to solve the task after the same amount of interaction with the environment, emphasizing the complexity of the learning task.

\subsubsection{Sparse Ball-in-a-Cup}

\begin{figure}[b!]
	\centering
	\begin{tikzpicture}
		\node (O0){\input{./img/ball-in-a-cup-default-distribution-evolution.tex}};
		\node [above right = -60pt and -5pt of O0] (O1) {\scalebox{0.09}{\includegraphics{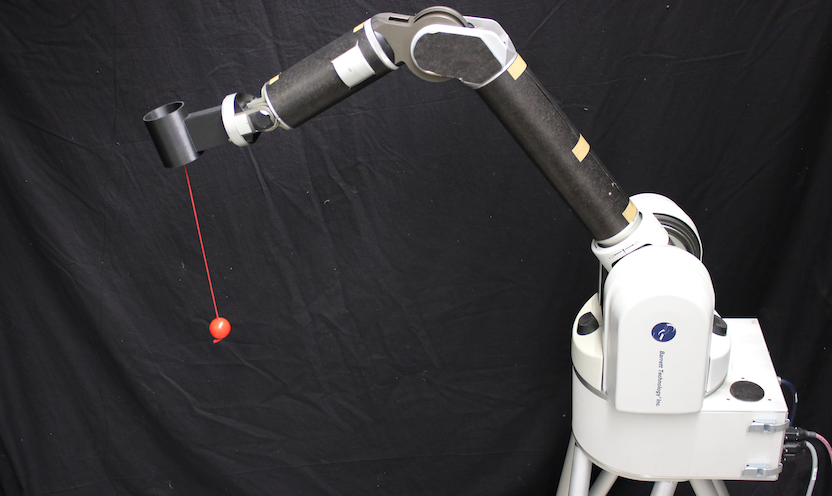}}};
		\node [below = -4pt of O1] (O2) {\scalebox{0.09}{\includegraphics{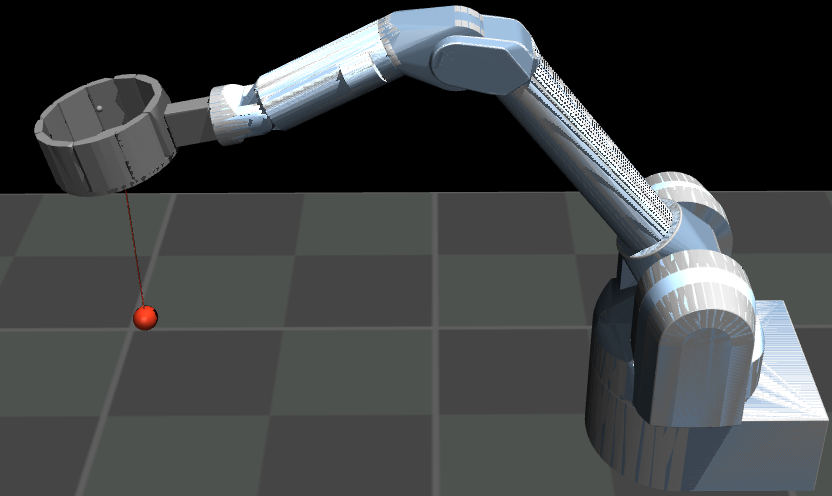}}};
	\end{tikzpicture}
	\vspace{-5pt}
	\caption[BIC Experiments]{Left: $50\%$-quantiles (thick lines) and intervals from $10\%$- to $90\%$-quantile (shaded areas) of the success rates for the sparse ball-in-a-cup task. Quantiles are computed from the $10$ best runs out of $20$. Middle: The sampling distribution $p\left(c \vert \cvec{\nu}\right)$ at different iterations (colored areas) of one SPRL run together with the target distribution $\mu(c)$ (black line). Right: Task visualization on the real robot (upper) and in simulation with a scale of $2.5$ (lower).}
	\label{fig:bic-results}
	\vspace{-10pt}
\end{figure}

We conclude this experimental evaluation with a ball-in-a-cup task, in which the reward function exhibits a significant amount of sparsity by only returning a reward of $1$ minus an L2 regularization term on the policy parameters, if the ball is in the cup after the policy execution, and $0$ otherwise. The robotic platform is a Barrett WAM, which we simulate using the MuJoCo physics engine \citep{todorov2012mujoco}. The policy represents again a ProMP encoding the desired position of the first, third and fifth joint of the robot. Achieving the desired task with a poor initial policy is an unlikely event, leading to mostly uninformative rewards and hence poor learning progress. However, as can be seen in Figure~\ref{fig:bic-results}, giving the learning agent control over the diameter of the cup significantly improves the learning progress by first training with larger cups and only progressively increasing the precision of the movement to work with smaller cups. Having access to only $16$ samples per iteration, the algorithms did not always learn to achieve the task. However, the final policies learned by SPRL outperform the ones learned by C-REPS, CMA-ES, GoalGAN and SAGG-RIAC. The movements learned in simulation were finally applied to the robot with a small amount of fine-tuning.

\section{Application to Step-Based Reinforcement Learning}
\label{sec:step-based-rl}

The experiments in the previous section demonstrate that the self-paced learning paradigm can indeed be beneficial in the episodic RL ---or black-box optimization--- setting, so that as a next step we want to investigate its application when using a stochastic policy of the form $\pi(\svec{a} \vert \svec{s}, \svec{c}, \cvec{\omega})$. In this setting, we derive an implementation of SPL that is agnostic to the RL algorithm of choice by using the possibility of updating the SPL objective (\ref{eq:kl-constrained-sprl}) in a block-coordinate manner w.r.t. $\cvec{\omega}$ and $\cvec{\nu}$. The resulting approximate implementation allows to create learning agents following the SPL paradigm using arbitrary RL algorithms by making use of the value functions that the RL algorithms estimate during policy optimization.

\subsection{Algorithmic Implementation}

\begin{algorithm}[t]
   \caption{Self-Paced Deep Reinforcement Learning (SPDL)}
   \label{alg:step-based-rl}
\begin{algorithmic}
   \STATE {\bfseries Input:} Initial context distribution- and policy parameters $\cvec{\nu}_0$ and $\cvec{\omega}_0$, Target context distribution $\mu(\svec{c})$, KL penalty proportion $\zeta$ and offset $K_{\alpha}$, Number of iterations $K$, Rollouts per policy update $M$, Relative entropy bound $\epsilon$
   \FOR{$k=1$ {\bfseries to} $K$}
   \STATE {\bfseries Agent Improvement:}
   \STATE Sample contexts: $\svec{c}_i \sim p(\svec{c} \vert \cvec{\nu}_k),\ i \in [1, M]$
   \STATE Rollout trajectories: $\cvec{\tau}_i \sim p(\cvec{\tau} \vert \svec{c}_i, \cvec{\omega}_k),\ i \in [1, M]$
   \STATE Obtain $\cvec{\omega}_{k + 1}$ from RL algorithm of choice using $\mathcal{D}_k = \left\{ (\svec{c}_i, \cvec{\tau}_i) \middle| i \in [1, M] \right\}$
   \STATE Estimate $\tilde{V}_{\cvec{\omega}_{k+1}}(\cvec{s}_{i, 0}, \svec{c}_i)$ (or use estimate of RL agent) for contexts $\svec{c}_i$
   \STATE {\bfseries Context Distribution Update:}
   \STATE \textbf{IF} $k \leq K_{\alpha}$: Obtain $\cvec{\nu}_{k+1}$ from (\ref{eq:spdl-approx-objective}) with $\alpha_k = 0$
   \STATE \textbf{ELSE}: Obtain $\cvec{\nu}_{k+1}$ optimizing (\ref{eq:spdl-approx-objective}), using $\alpha_k = \mathcal{B}(\cvec{\nu}_k, \mathcal{D}_k)$ (\ref{eq:alpha-computation})
   \ENDFOR
\end{algorithmic}
\end{algorithm}

\noindent When optimizing (\ref{eq:kl-constrained-sprl}) w.r.t. the policy parameters $\cvec{\omega}$ under the current context distribution $p(\svec{c} \vert \cvec{\nu}_k)$ using an RL algorithm of choice, a data set $\mathcal{D}_k$ of trajectories is generated
\begin{align*}
\mathcal{D}_k &{=}\left\{(\svec{c}_i, \cvec{\tau}_i)\ \middle|\ \svec{c}_i \sim p(\svec{c} \vert \cvec{\nu}_k), \cvec{\tau}_i \sim p(\cvec{\tau} \vert \svec{c}_i, \cvec{\omega}_k), i \in \left[1, M\right] \right\}. \\
\cvec{\tau}_i &= \{ (\svec{s}_{i,j}, \svec{a}_{i,j}, r_{i,j}) \vert \svec{a}_{i,j} {\sim} p(\svec{a} \vert \svec{s}_{i, j}, \svec{c}_i, \cvec{\omega}_k), \svec{s}_{i,j+1} {\sim} p_{\svec{c}_i}(\svec{s} \vert \svec{s}_{i,j}, \svec{a}_{i,j}), \svec{s}_{i,0} {\sim} p_{0, \svec{c}_i}(\svec{s}),  j {=} 1,\ldots \}.
\end{align*}
One unifying property of many RL algorithms is their reliance on estimating the state-value function $V_{\cvec{\omega}}(\svec{s}_0, \svec{c})$, each in their respective way, as a proxy to optimizing the policy. We make use of this approximated value function $\tilde{V}_{\cvec{\omega}}(\svec{s}_0, \svec{c})$ (note the $\sim$ indicating the approximation) to compute an estimate of the expected performance $J(\cvec{\omega}, \svec{c})$ in context~$\svec{c}$. Since $J(\cvec{\omega}, \svec{c}) = \mathbb{E}_{p_{0,\svec{c}}(\svec{s}_0)} \left[ V_{\cvec{\omega}}(\svec{s}_0, \svec{c}) \right]$, we can coarsely approximate the expectation w.r.t. $p_{0,\svec{c}}(\svec{s}_0)$ for a given context $\svec{c}_{i}$ with the initial state $\svec{s}_{i,0}$ contained in the set of trajectories $\mathcal{D}_k$. This yields an approximate form of (\ref{eq:kl-constrained-sprl}) given by
\begin{align}
\max_{\cvec{\nu}_{k+1}}\ &\frac{1}{M} \sum_{i=1}^M \frac{p\left(\svec{c}_i \vert \cvec{\nu}_{k+1}\right)}{p\left(\svec{c}_i \vert \cvec{\nu}_k\right)} \tilde{V}_{\cvec{\omega}}\left(\cvec{s}_{i,0}, \svec{c}_i\right) - \alpha_k \kldiv{p(\svec{c} \vert \cvec{\nu}_{k+1})}{\mu(\svec{c})} \nonumber \\ 
\text{s.t.}\ &\kldiv{p(\svec{c} \vert \cvec{\nu}_{k+1})}{p(\svec{c} \vert \cvec{\nu}_k)} {\leq} \epsilon. \label{eq:spdl-approx-objective}
\end{align}
The first term in objective (\ref{eq:spdl-approx-objective}) is an approximation to $\mathbb{E}_{p(\svec{c} \vert \cvec{\nu}_{k+1})} \left[ J(\cvec{\omega}, \svec{c}) \right]$ via importance-weights. The above objective can be solved using any constrained optimization algorithm. In our implementation, we use the trust-region algorithm implemented in the SciPy library \citep{2020SciPy-NMeth}. The two KL divergences in (\ref{eq:spdl-approx-objective}) can be computed in closed form since $\mu(\svec{c})$ and $p(\svec{c} \vert \cvec{\nu})$ are Gaussians in our implementations. However, for more complicated distributions, the divergences can also be computed using samples from the respective distributions and the corresponding (unnormalized) log-likelihoods. The resulting approach (SPDL) is summarized in Algorithm \ref{alg:step-based-rl}.

\subsection{Experiments}

We evaluate SPDL in three different environments (Figure \ref{fig:environments}) with different deep RL (DRL) algorithms: TRPO \citep{schulman2015trust}, PPO \citep{schulman2017proximal} and SAC \citep{haarnoja2018soft}. For all DRL algorithms, we use the implementations from the \texttt{Stable Baselines} library \citep{stable-baselines}. \footnote{Code for running the experiments can be found at \href{https://github.com/psclklnk/spdl}{\texttt{https://github.com/psclklnk/spdl}}.}
\\
\begin{wrapfigure}[20]{r}{8.29cm}
\vspace{-14pt}
\begin{tikzpicture}
\node (point1) {\scalebox{0.1}{\includegraphics{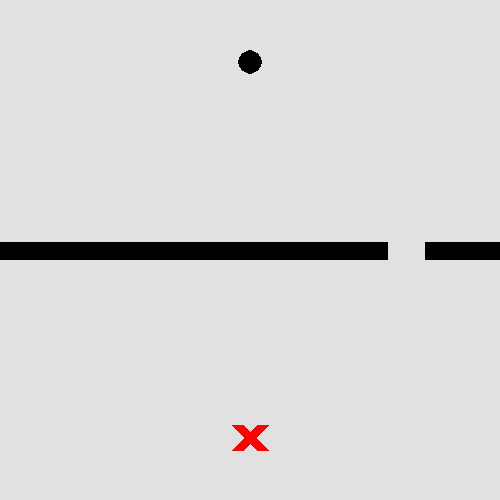}}};
\node [below = 5pt of point1] (point2) {\scalebox{0.1}{\includegraphics{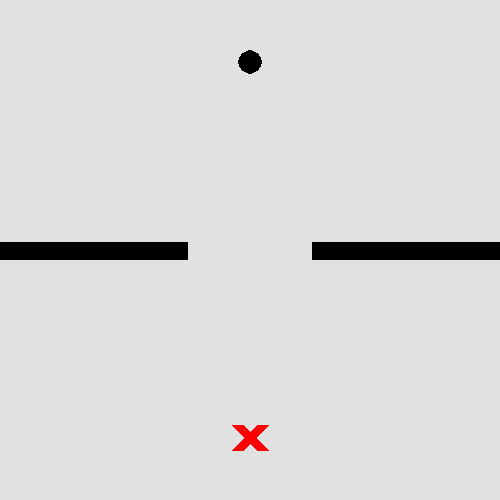}}};
\node [below = 0 pt of point2] {(a) Point-Mass};

\node [right = 5 pt of point1] (ant1) {\scalebox{0.07}{\includegraphics{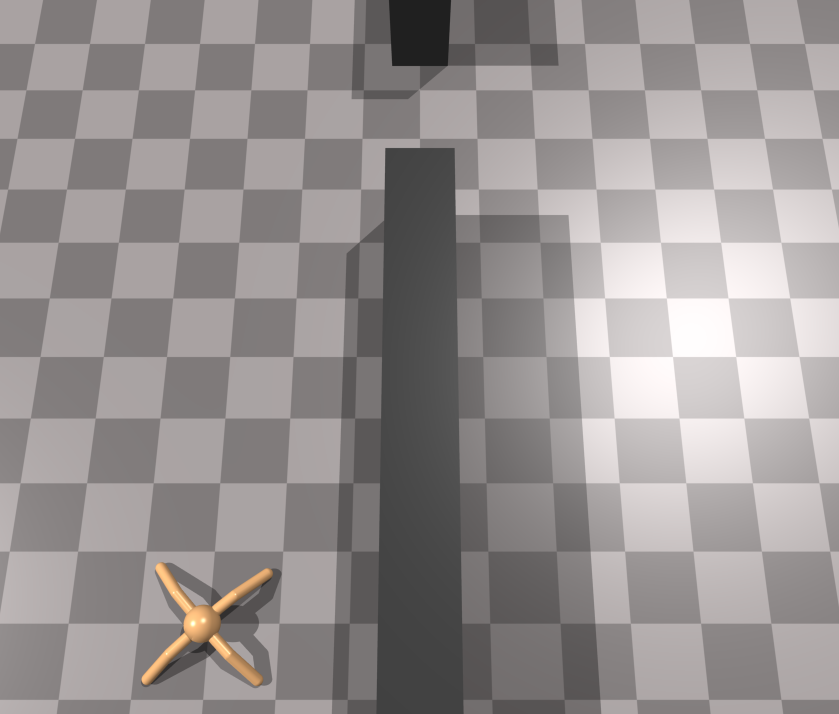}}};
\node [below = 5pt of ant1] (ant2) {\scalebox{0.07}{\includegraphics{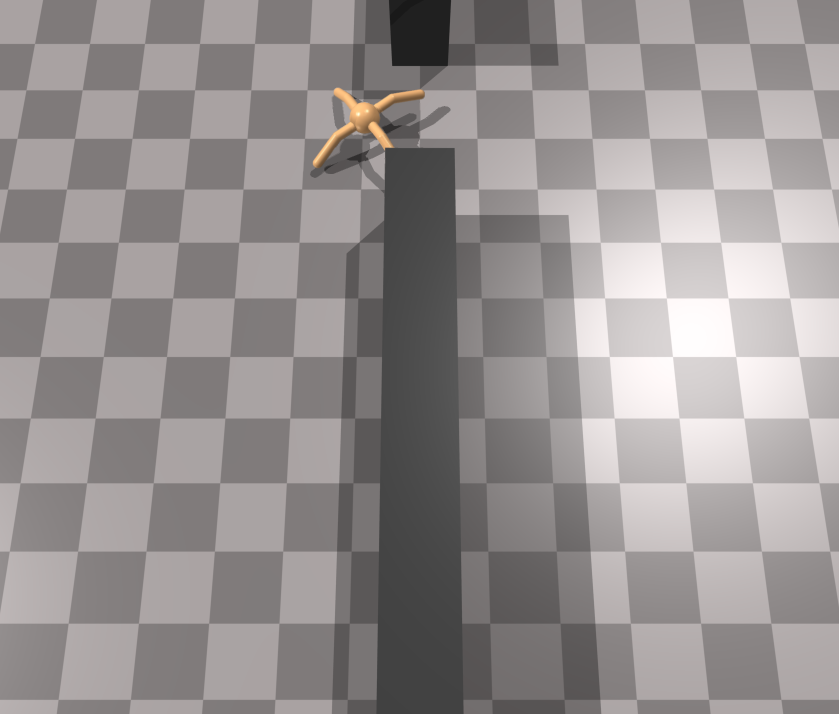}}};
\node [below = 0pt of ant2] {(b) Ant};

\node [below right = -57pt and 5 pt of ant1] (ball) {\scalebox{0.20}{\includegraphics{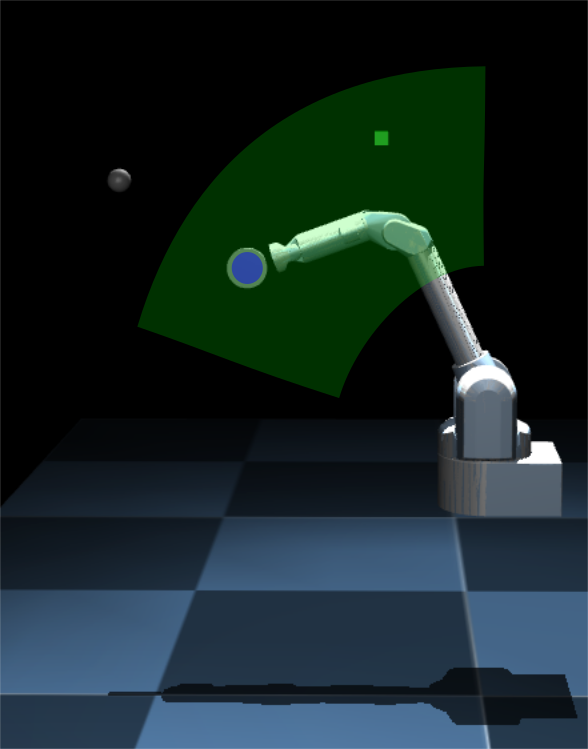}}};
\node [below = 0 pt of ball] {(c) Ball-Catching};
\end{tikzpicture}
\caption{Environments used for experimental evaluation. For the point-mass environment (a), the upper plot shows the target task. The shaded areas in picture (c) visualize the target distribution of ball positions (green) as well as the ball positions for which the initial policy succeeds (blue).}
\label{fig:environments}
\end{wrapfigure}
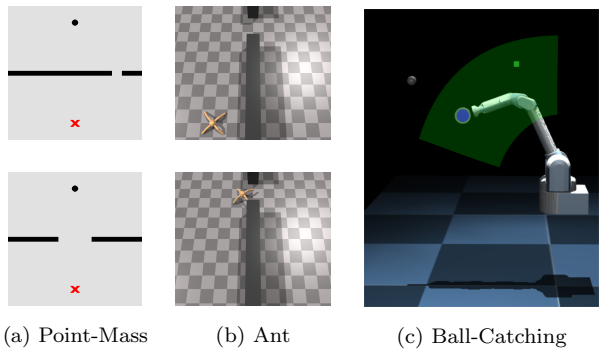The first environment for testing SPDL is again a point-mass environment but with an additional parameter to the context space, as we will detail in the corresponding section. The second environment extends the point-mass experiment by replacing the point-mass with a torque-controlled quadruped `ant', thus increasing the complexity of the underlying control problem and requiring the capacity of deep neural network function approximators used in DRL algorithms. Both of the mentioned environments will focus on learning a specific hard target task. The final environment is a robotic ball-catching environment. This environment constitutes a shift in curriculum paradigm as well as reward function. Instead of guiding learning towards a specific target task, this third environment requires to learn a ball-catching policy over a wide range of initial states (ball position and velocity). The reward function is sparse compared to the dense ones employed in the first two environments. To judge the performance of SPDL, we compare the obtained results to state-of-the-art CRL algorithms ALP-GMM~\citep{portelas2019teacher}, which is based on the concept of Intrinsic Motivation, and GoalGAN~\citep{florensa2018automatic}, which relies on the notion of a success indicator to define a curriculum. Further, we also compare to curricula consisting of tasks uniformly sampled from the context space (referred to as `Random' in the plots) and learning without a curriculum (referred to as `Default'). Additional details on the experiments as well as qualitative evaluations of them can be found in the appendix.

\subsubsection{Point-Mass Environment}

\begin{figure}[b!]
\centering
\includegraphics{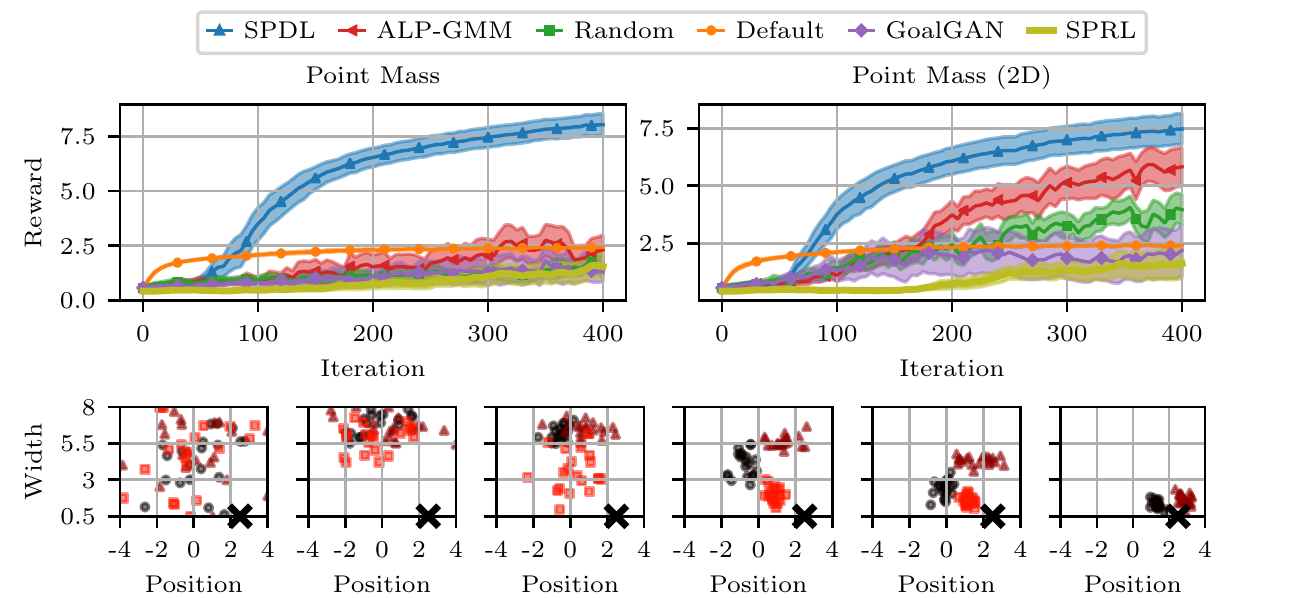}
\caption{Reward of different curricula in the point-mass (2D and 3D) environment for TRPO. Mean (thick line) and two times standard error (shaded area) is computed from $20$ algorithm runs. The lower plots show samples from the context distributions $p(\svec{c} \vert \cvec{\nu})$ in the point-mass 2D environment at iterations $0$, $20$, $30$, $50$, $65$ and $120$ (from left to right). Different colors and shapes of samples indicate different algorithm runs. The black cross marks the mean of the target distribution $\mu(\svec{c})$.}
\label{fig:point-mass-perf}
\vspace{-10pt}
\end{figure}

As previously mentioned, we again focus on a point-mass environment, where now the control policy is a neural network. Furthermore, the contextual variable $\svec{c} \in \mathbb{R}^{3}$ now  changes the width and position of the gate as well as the dynamic friction coefficient of the ground on which the point-mass slides. The target context distribution $\mu(\svec{c})$ is a narrow Gaussian with a negligible variance that encodes a small gate at a specific position and a dynamic friction coefficient of $0$. Figure \ref{fig:environments} shows two different instances of the environment, one of them being the target task.
\\
Figure \ref{fig:point-mass-perf} shows the results of two different experiments in this environment, one where the curriculum is generated over the full three-dimensional context space and one in which the friction parameter is fixed to its target value of $0$ so that the curriculum is generated only in a two-dimensional subspace. As Figure \ref{fig:point-mass-perf} and Table \ref{tab:point-mass} indicate, SPDL significantly increases the asymptotic reward on the target task compared to other methods. Increasing the dimension of the context space harms the performance of the other CRL algorithms. For SPDL, there is no statistically significant difference in performance across the two settings. This observation is in line with the hypothesis posed in Section \ref{sec:ep-rl-gate-exp}, that SPDL leverages the notion of $\mu(\svec{c})$ compared to other CRL algorithms that are not aware of it. As the context dimension increases, the volume of those parts in context space that carry non-negligible probability density according to $\mu(\svec{c})$ become smaller and smaller compared to the volume of the whole context space. Hence curricula that always target the whole context space tend to spend less time training on tasks that are relevant under $\mu(\svec{c})$. By having a notion of a target distribution, SPDL ultimately samples contexts that are likely according to $\mu(\svec{c})$, regardless of the dimension. The context distributions $p(\svec{c} \vert \cvec{\nu})$ visualized in Figure \ref{fig:point-mass-perf} show that the agent focuses on wide gates in a variety of positions in early iterations. Subsequently, the size of the gate is decreased and the position of the gate is shifted to match the target one. This process is carried out at different paces and in different ways, sometimes preferring to first shrink the width of the gate before moving its position while sometimes doing both simultaneously. More interestingly, the behavior of the curriculum is consistent with the one observed in Section \ref{sec:ep-rl-gate-exp}. 
We further see that the episodic version (SPRL), which we applied by defining the episodic RL policy $p(\cvec{\theta} \vert \svec{c}, \cvec{\omega})$ to choose the weights $\cvec{\theta}$ of a policy network for a given context $\svec{c}$, learns much slower compared to its step-based counterpart, requiring up to $800$ iterations to reach an average reward of $5$ (only the first $400$ are shown in Figure \ref{fig:point-mass-perf}). To keep the dimension of the context space moderate, the policy network for SPRL consisted of one layer of $21$ $\tanh$-activated hidden units, leading to $168$ and $189$ parameter dimensions in the two 2D and 3D context space instances. To make sure that the performance difference is not caused by different policy architectures, we also evaluated SPDL with this policy architecture, still significantly outperforming SPRL with an average reward of around $8$ after $800$ iterations.

\subsubsection{Ant Environment}

\begin{figure}[t]
\centering
\includegraphics{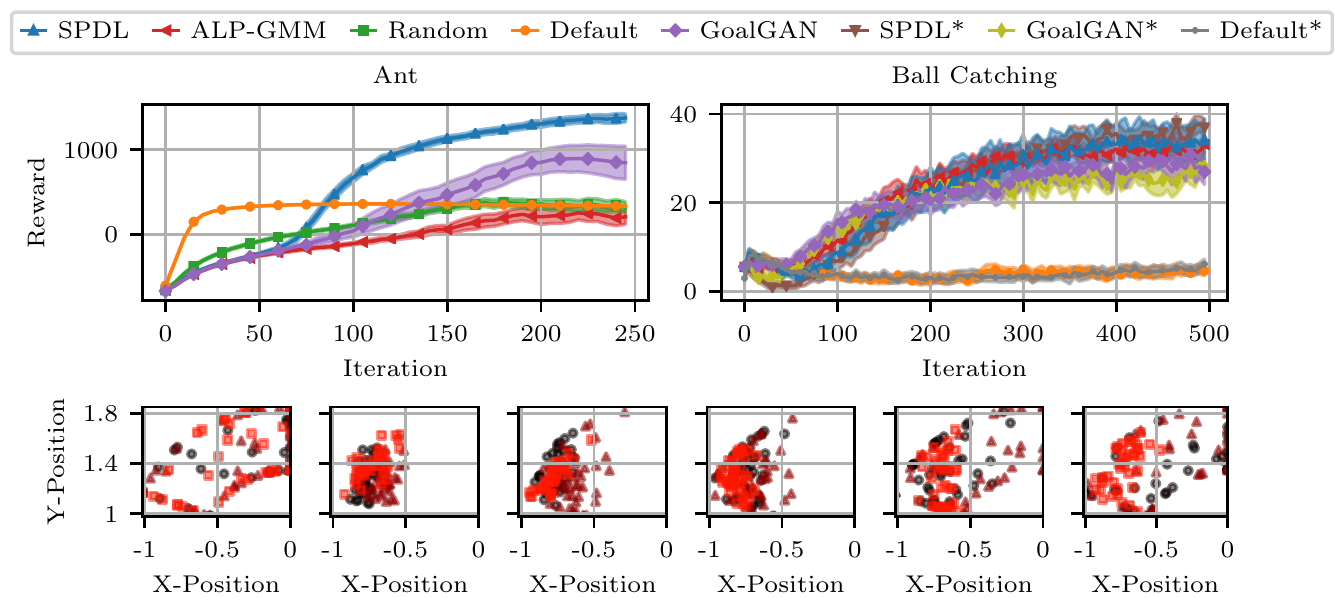}
\caption{Mean (thick line) and two times standard error (shaded area) of the reward achieved with different curricula in the ant environment for PPO and in the ball-catching environment for SAC (upper plots). The statistics are computed from $20$ seeds. For ball-catching, runs of SPDL/GoalGAN with an initialized context distribution and runs of Default learning without policy initialization are indicated by asterisks. The lower plots show ball positions in the `catching' plane sampled from the context distributions $p(\svec{c} \vert \cvec{\nu})$ in the ball-catching environment at iterations $0$, $50$, $80$, $110$, $150$ and $200$ (from left to right). Different sample colors and shapes indicate different algorithm runs. Given that $p(\svec{c} \vert \cvec{\nu})$ is initialized with $\mu(\svec{c})$, the samples in iteration $0$ visualize the target distribution.}
\label{fig:ant-perf}
\vspace{-7pt}
\end{figure}

We replace the point-mass in the previous environment with a four-legged ant similar to the one in the OpenAI Gym simulation environment \citep{brockman2016openai}. \footnote{We use the Nvidia Isaac Gym simulator \citep{isaac-gym} for this experiment.} The goal is to reach the other side of a wall by passing through a gate, whose width and position are determined by the contextual variable $\svec{c} \in \mathbb{R}^2$ (see Figure \ref{fig:environments}). 
In this environment, we were only able to evaluate the CRL algorithms using PPO. This is because the implementations of TRPO and SAC in the \texttt{Stable-Baselines} library do not allow to make use of the parallelization capabilities of the Isaac Gym simulator, leading to prohibitive running times (details in the appendix).
\\
Looking at Figure \ref{fig:ant-perf}, we see that SPDL allows the learning agent to escape the local optimum which results from the agent not finding the gate to pass through. ALP-GMM and a random curriculum do not improve the reward over directly learning on the target task. However, as we show in the appendix, both ALP-GMM and a random curriculum improve the qualitative performance, as they sometimes allow the ant to move through the gate. Nonetheless, this behavior is less efficient than the one learned by GoalGAN and SPDL, causing the action penalties in combination with the discount factor to prevent this better behavior from being reflected in the reward.

\subsubsection{Ball-Catching Environment}

Due to a sparse reward function and a broad target task distribution, this final environment is drastically different from the previous ones. In this environment, the agent needs to control a Barrett WAM robot to catch a ball thrown towards it. The reward function is sparse, only rewarding the robot when it catches the ball and penalizing excessive movements. In the simulated environment, the ball is considered caught if it is in contact with the end effector. The context $\svec{c} \in \mathbb{R}^3$ parameterizes the distance to the robot from which the ball is thrown as well as its target position in a plane that intersects the base of the robot. Figure \ref{fig:environments} shows the robot as well as the target distribution over the ball positions in the aforementioned `catching' plane. The context $\svec{c}$ is not visible to the policy, as it only changes the initial state distribution $p(\cvec{s}_0)$ via the encoded target position and initial distance to the robot. Given that the initial state is already observed by the policy, observing the context is superfluous. To tackle this learning task with a curriculum, we initialize the policy of the RL algorithms to hold the robot's initial position. This creates a subspace in the context space in which the policy already performs well, i.e. where the target position of the ball coincides with the initial end effector position. This can be leveraged by CRL algorithms.
\\
Since SPDL and GoalGAN support to specify the initial context distribution, we investigate whether this feature can be exploited by choosing the initial context distribution to encode the aforementioned tasks in which the initial policy performs well. When directly learning on the target context distribution without a curriculum, it is not clear whether the policy initialization benefits learning. Hence, we evaluate the performance both with and without a pre-trained policy when not using a curriculum.\begin{table*}[t]
\begin{center}
\vspace{-10pt}
\resizebox{\columnwidth}{!}{
\begin{small}
\begin{tabular}{lcccccr}
\toprule
 & PPO (P3D) & SAC (P3D) & PPO (P2D) & SAC (P2D) & TRPO (BC) & PPO (BC) \\
\midrule
ALP-GMM & $2.43\pm0.3$ & $4.68\pm0.8$ & $5.23\pm0.4$ & $5.11\pm0.7$ & $39.8\pm1.1$ & $46.5\pm0.7$\\
GoalGAN & $0.66\pm0.1$ & $2.14\pm0.6$ & $1.63\pm0.5$ & $1.34\pm0.4$ & $42.5\pm1.6$ & $42.6\pm2.7$\\
GoalGAN* & - & - & - & - & $45.8\pm1.0$ & $45.9\pm1.0$\\
SPDL & $\mathbf{8.45\pm0.4}$ & $6.85\pm0.8$ & $\mathbf{8.94 \pm 0.1}$ & $5.67\pm0.8$ & $47.0\pm2.0$ & $\mathbf{53.9\pm0.4}$ \\
SPDL* & - & - & - & - & $43.3\pm2.0$ & $49.3\pm1.4$ \\
Random & $0.67\pm0.1$ & $2.70\pm0.7$ & $2.49\pm0.3$ & $4.99\pm0.8$ & - & -\\
Default & $2.40\pm0.0$ & $2.47\pm0.0$ & $2.37\pm0.0$ & $2.40\pm0.0$ & $21.0\pm0.3$ & $22.1\pm0.3$\\
Default* & - & - & - & - & $21.2 \pm 0.3$ & $23.0 \pm 0.7$ \\
\bottomrule
\end{tabular}
\end{small}}
\end{center}
\caption{Average final reward and standard error of different curricula and RL algorithms in the two point-mass environments with three (P3D) and two (P2D) context dimensions as well as the ball-catching environment (BC). The data is computed from $20$ algorithm runs. Significantly better results according to Welch's t-test with $p<1\%$ are highlighted in bold. The asterisks mark runs of SPDL/GoalGAN with an initialized context distribution and runs of default learning without policy initialization.}
\label{tab:point-mass}
\vspace{-12pt}
\end{table*}
\\
Figure \ref{fig:ant-perf} and Table \ref{tab:point-mass} show the performance of the investigated curriculum learning approaches. We see that sampling tasks directly from the target distribution does not allow the agent to learn a meaningful policy, regardless of the initial one. Further, all curricula enable learning in this environment and achieve a similar reward. The results also highlight that initialization of the context distribution slightly improves performance for GoalGAN while slightly reducing performance for SPDL. The context distributions $p(\svec{c} \vert \cvec{\nu})$ visualized in Figure \ref{fig:ant-perf} indicate that SPDL shrinks the initially wide context distribution in early iterations to recover the subspace of ball target positions, in which the initial policy performs well. From there, the context distribution then gradually matches the target one. As in the point-mass experiment, this progress takes place at a differing pace, as can be seen in the visualizations of $p(\svec{c} \vert \cvec{\nu})$ in Figure~\ref{fig:ant-perf} for iteration $200$: Two of the three distributions fully match the target distribution while the third only covers half of it. \\
The similar performance across curriculum learning methods is indeed interesting. Clearly, the wide target context distribution $\mu(\svec{c})$ better matches the implicit assumptions made by both ALP-GMM and GoalGAN that learning should aim to accomplish tasks in the whole context space. However, both ALP-GMM and GoalGAN are built around the idea to sample tasks that promise a maximum amount of learning progress. This typically leads to a sampling scheme that avoids re-sampling tasks that the agent can already solve. However, SPDL achieves the same performance by simply growing the sampling distribution over time, not at all avoiding to sample tasks that the agent has mastered long ago. Hence, a promising direction for further improving the performance of CRL methods is to combine ideas of SPDL and methods such as ALP-GMM and GoalGAN.

\section{Improved $\alpha$-Schedule}
\label{sec:optimal-alpha}

\begin{figure}[t]
\vspace{-5pt}
\centering
\includegraphics{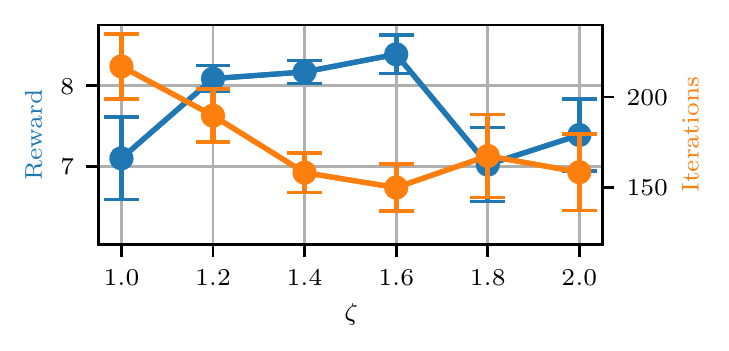}
\includegraphics{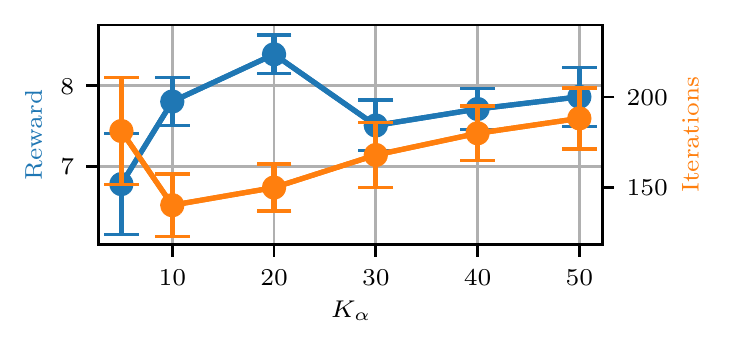}
\caption{Final performance (blue) of SPDL on the point mass (3D) environment for different values of $K_{\alpha}$ and $\zeta$ as well as the average number of iterations required to reach a reward larger or equal to $5$ on tasks sampled from $\mu(\svec{c})$ (orange). The results are computed from $20$ environment runs. The error bars indicate the standard error. When varying $\zeta$, $K_{\alpha}$ was fixed to a value of $20$. When varying $K_{\alpha}$, $\zeta$ was fixed to a value of $1.6$.}
\label{fig:perf-trade-off}
\vspace{-7pt}
\end{figure}

The evaluation in the previous settings showed that choosing the trade-off parameter $\alpha_k$ in each iteration according to (\ref{eq:alpha-computation}) was sufficient to improve the performance of the learner in the investigated experiments. However, the introduced schedule requires an offset parameter $K_{\alpha}$ as well as a penalty proportion $\zeta$ to be specified. Both these parameters need to be chosen adequately. If $K_{\alpha}$ is chosen too small, the agent may not have enough time to find a subspace of the context space containing tasks of adequate difficulty. If $K_{\alpha}$ is chosen too large, the learner wastes iterations focusing on tasks of small difficulty, not making progress towards the tasks likely under $\mu(\svec{c})$. The parameter $\zeta$ exhibits a similar trade-off behavior. Too small values will lead to an unnecessarily slow progression towards $\mu(\svec{c})$ while too large values lead to ignoring the competence of the learner on the tasks under $p(\svec{c} \vert \cvec{\nu})$, resulting in a poor final agent behavior because of a too speedy progression towards $\mu(\svec{c})$. This trade-off is visualized in Figure \ref{fig:perf-trade-off} for the point mass environment. 
\begin{table*}[b!]
\begin{center}
\begin{small}
\begin{tabular}{l|cc|cc}
\toprule
& \multicolumn{2}{c}{$\text{SPDL}(K_{\alpha}, \zeta)$} & \multicolumn{2}{c}{$\text{SPDL}(V_{\text{LB}})$}\\ \cline{2-3} \cline{4-5}
& Performance & Iterations to Threshold & Performance & Iterations to Threshold \\
\midrule
TRPO (P3D) & $8.04 \pm 0.25$ & $198 \pm 18$ & $7.79 \pm 0.28$ & $220 \pm 19$ \\
PPO (P3D) & $8.45 \pm 0.42$ & $165 \pm 14$ & $8.66 \pm 0.07$ & $\mathbf{120 \pm 10}$ \\
SAC (P3D) & $6.85 \pm 0.77$ & $94 \pm 8$ & $7.13 \pm 0.71$ & $\mathbf{67 \pm 4}$  \\
TRPO (P2D) & $7.47 \pm 0.33$ & $201 \pm 20$ & $7.67 \pm 0.2$ & $198 \pm 19$ \\
PPO (P2D) & $8.94 \pm 0.10$ & $132 \pm 5$ & $9.01 \pm 0.07$ & $\textcolor{brown}{\mathbf{119 \pm 3}}$ \\
SAC (P2D) & $5.67 \pm 0.77$ & $134 \pm 30$ & $6.56 \pm 0.82$ & $\textcolor{brown}{\mathbf{59 \pm 3}}$ \\
PPO (ANT) & $1371 \pm 23$ & $131 \pm 3$ & $1305 \pm 38$  & $131 \pm 2$ \\
TRPO (BC) & $47.0 \pm 2.0$ & $379 \pm 21$ & $50.0 \pm 1.5$ & $320 \pm 20$ \\
PPO (BC) & $53.9 \pm 0.4$ & $285 \pm 19$ & $51.6 \pm 1.7$ & $\textcolor{brown}{\mathbf{234 \pm 12}}$ \\
SAC (BC) & $34.1 \pm 2.3$ & $205 \pm 12$ & $34.1 \pm 1.3$ & $\mathbf{139 \pm 8}$ \\
TRPO (BC*) & $43.3 \pm 2.0$ & $354 \pm 18$ & $46.0 \pm 1.5$ & $\textcolor{brown}{\mathbf{285 \pm 20}}$ \\
PPO (BC*) & $49.3 \pm 1.4$ & $224 \pm 7$ & $51.8 \pm 0.5$ & $212 \pm 17$ \\
SAC (BC*) & $36.9 \pm 1.0$ & $235 \pm 23$ & $37.1 \pm 1.2$ & $\textcolor{brown}{\mathbf{173 \pm 20}}$ \\
\bottomrule
\end{tabular}
\end{small}
\end{center}
\caption{Comparison between the two SPDL heuristics on the point-mass (P3D and P2D), ant and ball-catching (BC) environments computed using $20$ runs. The asterisks mark runs of SPDL with an initialized context distribution. We compare both the final average reward $\pm$ standard error (Performance) as well as the average number of iterations required to reach $80\%$ of the lower of the two rewards (Iterations to Threshold). Statistically significant differences according to Welch's t-test are highlighted in \textbf{bold} for $p{<}1\%$  and \textcolor{brown}{\textbf{brown}} for $p{<}5\%$.}
\label{tab:spdlv2}
\vspace{-10pt}
\end{table*}
\noindent Despite the clear interpretation of the two parameters $K_{\alpha}$ and $\zeta$, which allows for a fairly straightforward tuning, we additionally explore another approach of choosing $\alpha_k$ in this section. This approach only requires to specify an expected level of performance $V_{\text{LB}}$ that the agent should maintain under the chosen context distribution $p(\svec{c} \vert \cvec{\nu})$. Assuming the decoupled optimization of the policy $\pi$ and the context distribution $p(\svec{c} \vert \cvec{\nu})$ investigated in the last section, this can be easily realized by rewriting (\ref{eq:kl-constrained-sprl}) as
\begin{align}
\min_{\cvec{\nu}}\ &\kldiv{p(\svec{c} \vert \cvec{\nu})}{\mu(\svec{c})} \nonumber \nonumber \\
\text{s.t.}\ &\mathbb{E}_{p(\svec{c} \vert \cvec{\nu})} \left[ J(\cvec{\omega}, \svec{c}) \right] \geq V_{\text{LB}} \nonumber \\
&\kldiv{p(\svec{c} \vert \cvec{\nu})}{p(\svec{c} \vert \cvec{\nu}')} \leq \epsilon. \label{eq:spdl-v2}
\end{align}
The main modification is to avoid the explicit trade-off between expected agent performance and KL divergence by minimizing the KL divergence w.r.t. $\mu(\svec{c})$ subject to a constraint on the expected agent performance. Investigating the Lagrangian of (\ref{eq:spdl-v2})
\begin{align*}
L(\cvec{\nu}, \alpha, \eta) =\ &\kldiv{p(\svec{c} \vert \cvec{\nu})}{\mu(\svec{c})} + \alpha (V_{\text{LB}} - \mathbb{E}_{p(\svec{c} \vert \cvec{\nu})} \left[ J(\cvec{\omega}, \svec{c}) \right]) \\
& \quad + \eta (\kldiv{p(\svec{c} \vert \cvec{\nu})}{p(\svec{c} \vert \cvec{\nu}')} - \epsilon), \qquad \qquad \alpha, \eta \geq 0
\end{align*}
we see that the constraint reintroduces a scalar $\alpha$ that trades-off the expected agent performance and KL divergence to $\mu(\svec{c})$. The value of this scalar is, however, now automatically chosen to fulfill the imposed constraint on the expected agent performance. For an implementation of (\ref{eq:spdl-v2}), we again replace $J(\cvec{\omega}, \svec{c})$ by an importance-sampled Monte Carlo estimate as done in (\ref{eq:spdl-approx-objective}). At this point, we have replaced the parameter $\zeta$ with $V_{\text{LB}}$. The benefit is a more intuitive choice of this hyperparameter since it is directly related to the expected performance, i.e. the quantity being optimized. Furthermore, we can easily remove the need for the offset parameter $K_{\alpha}$ by setting $\alpha_k {=} 0$ in (\ref{eq:spdl-approx-objective}) until we first reach $V_{\text{LB}}$. Consequently, this new schedule only requires one hyperparameter $V_{\text{LB}}$ to be specified. From then on, we compute the new context distribution by optimizing (\ref{eq:spdl-v2}). If during learning, the performance of the agent falls below $V_{\text{LB}}$ again, we simply do not change the context distribution until the performance exceeds $V_{\text{LB}}$ again. We now compare this new schedule with the schedule for $\alpha$ that is based on $K_{\alpha}$ and $\zeta$. The corresponding experiment data is shown in Table \ref{tab:spdlv2}. We can see that the final rewards achieved with the two heuristics are not significantly different according to Welch's t-test. This indicates that the simpler heuristic performs just as well in the investigated environments in terms of final reward. However, the often significantly smaller average number of iterations required to reach a certain average performance on $\mu(\svec{c})$ shows that the schedule based on $V_{\text{LB}}$ tends to lead to a faster progression towards $\mu(\svec{c})$. In a sense, this is not surprising, given that the explicit minimization of the KL divergence w.r.t. $\mu(\svec{c})$ under a constraint on the expected performance optimizes the trade-off between agent performance and KL divergence to $\mu(\svec{c})$ in each iteration. With the schedule based on $K_{\alpha}$ and $\zeta$, this trade-off was compressed into the parameter $\zeta$ that stayed constant for all iterations. Furthermore, the lower bound $V_{\text{LB}}$ on the achieved average reward under $p(\svec{c} \vert \cvec{\nu})$ exhibits certain similarities to the GoalGAN algorithm in which the context distribution resulted from a constraint of encoding only tasks of intermediate difficulty. In a sense, GoalGAN uses an upper and lower bound on the task difficulty. However, it enforces this constraint per context while our formulation enforces the lower bound in expectation.

\section{An Inference Perspective on Self-Paced Reinforcement Learning}
\label{sec:inference-view}

As noted in Section \ref{sec:sprl}, the KL divergence regularization w.r.t. $\cvec{\nu}$ in (\ref{eq:kl-constrained-sprl}) was done to stabilize the overall learning procedure. In this final section, we show that the resulting learning scheme can be connected to a modified version of the expectation-maximization algorithm, a well-known majorize-minimize algorithm for inference problems. Before we conclude this paper, we want to briefly point out this connection, especially highlighting a connection between self-paced learning and the concept of tempering \citep{kirkpatrick1983optimization,van1987simulated}.

\subsection{RL as Inference}

To establish the aforementioned connection, we need to introduce a probabilistic interpretation of the contextual RL problem as being an inference task \citep{dayan1997using,toussaint2006probabilistic,levine2018reinforcement}. In this formulation, the goal is to maximize the probability of an optimality event $\mathcal{O} \in \{0, 1\}$ that depends on the sum of rewards along a trajectory of states and actions $\cvec{\tau} = \{(\svec{s}_t, \svec{a}_t) \vert t = 0, 1,  \ldots \}$ 
\begin{align}
p(\mathcal{O} \vert \cvec{\tau}, \svec{c}) \propto \exp \left( R(\cvec{\tau}, \svec{c}) \right) = \exp \left( \sum_{t=0}^{\infty} r_{\svec{c}}(\svec{s}_t, \svec{a}_t) \right).
\end{align}
Together with the probability for a trajectory $\cvec{\tau}$ given the policy parameters $\cvec{\omega}$ 
\begin{align}
p(\cvec{\tau} \vert \svec{c}, \cvec{\omega}) = p_{0, \svec{c}}(\cvec{s}_0) \prod_{t \geq 0} \bar{p}_{\svec{c}}(\svec{s}_{t + 1} \vert \svec{s}_t, \svec{a}_t) \pi(\svec{s}_t \vert \svec{a}_t, \svec{c}, \cvec{\omega}) \label{eq:traj_prob}
\end{align}
we can marginalize over the trajectories $\cvec{\tau}$ generated by $p(\cvec{\tau} \vert \svec{c}, \cvec{\omega})$, i.e. generated by the agent. This results in a probabilistic interpretation of  the expected performance $J(\cvec{\omega}, \svec{c})$ in context $\svec{c}$ under policy parameters $\cvec{\omega}$. This probabilistic interpretation is given by the marginal likelihood of event $\mathcal{O}$ in an MDP $\mathcal{M}(\svec{c})$
\begin{align}
p(\mathcal{O} \vert \svec{c}, \cvec{\omega}) = \int p(\mathcal{O} \vert \cvec{\tau}, \svec{c}) p(\cvec{\tau} \vert \svec{c}, \cvec{\omega}) \dif \cvec{\tau}. \label{eq:rl-obs-prob}
\end{align}
The transition probabilities $\bar{p}_{\svec{c}}$ in (\ref{eq:traj_prob}) are a modified version of the original transition probabilities $p_{\svec{c}}$ that introduce a ``termination'' probability in each step that can occur with a probability of $1 - \gamma$ \citep[see][]{levine2018reinforcement}. This introduces the concept of a discounting factor $\gamma$ into the probabilistic model. The marginal likelihood (\ref{eq:rl-obs-prob}) is not exactly equivalent to the expected performance $J(\cvec{\omega}, \svec{c})$, however, as e.g. shown by \citet{levine2018reinforcement} or \citet{abdolmaleki2018maximum}, it motivates many successful RL algorithms and we refer to these works for a detailed discussion of this model as well as the differences to the ``default'' RL objective (\ref{eq:rl-objective}).  This is because we are more interested in the contextual version of this model, in which we introduce a context distribution $p(\svec{c} \vert \cvec{\nu})$ to obtain a probabilistic interpretation of the contextual RL objective
\begin{align}
p(\mathcal{O} \vert \cvec{\omega}, \cvec{\nu}) = \int p(\mathcal{O} \vert \cvec{\tau}, \svec{c}) p(\cvec{\tau} \vert \svec{c}, \cvec{\omega}) p(\svec{c} \vert \cvec{\nu}) \dif \svec{c} \dif \cvec{\tau} = \int p(\mathcal{O} \vert \svec{c}, \cvec{\omega}) p(\svec{c} \vert \cvec{\nu}) d\svec{c}. \label{eq:RL-LVM}
\end{align}
When not making use of a curriculum, we would simply set $p(\svec{c} \vert \cvec{\nu}) = \mu(\svec{c})$ and only optimize w.r.t. the policy parameters $\cvec{\omega}$. The above models (\ref{eq:rl-obs-prob}) and (\ref{eq:RL-LVM}) are called latent variable models (LVMs), as the trajectories $\cvec{\tau}$ (as well as the contexts $\svec{c}$) are marginalized out to form the likelihood of the event $\mathcal{O}$. These marginalizations make the direct optimization w.r.t. $\cvec{\omega}$ and $\cvec{\nu}$ challenging. The so-called expectation-maximization algorithm is commonly applied to split this complicated optimization into two simpler steps: The E- and M-Step
\begin{align}
\text{E-Step}:\ & q^k(\cvec{\tau}, \svec{c}) = \argmin_{q(\cvec{\tau}, \svec{c})}\ \kldiv{q(\cvec{\tau}, \svec{c})}{p(\cvec{\tau}, \svec{c} \vert \mathcal{O}, \cvec{\omega}^k, \cvec{\nu}^k)} \label{eq:rl-e-step} \\
\text{M-Step}:\ & \cvec{\omega}^{k+1}, \cvec{\nu}^{k+1} = \argmax_{\cvec{\omega}, \cvec{\nu}} \ \mathbb{E}_{q^k(\cvec{\tau}, \svec{c})} \left[ \log(p(\mathcal{O}, \cvec{\tau}, \svec{c} \vert \cvec{\omega}, \cvec{\nu}))\right]. \label{eq:rl-m-step}
\end{align}
Iterating between these two steps is guaranteed to find a local optimum of the marginal likelihood $p(\mathcal{O} \vert \cvec{\omega}, \cvec{\nu})$ and has been shown by e.g. \citet{abdolmaleki2018maximum} to motivate well-known RL algorithms.

\subsection{Connection to Self-Paced Reinforcement Learning}

At this point, two simple reformulations are required to establish the connection between the KL-regularized objective (\ref{eq:kl-constrained-sprl}) and the expectation-maximization algorithm on LVM (\ref{eq:RL-LVM}). First, we can reformulate the M-Step as an M-projection (i.e. a maximum-likelihood fit of the parametric model $q(\cvec{\tau}, \svec{c} \vert \cvec{\omega}, \cvec{\nu})$ to $q^k(\cvec{\tau}, \svec{c})$)
\begin{align*}
\argmax_{\cvec{\omega}, \cvec{\nu}} \ \mathbb{E}_{q^k(\cvec{\tau}, \svec{c})} \left[ \log(p(\mathcal{O}, \cvec{\tau}, \svec{c} \vert \cvec{\omega}, \cvec{\nu}))\right] = \argmin_{\cvec{\omega}, \cvec{\nu}} \kldiv{q^k(\cvec{\tau}, \svec{c})}{q(\cvec{\tau}, \svec{c} \vert \cvec{\omega}, \cvec{\nu})}.
\end{align*}
Second, the E-Step can, for this particular model, be shown to be equivalent to a KL-regularized RL objective
\begin{align*}
&\argmin_{q(\cvec{\tau}, \svec{c})}\ \kldiv{q(\cvec{\tau}, \svec{c})}{p(\cvec{\tau}, \svec{c} \vert \mathcal{O}, \cvec{\omega}^k, \cvec{\nu}^k)} \\
=\ &\argmax_{q(\cvec{\tau}, \svec{c})} \mathbb{E}_{q(\cvec{\tau}, \svec{c})} \left[ R\left( \cvec{\tau}, \svec{c} \right) \right] - \kldiv{q(\cvec{\tau}, \svec{c})}{p(\cvec{\tau}, \svec{c} \vert \cvec{\omega}^k, \cvec{\nu}^k)},
\end{align*}
in which we penalize a deviation of the policy and context distribution from the current parametric distribution $p(\cvec{\tau}, \svec{c} \vert \cvec{\omega}^k, \cvec{\nu}^k)$. Adding a term $-\alpha \kldiv{q(\svec{c})}{\mu(\svec{c})}$ and optimizing this modified E-Step only w.r.t. the context distribution $q(\svec{c})$ while keeping $q(\cvec{\tau} \vert \svec{c})$ fixed at $p(\cvec{\tau} \vert \svec{c}, \cvec{\omega}^k)$, we obtain
\begin{align}
\argmax_{q(\svec{c})} \mathbb{E}_{q(\svec{c})} \left[ \mathbb{E}_{p(\cvec{\tau} \vert \svec{c}, \cvec{\omega}^k)} \left[ R\left( \cvec{\tau}, \svec{c} \right) \right] \right] - \alpha \kldiv{q(\svec{c})}{\mu(\svec{c})} - \kldiv{q(\svec{c})}{p(\svec{c} \vert \cvec{\nu}^k)}. \label{eq:mod-e-step}
\end{align}
This result resembles (\ref{eq:kl-constrained-sprl}), where however the optimization is carried out w.r.t. $q(\svec{c})$ instead of $\cvec{\nu}$ and the KL divergence w.r.t. $p(\svec{c} \vert \cvec{\nu}^k)$ is treated as a penalty term instead of a constraint. Not fitting the parameters $\cvec{\nu}^{k+1}$ directly but in a separate (M-)step is also done by the \mbox{C-REPS} algorithm and our episodic RL implementation of SPL. Hence, in the light of these results, the step-based implementation can be interpreted as skipping an explicit M-Step and directly optimizing the E-Step w.r.t. to the parametric policy. Such a procedure can be found in popular RL algorithms, as detailed by \cite{abdolmaleki2018maximum}.

\subsection{Self-Paced Learning as Tempering}

The previously derived E-Step has a highly interesting connection to a concept in the inference literature called \textit{tempering} \citep{kirkpatrick1983optimization,van1987simulated}. This connection is revealed by showing that the penalty term $\alpha \kldiv{q(\svec{c})}{\mu(\svec{c})}$ in the modified E-Step (\ref{eq:mod-e-step}) results in an E-Step to a modified target distribution. That is
\begin{align}
&\argmin_{q(\svec{c})}\ \kldiv{q(\svec{c})}{p(\svec{c} \vert \mathcal{O}, \cvec{\omega}^k, \cvec{\nu}^k)} + \alpha \kldiv{q(\svec{c})}{\mu(\svec{c})} \nonumber \\
=\ & \argmin_{q(\svec{c})}\ \kldiv{q(\svec{c})}{\frac{1}{Z} p(\svec{c} \vert \mathcal{O}, \cvec{\omega}^k, \cvec{\nu}^k)^{\frac{1}{1 + \alpha}} \mu(\svec{c})^{\frac{\alpha}{1 + \alpha}}}. \label{eq:sprl-path}
\end{align}
The modified target distribution in (\ref{eq:sprl-path}) is performing an interpolation between $\mu(\svec{c})$ and $p(\svec{c} \vert \mathcal{O}, \cvec{\omega}^k, \cvec{\nu}^k)$ based on the parameter $\alpha$. Looking back at the sampling distribution induced by the probabilistic SPL objective (\ref{eq:pspl-objective}) for the regularizer $f_{\text{KL},i}$ (see Equation \ref{eq:sprl-weights} in the appendix)
\begin{align}
p(\svec{c} \vert \alpha, \cvec{\omega}) \propto \cvec{\nu}^*_{\text{KL}, \svec{c}}(\alpha, \cvec{\omega}) = \mu(\svec{c}) \exp\left( J(\cvec{\omega}, \svec{c}) \right)^{\frac{1}{\alpha}}, \label{eq:sprl-path-2}
\end{align}
we can see that, similarly to the modified E-Step, the distribution encoded by $\cvec{\nu}^*_{\text{KL}, \svec{c}}$, i.e. the optimizers of (\ref{eq:pspl-objective}), interpolates between $\mu(\svec{c})$ and the distribution $p(\svec{c} \vert \cvec{\omega}) \propto \exp\left( J(\cvec{\omega}, \svec{c}) \right)$. Both of these distributions would be referred to as tempered distributions in the inference literature.
\\
The concept of tempering has been explored in the inference literature as a tool to improve inference methods when sampling from or finding modes of a distribution $\mu(\svec{c})$ with many isolated modes of density \citep{kirkpatrick1983optimization,Marinari1992Simulated,ueda1995deterministic}. The main idea is to not directly apply inference methods to $\mu(\svec{c})$ but to make use of a tempered distribution $p_{\alpha}(\svec{c})$ which interpolates between $\mu(\svec{c})$ and a user-chosen reference distribution $\rho(\svec{c})$ from which samples can be easily drawn by the employed inference method (e.g. a Gaussian distribution). Doing repeated inference for varying values of $\alpha$ allows to explore the isolated modes more efficiently and with that yielding more accurate samples from $\mu(\svec{c})$. Intuitively, initially sampling from $\rho(\svec{c})$, chosen to be free from isolated modes, and gradually progressing towards $\mu(\svec{c})$ while using the previous inferences as initializations avoids getting stuck in isolated modes of $\mu(\svec{c})$ that encode comparatively low density. This technique makes the inference algorithm less dependent on a good initialization.
\\
We can easily identify both (\ref{eq:sprl-path}) and (\ref{eq:sprl-path-2}) to be particular tempered distributions $p_{\alpha}(\svec{c})$. There, however, seems to be a striking difference to the aforementioned tempering scheme: The target density $\mu(\svec{c})$ is typically trivial, not requiring any advanced inference machinery. However, although $\mu(\svec{c})$ may be trivial from an inference perspective, the density $p(\cvec{\omega} \vert \mathcal{O}) \propto  p(\cvec{\omega}) \int p(\mathcal{O} \vert \svec{c}, \cvec{\omega})  \mu(\svec{c}) \dif \svec{c}$, i.e. the posterior over policy parameters, is highly challenging for contexts $\svec{c}$ distributed according to $\mu(\svec{c})$. This is because it may contain many, highly isolated modes, many of which only encode suboptimal behavior. In these cases, tempering helps to achieve better performance when employed in combination with RL. For low values of $\alpha$, it is easier to find high-density modes of $p_{\alpha}(\cvec{\omega} \vert \mathcal{O}) \propto p(\cvec{\omega}) \int p(\mathcal{O} \vert \svec{c}, \cvec{\omega})  p_{\alpha}(\svec{c}) d\svec{c}$. These modes can then be ``tracked'' by the RL algorithm while increasing the value of $\alpha$.
\\
The connection between SPL and the concept of tempering yields interesting insights into the problem of choosing both a good schedule for $\alpha$ and also the general design of $p_{\alpha}(\svec{c})$. As introduced in Section \ref{sec:prel-self-paced-learning}, the particular choice of the self-paced regularizer $f(\alpha, \cvec{\nu})$, and hence the regularizer $F_{\alpha}(l)$, is closely related to the particular form of $p_{\alpha}(\svec{c})$. A ubiquitous decision is the choice of the particular regularizer or tempered distribution for a given problem. \cite{gelman1998simulating} show that the particular choice of $p_{\alpha}$ has a tremendous effect on the error of Monte Carlo estimates of ratios between normalization constants. Furthermore, they compute the optimal form of $p_{\alpha}$ for a Gaussian special case that reduces the variance of the Monte Carlo estimator. It may be possible to draw inspiration from their techniques to design regularizers specialized for problems that fulfill particular properties. \\
For the application of SPL to RL, another important design decision is the schedule of $\alpha$. The value of $\alpha$ should be increased as fast as possible while ensuring the stability of the RL agent. We proposed two schedules that accomplished this task sufficiently well. However, there may be a tremendous margin for improvement. In the inference literature, people readily investigated the problem of choosing $\alpha$, as they face a similar trade-off problem between required computation time of inference methods and the usefulness of their results \citep{mandt2016variational,graham2017continuously,luo2018thermostat}. Again, it may be possible to draw inspiration from these works to design better schedules for $\alpha$ in RL problems.

\section{Conclusion and Discussion}

We have presented an interpretation of self-paced learning as inducing a sampling distribution over tasks in a reinforcement learning setting when using the KL divergence w.r.t. a target distribution $\mu(\svec{c})$ as a self-paced regularizer. This view renders the induced curriculum as an approximate implementation of a regularized contextual RL objective that samples training tasks based on their contribution to the overall gradient of the objective. Furthermore, we identified our approximate implementations to be a modified version of the expectation-maximization algorithm applied to the common latent variable model for RL. These, in turn, revealed connections to the concept of tempering in the inference literature. 
\\
These observations motivate further theoretical investigations, such as identifying the particular regularized objective that is related to our approximate implementation (\ref{eq:con-rl-objective}). Furthermore, we only explored the KL divergence as a self-paced regularizer. Although we showed that the probabilistic interpretation of SPL does not hold for arbitrary regularizers, it may be possible to derive the presented results for a wider class of regularizers, such as f-divergences.
\\
From an experimental point of view, we focused particularly on RL tasks with a continuous context space in this work. In the future, we want to conduct experiments in discrete context spaces, where we do not need to restrict the distribution to some tractable analytic form since we can exactly represent discrete probability distributions.
\\
Our implementations of the SPL scheme for RL demonstrated remarkable performance across RL algorithms and tasks. The presented algorithms are, however, by far no perfect realizations of the theoretical concept. The proposed ways of choosing $\alpha$ in each iteration are just ad-hoc choices. At this point, insights gained through the inference perspective into our curriculum generation scheme presented in Section \ref{sec:inference-view} may be particularly useful. Furthermore, the use of Gaussian context distributions is a major limitation that restricts the flexibility of the context distribution. Specifically in higher-dimensional context spaces, such a restriction could lead to poor performance. Here, it may be possible to use advanced inference methods \citep{liu2018understanding,wibisono2018sampling} to sample from the distribution (\ref{eq:sprl-path}) without approximations even in continuous spaces.

\acks{This project has received funding from the DFG project PA3179/1-1 (ROBOLEAP) and from the European Union’s Horizon 2020 research and innovation programme under grant agreement No. 640554 (SKILLS4ROBOTS). Calculations for this research were conducted on the Lichtenberg high performance computer of the TU Darmstadt.}

\appendix

\section{Proof of Theorem 1}

We begin by restating the theorem from the main text
\retheo*
\begin{proof}
\noindent To prove the theorem, we make use of the result established by \cite{meng2017theoretical} that optimizing the SPL objective alternatingly w.r.t. $\cvec{\nu}$ and $\cvec{\omega}$
\begin{align*}
\cvec{\nu}^*, \cvec{\omega}^* = \argmin_{\cvec{\nu}, \cvec{\omega}}\ r(\cvec{\omega}) + \sum_{i=1}^N \left( \nu_i l(\svec{x}_i, y_i, \cvec{\omega}) + f(\alpha, \nu_i) \right), \qquad \alpha > 0.  \tag{\ref{eq:spl-objective}}
\end{align*}
is a majorize-minimize scheme applied to the objective
\begin{align*}
\min_{\cvec{\omega}} r(\cvec{\omega}) + \sum_{i=1}^N F_{\alpha}(l(\svec{x}_i, y_i, \cvec{\omega})), \qquad F_{\alpha}(l(\svec{x}_i, y_i, \cvec{\omega})) = \int_0^{l(\svec{x}_i, y_i, \cvec{\omega})} \nu^*(\alpha, \iota) \dif \iota. \tag{\ref{eq:spl-int-form}}
\end{align*}
Based on this result, the proof of Theorem \ref{theo:1} requires three steps: \textbf{First}, we need to show that the function
\begin{align*}
f_{\text{KL},i}(\alpha, \nu) = \alpha \nu \left( \log(\nu) - \log(\mu(c{=}i)) \right) - \alpha \nu, \tag{\ref{eq:pspl-kl-regularizer}}
\end{align*}
is a valid self-paced regularizer for objective (\ref{eq:spl-objective}) and that the corresponding objective (\ref{eq:spl-int-form}) has the form of the second objective in Theorem \ref{theo:1}. \textbf{Second}, we need to show the equivalence between the SPL objective (\ref{eq:spl-objective}) and the probabilistic objective (\ref{eq:pspl-objective}) for the regularizer $f_{\text{KL},i}$. \textbf{Finally}, we need to show that objective (\ref{eq:pspl-objective}) corresponds to the first objective in Theorem \ref{theo:1} when using $f_{\text{KL},i}$. We begin by restating the axioms of self-paced regularizers defined by \cite{jiang2015self} to prove the first of the three points. Again making use of the notation $\nu^*(\alpha, l) = \argmin_{\nu} \nu l + f(\alpha, \nu)$, these axioms are
\begin{enumerate}
\item $f(\alpha, \nu)$ is convex w.r.t. $\nu$
\item $\nu^*(\alpha, l)$ is monotonically decreasing w.r.t. $l$ and it holds that $\lim_{l \to 0} \nu^*(\alpha, l) = 1$ as well as $\lim_{l \to \infty} \nu^*(\alpha, l) = 0$
\item $\nu^*(\alpha, l)$ is monotonically decreasing w.r.t. $\alpha$ and it holds that $\lim_{\alpha \to \infty} \nu^*(\alpha, l) \leq 1$ as well as $\lim_{\alpha \to 0} \nu^*(\alpha, l) = 0$.
\end{enumerate}
It is important to note that, due to the term $\mu(c{=}i)$ in (\ref{eq:pspl-kl-regularizer}), there is now an individual regularizer $f_{\text{KL},i}$ for each sample. This formulation is in line with the theory established by \cite{meng2017theoretical} and simply corresponds to an individual regularizer $F_{\alpha, i}$ for each sample in (\ref{eq:spl-int-form}). Inspecting the second derivative of $f_{\text{KL},i}$ w.r.t. $\nu$, we see that $f_{\text{KL},i}(\alpha, \nu)$ is convex w.r.t. $\nu$. Furthermore, the solution to the SPL objective (\ref{eq:spl-objective}) 
\begin{align}
\nu_{\text{KL},i}^*(\alpha, l) = \mu(c{=}i) \exp \left(-\frac{1}{\alpha} l \right) \label{eq:sprl-weights}
\end{align}
fulfills above axioms except for $\lim_{l \to 0} \nu_{\text{KL},i}^*(\alpha, l) = 1$, since $\lim_{l \to 0} \nu_{\text{KL},i}^*(\alpha, l) = \mu(c{=}i)$. However, we could simply remove the log-likelihood term $\log(\mu(c{=}i))$ from $f_{\text{KL},i}(\alpha, \nu_i)$ and pre-weight each sample with $\mu(c{=}i)$, which would yield exactly the same curriculum while fulfilling all axioms. We stick to the introduced form, as it eases the connection of $f_{\text{KL},i}$ to the KL divergence between $p(c \vert \cvec{\nu})$ and $\mu(c)$. Given that we have ensured that $f_{\text{KL}, i}$ is a valid self-paced regularizer, we know that optimizing the SPL objective (\ref{eq:spl-objective}) under $f_{\text{KL},i}$ corresponds to employing the non-convex regularizer
\begin{align}
F_{\text{KL}, \alpha, i}(l(\svec{x}_i, y_i, \cvec{\omega})) = \int_0^{l(\svec{x}_i, y_i, \cvec{\omega})} \nu_{\text{KL},i}^*(\alpha, \iota)  \dif \iota = \mu(c {=} i) \alpha \left( 1 - \exp \left( -\frac{1}{a} l(\svec{x}_i, y_i, \cvec{\omega}) \right) \right).
\end{align}
Put differently, optimizing the SPL objective (\ref{eq:spl-objective}) with $r(\cvec{\omega}) = 0$ under $f_{\text{KL},i}$ corresponds to optimizing
\begin{align*}
\min_{\cvec{\omega}} \sum_{i=1}^N F_{\text{KL}, \alpha, i}(l(\svec{x}_i, y_i, \cvec{\omega})) = \min_{\cvec{\omega}} \mathbb{E}_{\mu(c)} \left[ \alpha \left( 1 - \exp \left( -\frac{1}{a} l(\svec{x}_c, y_c, \cvec{\omega}) \right) \right) \right],
\end{align*}
as stated in Theorem \ref{theo:1}. As a next step, we notice that entries in the optimal $\cvec{\nu}$ for a given $\cvec{\omega}$ and $\alpha$ in the probabilistic SPL objective (\ref{eq:pspl-objective}) are proportional to $\nu_{\text{KL},i}^*(\alpha, l)$ in (\ref{eq:sprl-weights}), where the factor of proportionality $Z$ simply rescales the variables $\nu_{\text{KL},i}^*$ so that they fulfill the normalization constraint of objective (\ref{eq:pspl-objective}). Since 
$$
\mathbb{E}_{p(c \vert \cvec{\nu})} \left[ f(\svec{x}_c, y_c, \cvec{\omega}) \right] = \sum_{i=1}^N \nu_i f(\svec{x}_i, y_i, \cvec{\omega})
$$ 
by definition of $p(c \vert \cvec{\nu})$ introduced in Section \ref{sec:prob-self-paced-learning}, we see that consequently the only difference between (\ref{eq:spl-objective}) and (\ref{eq:pspl-objective}) for this particular regularizer is a different weighting of the regularization term $r(\cvec{\omega})$ throughout the iterations of SPL. More precisely, $r(\cvec{\omega})$ is weighted by the aforementioned factor of proportionality $Z$. Since $r(\cvec{\omega}) = 0$ in Theorem \ref{theo:1}, SPL (\ref{eq:spl-objective}) and the probabilistic interpretation introduced in this paper (\ref{eq:pspl-objective}) are exactly equivalent, since a constant scaling does not change the location of the optima w.r.t $\cvec{\omega}$ in both (\ref{eq:spl-objective}) and (\ref{eq:pspl-objective}). Consequently, we are left with proving that the PSPL objective (\ref{eq:pspl-objective}) under $f_{\text{KL}, i}$ and $r(\cvec{\omega}) = 0$ is equal to the first objective in Theorem \ref{theo:1}. This reduces to proving that $\sum_{i=1}^N f_{\text{KL},i}(\alpha, \nu_i)$ is equal to the KL divergence between $p(c \vert \cvec{\nu})$ and $\mu(c)$. Remembering $p(c{=}i \vert \cvec{\nu}) = \nu_i$, it follows that
\begin{align*}
\sum_{i=1}^N f_{\text{KL},i}(\alpha, \nu_i) &= \alpha \sum_{i=1}^N p(c{=}i \vert \cvec{\nu}) \left( \log(p(c{=}i \vert \cvec{\nu})) - \log(\mu(c{=}i)) \right)  - \alpha \sum_{i=1}^N p(c{=}i \vert \cvec{\nu}) \\
&= \alpha \kldiv{p(c{=}i \vert \cvec{\nu})}{\mu(c{=}i)} - \alpha.
\end{align*}
The removal of the sum in the second term is possible because $\sum_{i=1}^N p(c{=}i \vert \cvec{\nu}) = 1$ per definition of a probability distribution. Since the constant value $\alpha$ does not change the optimization w.r.t. $\cvec{\nu}$,  this proves the desired equivalence and with that Theorem \ref{theo:1}.
\end{proof}

\section{Self-Paced Episodic Reinforcement Learning Derivations}

This appendix serves to highlight some important details regarding the derivation of the weights (\ref{eq:episodic-pol-update}) and (\ref{eq:episodic-con-update}) as well as the dual objective (\ref{eq:sprl_dual}). The most notable detail is the introduction of an additional distribution $q(\svec{c})$ that takes the role of the marginal $\int q(\cvec{\theta}, \svec{c}) \dif \cvec{\theta}$ as well as the regularization of this additional distribution via a KL divergence constraint w.r.t. to the previous marginal $p(\svec{c}) = \int p(\cvec{\theta}, \svec{c}) \dif \cvec{\theta}$. This yields the following objective
\begin{alignat*}{2}
\max_{q(\cvec{\theta}, \svec{c}), q(\svec{c})}\ &\mathbb{E}_{q(\cvec{\theta}, \svec{c})} \left[ r\left( \cvec{\theta}, \svec{c} \right) \right] - \alpha \kldiv{q(\svec{c})}{\mu(\svec{c})} && \\
\text{s.t.}\ &\kldiv{q(\cvec{\theta}, \svec{c})}{p(\cvec{\theta}, \svec{c})} \leq \epsilon && \int q(\cvec{\theta}, \svec{c}) \dif \svec{c} \dif \cvec{\theta} = 1 \\
&\kldiv{q(\svec{c})}{p(\svec{c})} \leq \epsilon && \int q(\svec{c}) \dif \svec{c} = 1 \\
&\int q(\cvec{\theta}, \svec{c}) \dif \cvec{\theta} = q(\svec{c}) \quad \forall \svec{c} \in \mathcal{C}. &&
\end{alignat*}
However, these changes are purely of technical nature as they allow to derive numerically stable weights and duals. It is straightforward to verify that $\kldiv{q(\cvec{\theta}, \svec{c})}{p(\cvec{\theta}, \svec{c})} {\leq} \epsilon$ implies $\kldiv{q(\svec{c})}{p(\svec{c})} {\leq} \epsilon$. Hence, the constraint $\int q(\cvec{\theta}, \svec{c}) \dif \cvec{\theta} = q(\svec{c})$ guarantees that a solution $q(\cvec{\theta}, \svec{c})$ to above optimization problem is also a solution to (\ref{eq:sprl-episodic}). The dual as well as the weighted updates now follow from the Lagrangian
\begin{align}
\mathcal{L}(q, V, \eta_q, \eta_{\tilde{q}}, \lambda_q, \lambda_{\tilde{q}}) =\ \mathbb{E}_{q(\cvec{\theta}, \svec{c})} &\left[ r\left( \cvec{\theta}, \svec{c} \right) \right] - \alpha \kldiv{q(\svec{c})}{\mu(\svec{c})} \nonumber \\
&+ \eta_q \left( \epsilon - \kldiv{q(\cvec{\theta}, \svec{c})}{p(\cvec{\theta}, \svec{c})} \right) + \lambda_q \left( 1 - \int q(\cvec{\theta}, \svec{c}) \dif \svec{c} \dif \cvec{\theta} \right) \nonumber \\
&+ \eta_{\tilde{q}} \left( \epsilon - \kldiv{q(\svec{c})}{p(\svec{c})} \right) + \lambda_{\tilde{q}} \left( 1 - \int q(\svec{c}) \dif \svec{c} \right) \nonumber \\
& + \int V(\svec{c}) \left( \int q(\cvec{\theta}, \svec{c}) \dif \cvec{\theta} - q(\svec{c}) \right) \dif \svec{c}. \label{eq:lag}
\end{align}
Note that we slightly abuse notation and overload the argument $q$ in the definition of the Lagrangian. The update equations (\ref{eq:episodic-pol-update}) and (\ref{eq:episodic-con-update}) follow from the two conditions $\frac{\partial L}{\partial q(\cvec{\theta}, \svec{c})} = 0$ and $\frac{\partial L}{\partial q(\svec{c})} \mathcal{L} = 0$. Inserting (\ref{eq:episodic-pol-update}) and (\ref{eq:episodic-con-update}) into equation (\ref{eq:lag}) then allows to derive the dual (\ref{eq:sprl_dual}). We refer to \cite{van2017non} for detailed descriptions on the derivations in the non-contextual setting, which however generalize to the one investigated here.

\section{Regularized Policy Updates}

In order to enforce a gradual change in policy and context distribution not only during the computation of the weights via equations (\ref{eq:episodic-pol-update}) and (\ref{eq:episodic-con-update}) but also during the actual inference of the new policy and context distribution, the default weighted linear regression and weighted maximum likelihood objectives need to be regularized. Given a data set of $N$ weighted samples 
\begin{align*}
D &= \left\{(w_i^{\svec{x}}, w_i^{\svec{y}}, \svec{x}_i, \svec{y}_i) \vert i = 1,\ldots, N \right\},
\end{align*}
with $\svec{x}_i \in \mathbb{R}^{d_{\svec{x}}}, \svec{y}_i \in \mathbb{R}^{d_{\svec{y}}}$, the task of fitting a joint-distribution
\begin{align*}
q(\svec{x}, \svec{y}) &= q_{\svec{y}}(\svec{y} \vert \svec{x}) q_{\svec{x}}(\svec{x}) = \mathcal{N}(\svec{y} \vert \cvec{A} \phi(\svec{x}), \cvec{\Sigma}_{\svec{y}}) \mathcal{N}(\svec{x} \vert \cvec{\mu}_{\svec{x}}, \cvec{\Sigma}_{\svec{x}})
\end{align*}
to $D$ while limiting the change with regards to a reference distribution
\begin{align*}
p(\svec{x}, \svec{y}) &= p_{\svec{y}}(\svec{y} \vert \svec{x}) p_{\svec{x}}(\svec{x}) = \mathcal{N}(\svec{y} \vert \tilde{\cvec{A}} \phi(\svec{x}), \tilde{\cvec{\Sigma}}_{\svec{y}}) \mathcal{N}(\svec{x} \vert \tilde{\cvec{\mu}}_{\svec{x}}, \tilde{\cvec{\Sigma}}_{\svec{x}}),
\end{align*}
with feature function $\phi: \mathbb{R}^{d_x} \mapsto \mathbb{R}^{o}$, can be expressed as a constrained optimization problem
\begin{align*}
\max_{\cvec{A}, \cvec{\Sigma}_{\svec{y}}, \cvec{\mu}_{\svec{x}}, \cvec{\Sigma}_{\svec{x}}} &\sum_{i = 1}^N ( w^{\svec{x}}_i \log(q_{\svec{x}}(\svec{x}_i)) + w^{\svec{y}}_i \log(q_{\svec{y}}(\svec{y}_i \vert \svec{x}_i)) ) \\
\mathrm{s.t.}\ & \kldiv{p}{q} \approx \frac{1}{N} \sum_{i = 1}^N \kldiv{p_{\svec{y}}(\cdot \vert \svec{x}_i)}{q_{\svec{y}}(\cdot \vert \svec{x}_i)} + \kldiv{p_{\svec{x}}}{q_{\svec{x}}} \leq \epsilon.
\end{align*}
Note that we employ the reverse KL divergence in the constraint as this is the only form that allows for a closed form solution w.r.t. the parameters of the Gaussian distribution. Due to the unimodal nature of Gaussian distributions as well as the typically small value of $\epsilon$ this is a reasonable approximation. Since the distributions $p_{\svec{x}}$, $p_{\svec{y}}$, $q_{\svec{x}}$ and $q_{\svec{y}}$ are Gaussians, the KL divergences can be expressed analytically. Setting the derivative of the Lagrangian with respect to the optimization variables to zero yields to following expressions of the optimization variables in terms of the multiplier $\eta$ and the samples from $D$
\begin{align*}
\cvec{A} &= \left[ \sum_{i = 1}^N \left(w_i \svec{y}_i + \frac{\eta}{N} \tilde{\cvec{A}} \phi(\svec{x}_i) \right) \phi(\svec{x}_i)^T \right] \left[ \sum_{i = 1}^N \left(w_i + \frac{\eta}{N}\right) \phi(\svec{x}_i) \phi(\svec{x}_i)^T \right]^{-1}, \\
\\
\cvec{\Sigma}_{\svec{y}} &= \frac{\sum_{i = 1}^N w_i \Delta\svec{y}_i \Delta\svec{y}_i^T + \eta \tilde{\cvec{\Sigma}}_{\svec{y}}+ \frac{\eta}{N} \Delta\cvec{A} \sum_{i = 1}^N \phi(\svec{x}_i) \phi(\svec{x}_i)^T \Delta\cvec{A}^T}{\sum_{i = 1}^N w_i + \eta}, \\
\\
\cvec{\mu}_{\svec{x}} &= \frac{\sum_{i =1}^N w_i \svec{x}_i + \eta \tilde{\cvec{\mu}}_{\svec{x}}}{\sum_{i = 1}^N w_i + \eta}, \\
\\
\cvec{\Sigma}_{\svec{x}} &= \frac{\sum_{i = 1}^N w_i (\svec{x}_i - \cvec{\mu}_{\svec{x}}) (\svec{x}_i - \cvec{\mu}_{\svec{x}})^T + \eta \left( \tilde{\cvec{\Sigma}}_{\svec{x}} + (\cvec{\mu}_{\svec{x}} - \tilde{\cvec{\mu}}_{\svec{x}}) (\cvec{\mu}_{\svec{x}} - \tilde{\cvec{\mu}}_{\svec{x}})^T \right)}{\sum_{i = 1}^N w_i + \eta},
\end{align*}
with $\Delta\svec{y}_i = \svec{y}_i - \cvec{A} \phi(\svec{x}_i)$ and $\Delta\cvec{A} = \cvec{A} - \tilde{\cvec{A}}$. Above equations yield a simple way of enforcing the KL bound on the joint distribution: Since $\eta$ is zero if the constraint on the allowed KL divergence is not active, $\cvec{A}$, $\cvec{\Sigma}_{\svec{y}}$, $\cvec{\mu}_{\svec{x}}$ and $\cvec{\Sigma}_{\svec{x}}$ can be first computed with $\eta = 0$ and only if the allowed KL divergence is exceeded, $\eta$ needs to be found by searching the root of
$$
f(\eta) = \epsilon - \frac{1}{N} \sum_{i = 1}^N \kldiv{p_{\svec{y}}(\cdot \vert \svec{x}_i)}{q_{\svec{y}}(\cdot \vert \svec{x}_i)} + \kldiv{p_{\svec{x}}}{q_{\svec{x}}},
$$
where $q_{\svec{y}}$ and $q_{\svec{x}}$ are expressed as given by above formulas and hence implicitly depend on $\eta$. As this is a one-dimensional root finding problem, simple algorithms can be used for this task.

\section{Experimental Details}

This section is composed of further details on the experiments in sections \ref{sec:episodic-rl} and \ref{sec:step-based-rl}, which were left out in the main paper to improve readability. The details are split between the episodic- and step-based scenarios as well as the individual experiments conducted in them.\\
To conduct the experiments, we use the implementation of ALP-GMM, GoalGAN and SAGG-RIAC provided in the repositories accompanying the papers from \cite{florensa2018automatic} and \cite{portelas2019teacher} as well as the CMA-ES implementation from \cite{hansen2019pycma}. The employed hyperparameters are discussed in the corresponding sections. \\
Conducting the experiments with SPRL and SPDL, we found that restricting the standard deviation of the context distribution $p(\svec{c} \vert \cvec{\nu})$ to stay above a certain lower bound $\cvec{\sigma_{\text{LB}}}$ helps to stabilize learning when generating curricula for narrow target distributions. This is because the Gaussian distributions have a tendency to quickly reduce the variance of the sampling distribution in this case. In combination with the KL divergence constraint on subsequent context distributions, this slows down progression towards the target distribution. Although we could enforce aforementioned lower bound via constraints on the distribution $p(\svec{c} \vert \cvec{\nu})$, we simply clip the standard deviation until the KL divergence w.r.t. the target distribution $\mu(\svec{c})$ falls below a certain threshold $D_{\text{KL}_{\text{LB}}}$. This threshold was chosen such that the distribution with the clipped standard deviation roughly ``contains'' the mean of target distribution within its standard deviation interval. The specific values of $D_{\text{KL}_{\text{LB}}}$ and $\cvec{\sigma_{\text{LB}}}$ are listed for the individual experiments.

\subsection{Episodic Setting}

\begin{table}[b]
	\begin{center}
		\begin{small}
			\begin{sc}
				\begin{tabular}{lccccccccr}
					\toprule
					& $\epsilon$ & $n_{\text{samples}}$ & Buffer Size & $\zeta$ & $K_{\alpha}$ & $\cvec{\sigma_{\text{LB}}}$ & $D_{\text{KL}_{\text{LB}}}$ \\
					\midrule
					Gate ``Global'' & $0.25$ & $100$ & $10$ & $0.002$ & $140$ & - & - \\
					Gate ``Precision'' & $0.4$ & $100$ & $10$ & $0.02$ & $140$ & - & - \\
					Reacher & $0.5$ & $50$ & $10$ & $0.15$ & $90$ & $[0.005\ 0.005]$ & $20$ \\
					Ball-in-a-cup & $0.35$ & $16$ & $5$ & $3.0$ & $15$ & $0.1$ & $200$ \\
					\bottomrule
				\end{tabular}
			\end{sc}
		\end{small}
	\end{center}
	\caption[Experimental Details]{Important parameters of SPRL and C-REPS in the conducted experiments. The meaning of the symbols correspond to those presented in the  algorithm from the main text and introduced in this appendix.}
	\label{table:exp-det}
	\vskip -0.1in
\end{table}
For the visualization of the success rate as well as the computation of the success indicator for the GoalGAN algorithm, the following definition is used: An experiment is considered successful, if the distance between final- and desired state ($\svec{s}_f$ and $\svec{s}_g$) is less than a given threshold $\tau$
\begin{align*}
	\text{Success}\left(\cvec{\theta}, \svec{c}\right) = \begin{cases}
		1,\ \text{if}\  \norm{\svec{s}_f\left(\cvec{\theta}\right) - \svec{s}_g\left(\svec{c}\right)}_{2} < \tau, \\
		0,\ \text{else}.
	\end{cases}
\end{align*}
For the Gate and reacher environment, the threshold is fixed to $0.05$, while for the ball-in-a-cup environment, the threshold depends on the scale of the cup and the goal is set to be the center of the bottom plate of the cup. \\
The policies are chosen to be conditional Gaussian distributions $\mathcal{N}(\cvec{\theta} \vert \cvec{A} \phi(\svec{c}), \cvec{\Sigma}_{\cvec{\theta}})$, where $\phi(\svec{c})$ is a feature function. SPRL and C-REPS both use linear policy features in all environments. \\
In the reacher and the ball-in-a-cup environment, the parameters $\cvec{\theta}$ encode a feed-forward policy by weighting several Gaussian basis functions over time
\begin{align*}
\cvec{u}_i\left(\cvec{\theta}\right) = \cvec{\theta}^T \cvec{\psi}\left(t_i\right), \quad \cvec{\psi}_j\left(t_i\right) = \frac{b_j\left(t_i\right)}{\sum_{l=1}^L b_l\left(t_i\right)}, \quad b_j\left(t_i\right) = \exp\left(\frac{\left(t_i - c_j\right)^2}{2L}\right),
\end{align*}
where the centers $c_j$ and length $L$ of the basis functions are chosen individually for the experiments. With that, the policy represents a Probabilistic Movement Primitive \citep{paraschos-promps}, whose mean and covariance matrix are progressively shaped by the learning algorithm to encode movements with high reward.
\\
In order to increase the robustness of SPRL and C-REPS while reducing the sample complexity, an experience buffer storing samples of recent iterations is used. The size of this buffer dictates the number of past iterations, whose samples are kept. Hence, in every iteration, C-REPS and SPRL work with $N_{\text{SAMPLES}} \times \text{BUFFER SIZE}$ samples, from which only $N_{\text{SAMPLES}}$ are generated by the policy of the current iteration.
\\
As the employed CMA-ES implementation only allows to specify one initial variance for all dimensions of the search distribution, this variance is set to the maximum of the variances contained in the initial covariance matrices used by SPRL and C-REPS.
\\
For the GoalGAN algorithm, the percentage of samples that are drawn from the buffer containing already solved tasks is fixed to $20\%$. The noise added to the samples of the GAN $\delta_{\text{NOISE}}$ and the number of iterations that pass between the training of the GAN $n_{\text{ROLLOUT}_{\text{GG}}}$ are chosen individually for the experiments.
\\
The SAGG-RIAC algorithm requires, besides the probabilities for the sampling modes which are kept as in the original paper, two hyperparameters to be chosen: The maximum number of samples to keep in each region $n_{\text{GOALS}}$ as well as the maximum number of recent samples for the competence computation $n_{\text{HIST}}$. 
\\
Tables \ref{table:exp-det} and \ref{table:exp-det-2} show the aforementioned hyperparameters of C-REPS, SPRL, GoalGAN and SAGG-RIAC for the different environments.

\begin{table}[t]
	\begin{center}
		\begin{small}
			\begin{sc}
				\begin{tabular}{lccccccr}
					\toprule
					& $\delta_{\text{NOISE}}$ & $n_{\text{ROLLOUT}_{\text{GG}}}$ & $n_{\text{GOALS}}$ & $n_{\text{HIST}}$ \\
					\midrule
					Gate ``Global'' & $0.05$ & $5$ & $100$ & $500$ \\
					Gate ``Precision'' & $0.05$ & $5$ & $100$ & $200$ \\
					Reacher & $0.1$ & $5$ & $80$ & $300$ \\
					Ball-in-a-cup & $0.05$ & $3$ & $50$ & $120$  \\
					\bottomrule
				\end{tabular}
			\end{sc}
		\end{small}
	\end{center}
	\caption[Experimental Details]{Important parameters of GoalGAN and SAGG-RIAC in the conducted experiments. The meaning of the symbols correspond to those introduced in this appendix.}
	\label{table:exp-det-2}
	\vskip -0.1in
\end{table}

\subsubsection{Point-Mass Experiment}

\begin{figure}
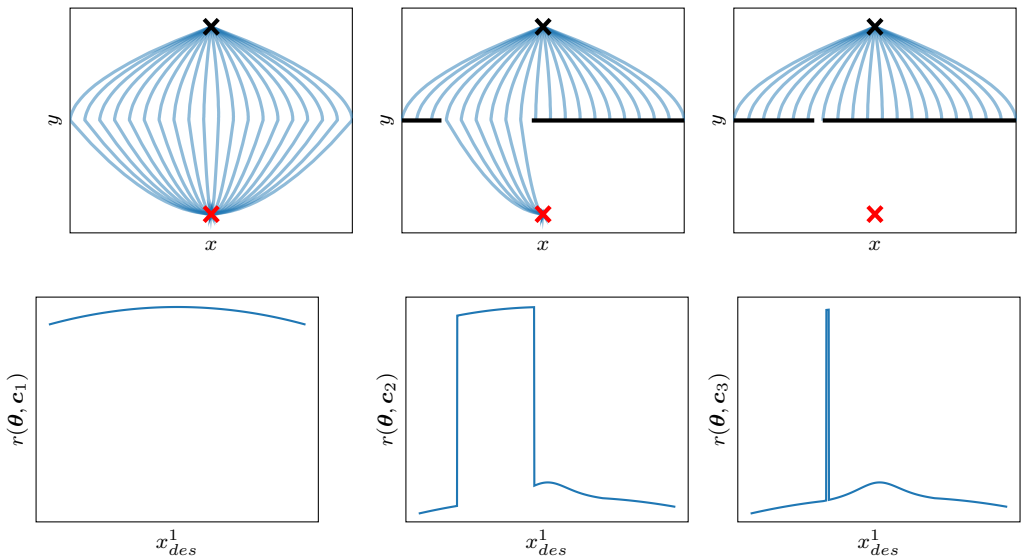

\centering
\begin{tikzpicture}
\node (leftnode) at (0, 0) {\input{img/gate-examples-20.tex}};
\node [below left = 0.25 and -3.96 of leftnode] {\input{img/gate-cost-function-20-3.tex}};

\node[right = 0.0 of leftnode] (middlenode) {\input{img/gate-examples-3.tex}};
\node [below = 0.25 of middlenode] {\input{img/gate-cost-function-3-3.tex}};

\node[right = 0.0 of middlenode] (rightnode) {\input{img/gate-examples-0.1.tex}};
\node [below = 0.25 of rightnode] {\input{img/gate-cost-function-0.1-3.tex}};
\end{tikzpicture}
\caption[Test]{The columns show visualizations of the point-mass trajectories (upper plots) as well as the obtained rewards (lower plots) in the point-mass task, when the desired position of the first controller is varied while all other parameters are kept fixed such that a stable control law is obtained. In every column, the gate is positioned at $x=4.0$ while the size of it varies from $20$ (left), over $3$ (middle) to $0.1$ (right).}
\label{fig:gate-reward-vis}
\end{figure}

The linear system that describes the behavior of the point-mass is given by
\begin{equation*}
	\begin{bmatrix}
		\dot{x} \\
		\dot{y}
	\end{bmatrix} = \begin{bmatrix}
		5 \\
		-1
	\end{bmatrix} + \cvec{u} + \cvec{\delta},\quad \cvec{\delta} \sim \mathcal{N}\left(\cvec{0}, 2.5 \times 10^{-3} \cvec{I} \right).
\end{equation*}
The point-mass is controlled by two linear controllers
\begin{align*}
	\text{C}_i\left(x, y\right) = \cvec{K}_i \begin{bmatrix}
		x_i - x \\
		y_i - y
	\end{bmatrix} + \cvec{k}_i,\ i\in \left[1, 2\right], \ \ \cvec{K}_i \in \mathbb{R}^{2 \times 2},\ \cvec{k}_i \in \mathbb{R}^2,\ x_i, y_i \in \mathbb{R},
\end{align*}
where $x$ is the $x$-position of the point-mass and $y$ its position on the $y$-axis. The episode reward exponentially decays with the final distance to the goal. In initial iterations of the algorithm, the sampled controller parameters sometimes make the control law unstable, leading to very large penalties due to large actions and hence to numerical instabilities in SPRL and C-REPS because of very large negative rewards. Because of this, the reward is clipped to always be above $0$.
\\
Table~\ref{table:exp-det} shows that a large number of samples per iteration for both the ``global'' and ``precision'' setting are used. This is purposefully done to keep the influence of the sample size on the algorithm performance as low as possible, as both of these settings serve as a first conceptual benchmark of our algorithm.
\\
Figure~\ref{fig:gate-reward-vis} helps in understanding, why SPRL drastically improves upon C-REPS especially in the ``precision'' setting, even with this large amount of samples. For narrow gates, the reward function has a local maximum which tends to attract both C-REPS and CMA-ES, as the chance of sampling a reward close to the true maximum is very unlikely. By first training on contexts in which the global maximum is more likely to be observed and only gradually moving towards the desired contexts, SPRL avoids this sub-optimal solution.

\subsubsection{Reacher Experiment}

In the reacher experiment, the ProMP encoded by the policy $\pi$ has $20$ basis functions of width $L=0.03$. The centers are evenly spread in the interval $[-0.2, 1.2]$ and the time interval of the movement is normalized to lie in the interval $[0, 1]$ when computing the activations of the basis functions. Since the robot can only move within the $xy$-plane, $\cvec{\theta}$ is a $40$-dimensional vector. As in the previous experiment, the episode reward decays exponentially with the final distance to the goal. As we can see in Table~\ref{table:exp-det}, the number of samples in each iteration was decreased to $50$, which in combination with the increased dimensionality of $\cvec{\theta}$ makes the task more challenging.
\\
As in the step-based setting, the PPO results are obtained using the version from the \texttt{Stable Baselines} library \citep{stable-baselines}. A step-based version of the reacher experiment is used, in which the reward function is given by
$$
r(\cvec{s}, \cvec{a}) = \exp \left(-2.5 \sqrt{(x - x_g)^2 + (y - y_g)^2}\right),
$$
where $\cvec{s} = (x\ \dot{x}\ y\ \dot{y})$ is the position and velocity of the end-effector, $\cvec{a} = (a_x\ a_y)$ the desired displacement of the end-effector (just as in the regular reacher task from the OpenAI Gym simulation environment) and $x_g$ and $y_g$ is the $x-$ and $y-$ position of the goal. When an obstacle is touched, the agent is reset to the initial position. This setup led to the best performance of PPO, while resembling the structure of the episodic learning task used by the other algorithms (a version in which the episode ends as soon as an obstacle is touched led to a lower performance of PPO).
\\
To ensure that the poor performance of PPO is not caused by an inadequate choice of hyperparameters, PPO was run on an easy version of the task in which the two obstacle sizes were set to $0.01$, where it encountered no problems in solving the task.
\\
Every iteration of PPO uses $3600$ environment steps, which corresponds to $24$ trajectory executions in the episodic setting. PPO uses an entropy coefficient of $10^{-3}$, $\gamma=0.999$ and $\lambda=1$. The neural network that learns the value function as well as the policy has two dense hidden layers with $164$ neurons and $\tanh$ activation functions. The number of minibatches is set to $5$ while the number of optimization epochs is set to $15$. The standard deviation in each action dimension is initialized to $1$, giving the algorithm enough initial variance, as the actions are clipped to the interval $[-1, 1]$ before being applied to the robot.

\subsubsection{Ball-in-a-Cup Experiment}

For the ball-in-a-cup environment, the $9$ basis functions of the ProMP are spread over the interval $[-0.01, 1.01]$ and have width $L=0.0035$. Again, the time interval of the movement is normalized to lie in the interval $[0,1]$ when computing the basis function activations. The ProMP encodes the offset of the desired position from the initial position.
By setting the first and last two basis functions to $0$ in each of the three dimensions, the movement always starts in the initial position and returns to it after the movement execution. All in all, $\theta$ is a $15$-dimensional vector. The reward function is defined as
\begin{align*}
r(\cvec{\theta}, \svec{c}) & = \begin{cases}%
1 - 0.07 \cvec{\theta}^T \cvec{\theta}&,\ \text{if successful} \\
0&,\ \text{else}%
\end{cases}.
\end{align*}
This encodes a preference over movements that deviate as little as possible from the initial position while still solving the task.
\\
Looking back at Table~\ref{table:exp-det}, the value of $\zeta$ stands out, as it is significantly higher than in the other experiments. We suppose that such a large value of $\zeta$ is needed because of the shape of the reward function, which creates a large drop in reward if the policy is sub-optimal. Because of this, the incentive required to encourage the algorithm to shift probability mass towards contexts in which the current policy is sub-optimal needs to be significantly higher than in the other experiments.
\\
After learning the movements in simulation, the successful runs were executed on the real robot. Due to simulation bias, just replaying the trajectories did not work satisfyingly. At this stage, we could have increased the variance of the movement primitive and re-trained on the real robot. As sim-to-real transfer is, however, not the focus of this paper, we decided to manually adjust the execution speed of the movement primitive by a few percent, which yielded the desired result.

\subsection{Step-Based Setting}
\label{ap:exp-det}

\begin{table}[t]
	\begin{center}
	\resizebox{\columnwidth}{!}{
		\begin{small}
			\begin{sc}
				\begin{tabular}{lccccccr}
					\toprule
					& $K_{\alpha}$ & $\zeta$ & $K_{\text{offset}}$ & $V_{\text{LB}}$ & $n_{\text{step}}$ & $\cvec{\sigma}_{\text{LB}}$ & $D_{\text{KL}_{LB}}$ \\
					\midrule
					Point-Mass (TRPO) & $20$ & $1.6$ & $5$ & $3.5$ & $2048$ & $[0.2\ \ 0.1875\ \ 0.1]$ & $8000$ \\
					Point-Mass (PPO) & $10$ & $1.6$ & $5$ & $3.5$ & $2048$ & $[0.2\ \ 0.1875\ \ 0.1]$ & $8000$ \\
					Point-Mass (SAC) & $25$ & $1.1$ & $5$ & $3.5$ & $2048$ & $[0.2\ \ 0.1875\ \ 0.1]$ & $8000$ \\
					Ant (PPO) & $15$ & $1.6$ & $10$ & $600$ & $81920$ & $[1\ \ 0.5]$ & $11000$\\
					Ball-Catching (TRPO) & $70$ & $0.4$ & $5$ & $42.5$ & $5000$ & - & -\\
					Ball-Catching* (TRPO) & $0$ & $0.425$ & $5$ & $42.5$ & $5000$ & - & -\\
					Ball-Catching (PPO) & $50$ & $0.45$ & $5$ & $42.5$ & $5000$ & - & -\\
					Ball-Catching* (PPO) & $0$ & $0.45$ & $5$ & $42.5$ & $5000$ & - & -\\
					Ball-Catching (SAC) & $60$ & $0.6$ & $5$ & $25$ & $5000$ & - & -\\
					Ball-Catching* (SAC) & $0$ & $0.6$ & $5$ & $25$ & $5000$ & - & -\\
					\bottomrule
				\end{tabular}
			\end{sc}
		\end{small}}
	\end{center}
	\caption[Experimental Details]{Hyperparameters for the SPDL algorithm per environment and RL algorithm. The asterisks in the table mark the ball-catching experiments with an initialized context distribution.}
	\label{table:spdl-det}
	\vskip -0.1in
\end{table}

The parameters of SPDL for different environments and RL algorithms are shown in Table \ref{table:spdl-det}. Opposed to the sketched algorithm in the main paper, we specify the number of steps $n_{\text{STEP}}$ in the environment between context distribution updates instead of the number of trajectory rollouts. The additional parameter $K_{\text{OFFSET}}$ describes the number of RL algorithm iterations that take place before SPDL is allowed the change the context distribution. We used this in order to improve the estimate regarding task difficulty, as for completely random policies, task difficulty is not as apparent as for slightly more structured ones. This procedure corresponds to providing parameters of a minimally pre-trained policy as $\cvec{\omega}_0$ in the algorithm sketched in the main paper. We selected the best $\zeta$ for every RL algorithm by a simple grid-search in an interval around a reasonably working parameter that was found by simple trial and error. For the point-mass environment, we only tuned the hyperparameters for SPDL in the experiment with a three-dimensional context space and reused them for the two-dimensional context space.  \\
Since the step-based algorithm makes use of the value function estimated by the individual RL algorithms, particular regularizations of RL algorithms can affect the curriculum. SAC, for example, estimates a ``biased'' value function due to the employed entropy regularization. This bias caused problems for our algorithm when working with the $\alpha$-heuristic based on $V_{\text{LB}}$. Because of this, we simply replace the value estimates for the contexts by their sample return when working with SAC and $V_{\text{LB}}$. This is an easy way to obtain an unbiased, yet noisier estimate of the value of a context. Furthermore, the general advantage estimation (GAE) employed by TRPO and PPO can introduce bias in the value function estimates as well. For the ant environment, we realized that this bias is particularly large due to the long time horizons. Consequently, we again made use of the sample returns to estimate the value functions for the sampled contexts. In all other cases and environments, we used the value functions estimated by the RL algorithms. \\
For ALP-GMM we tuned the percentage of random samples drawn from the context space $p_{\text{RAND}}$, the number of policy rollouts between the update of the context distribution $n_{\text{ROLLOUT}}$ as well as the maximum buffer size of past trajectories to keep $s_{\text{BUFFER}}$. For each environment and algorithm, we did a grid-search over 
\begin{align*}
(p_{\text{RAND}}, n_{\text{ROLLOUT}}, s_{\text{BUFFER}}) \in \{0.1, 0.2, 0.3\} \times \{25, 50, 100, 200\} \times \{500, 1000, 2000\}.
\end{align*}
\\
For GoalGAN we tuned the amount of random noise that is added on top of each sample $\delta_{\text{NOISE}}$, the number of policy rollouts between the update of the context distribution $n_{\text{ROLLOUT}}$ as well as the percentage of samples drawn from the success buffer $p_{\text{SUCCESS}}$. For each environment and algorithm, we did a grid-search over
\begin{align*}
(\delta_{\text{NOISE}}, n_{\text{ROLLOUT}}, p_{\text{SUCCESS}}) \in \{0.025, 0.05, 0.1\} \times \{25, 50, 100, 200\} \times \{0.1, 0.2, 0.3\}.
\end{align*}
The results of the hyperparameter optimization for GoalGAN and ALP-GMM are shown in Table \ref{table:cl-hps}.
\\
Since for all environments, both initial- and target distribution are Gaussians with independent noise in each dimension, we specify them in Table \ref{table:spdl-det-2} by providing their mean $\cvec{\mu}$ and the vector of standard deviations for each dimension $\cvec{\delta}$. When sampling from a Gaussian, the resulting context is clipped to stay in the defined context space.
\\
The experiments were conducted on a computer with an AMD Ryzen 9 3900X 12-Core Processor, an Nvidia RTX 2080 graphics card and 64GB of RAM.

\begin{table}[t!]
	\begin{center}
	\resizebox{\columnwidth}{!}{
		\begin{small}
			\begin{sc}
				\begin{tabular}{lccccccr}
					\toprule
					& $p_{\text{RAND}}$ & $n_{\text{ROLLOUT}_{\text{AG}}}$ & $s_{\text{BUFFER}}$ & $\delta_{\text{NOISE}}$ & $n_{\text{ROLLOUT}_{\text{GG}}}$ & $p_{\text{SUCCESS}}$ \\
					\midrule
					Point-Mass 3D (TRPO) & $0.1$ & $100$ & $1000$ & $0.05$ & $200$ & $0.2$ \\
					Point-Mass 3D (PPO) & $0.1$ & $100$ & $500$ & $0.025$ & $200$ & $0.1$ \\
					Point-Mass 3D (SAC) & $0.1$ & $200$ & $1000$ & $0.1$ & $100$ & $0.1$ \\
					Point-Mass 2D (TRPO) & $0.3$ & $100$ & $500$ & $0.1$ & $200$ & $0.2$ \\
					Point-Mass 2D (PPO) & $0.2$ & $100$ & $500$ & $0.1$ & $200$ & $0.3$ \\
					Point-Mass 2D (SAC) & $0.2$ & $200$ & $1000$ & $0.025$ & $50$ & $0.2$ \\
					Ant (PPO) & $0.1$ & $50$ & $500$ & $0.05$ & $125$ & $0.2$\\
					Ball-Catching (TRPO) & $0.2$ & $200$ & $2000$ & $0.1$ & $200$ & $0.3$\\
					Ball-Catching (PPO) & $0.3$ & $200$ & $2000$ & $0.1$ & $200$ & $0.3$\\
					Ball-Catching (SAC) & $0.3$ & $200$ & $1000$ & $0.1$ & $200$ & $0.3$\\
					\bottomrule
				\end{tabular}
			\end{sc}
		\end{small}}
	\end{center}
	\caption[Experimental Details]{Hyperparameters for the ALP-GMM and GoalGAN algorithm per environment and RL algorithm. The abbreviation AG is used for ALP-GMM, while GG stands for GoalGAN.}
	\label{table:cl-hps}
	\vskip -0.1in
\end{table}

\begin{table*}[t]
	\begin{center}
		\resizebox{\columnwidth}{!}{
		\begin{small}
			\begin{sc}
				\begin{tabular}{lccccr}
					\toprule
					& $\cvec{\mu}_{\text{init}}$ & $\cvec{\delta}_{\text{init}}$ & $\cvec{\mu}_{\text{target}}$ & $\cvec{\delta}_{\text{target}}$ \\
					\midrule
					Point-Mass & $[0\ \ 4.25\ \ 2]$ & $[2\ \ 1.875\ \ 1]$ & $[2.5\ \ 0.5\ \ 0]$ & $[0.004\ \ 0.00375\ \ 0.002]$ \\
					Ant & $[0\ \ 8]$ & $[3.2\ \ 1.6]$ & $[-8\ \ 3]$ & $[0.01\ \ 0.005]$ \\
					Ball-Catching & $[0.68\ \ 0.9\ \ 0.85]$ & $[0.03\ \ 0.03\ \ 0.3]$ & $[1.06\ \ 0.85\ \ 2.375]$ & $[0.8\ \ 0.38\ \ 1]$ \\
					\bottomrule
				\end{tabular}
			\end{sc}
		\end{small}}
	\end{center}
	\caption[Experimental Details]{Mean and standard deviation of target and initial distributions per environment.}
	\label{table:spdl-det-2}
	\vskip -0.1in
\end{table*}

\subsubsection{Point-Mass Environment}

\begin{figure}[t]
\begin{subfigure}{0.2\textwidth}
	\centering
	\includegraphics{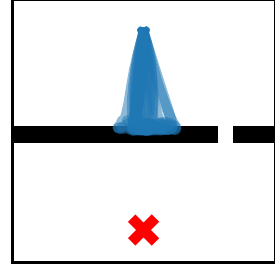}
	\caption{Default}
    \label{fig:doc1}
\end{subfigure}\begin{subfigure}{0.2\textwidth}
	\centering
	\includegraphics{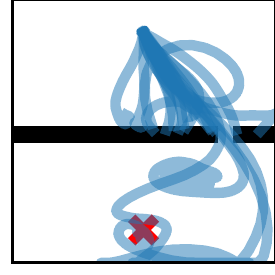}
	\caption{Random}
\end{subfigure}\begin{subfigure}{0.2\textwidth}
	\centering
	\includegraphics{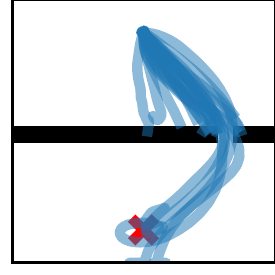}
	\caption{ALP-GMM}
\end{subfigure}\begin{subfigure}{0.2\textwidth}
	\centering
	\includegraphics{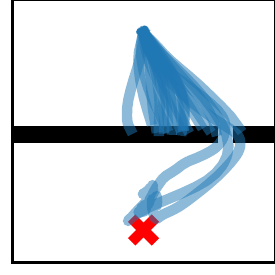}
	\caption{GoalGAN}
\end{subfigure}\begin{subfigure}{0.2\textwidth}
	\centering
	\includegraphics{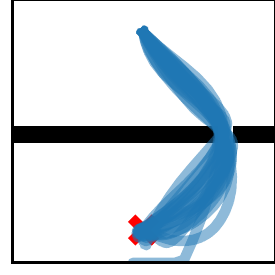}
	\caption{SPDL}
\end{subfigure}
\vspace{10pt}
\begin{subfigure}{0.2\textwidth}
	\centering
	\includegraphics{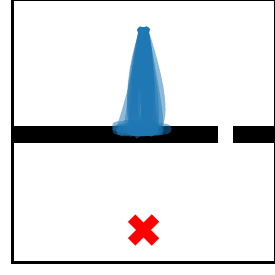}
	\caption{Default}
    \label{fig:doc1}
\end{subfigure}\begin{subfigure}{0.2\textwidth}
	\centering
	\includegraphics{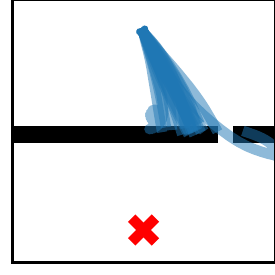}
	\caption{Random}
\end{subfigure}\begin{subfigure}{0.2\textwidth}
	\centering
	\includegraphics{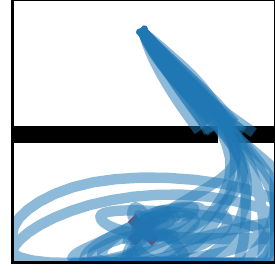}
	\caption{ALP-GMM}
\end{subfigure}\begin{subfigure}{0.2\textwidth}
	\centering
	\includegraphics{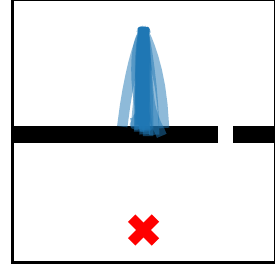}
	\caption{GoalGAN}
\end{subfigure}\begin{subfigure}{0.2\textwidth}
	\centering
	\includegraphics{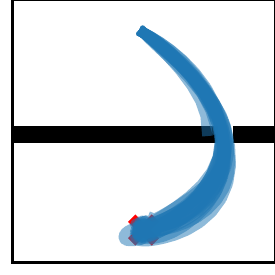}
	\caption{SPDL}
\end{subfigure}
\vspace{10pt}
\begin{subfigure}{0.2\textwidth}
	\centering
	\includegraphics{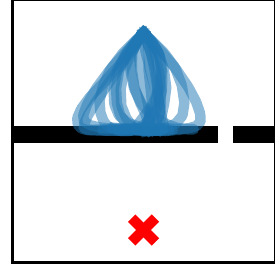}
	\caption{Default}
    \label{fig:doc1}
\end{subfigure}\begin{subfigure}{0.2\textwidth}
	\centering
	\includegraphics{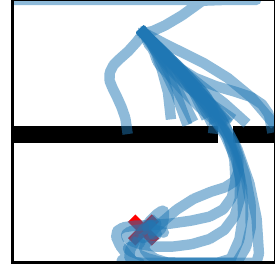}
	\caption{Random}
\end{subfigure}\begin{subfigure}{0.2\textwidth}
	\centering
	\includegraphics{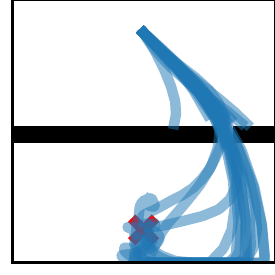}
	\caption{ALP-GMM}
\end{subfigure}\begin{subfigure}{0.2\textwidth}
	\centering
	\includegraphics{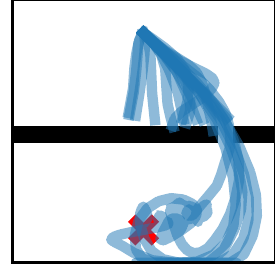}
	\caption{GoalGAN}
\end{subfigure}\begin{subfigure}{0.2\textwidth}
	\centering
	\includegraphics{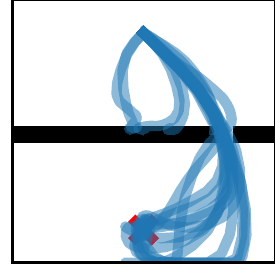}
	\caption{SPDL}
\end{subfigure}
\caption{Visualizations of policy rollouts in the point-mass environment (three context dimensions) with policies learned using different curricula and RL algorithms. Each rollout was generated using a policy learned with a different seed. The first row shows results for TRPO, the second for PPO and the third shows results for SAC.}
\label{fig:point_mass_plot}
\vspace{-10pt}
\end{figure}

The state of this environment is comprised of the position and velocity of the point-mass $\cvec{s} = [x\ \dot{x}\ y\ \dot{y}]$. The actions correspond to the force applied in x- and y-dimension ${\cvec{a} = [F_x\ F_y]}$. The context encodes position and width of the gate as well as the dynamic friction coefficient of the ground on which the point-mass slides ${\svec{c} = [p_g\ w_g\ \mu_k] \in [-4, 4] \times [0.5, 8] \times [0, 4] \subset \mathbb{R}^3}$. The dynamics of the system are defined by
\begin{align*}
\begin{pmatrix}
\dot{x}\\
\ddot{x}\\
\dot{y}\\
\ddot{y}
\end{pmatrix} = \begin{pmatrix}
0 & 1 & 0 & 0 \\
0 & -\mu_k & 0 & 0 \\
0 & 0 & 0 & 1 \\
0 & 0 & 0 & -\mu_k
\end{pmatrix} \cvec{s} + \begin{pmatrix}
0 & 0 \\ 
1 & 0 \\
0 & 0 \\ 
0 & 1
\end{pmatrix} \cvec{a}.
\end{align*}
The $x$- and $y$- position of the point-mass is enforced to stay within the space $[-4, 4] \times [-4, 4]$. The gate is located at position $[p_g\ 0]$. If the agent crosses the line $y = 0$, we check whether its $x$-position is within the interval ${[p_g - 0.5 w_g, p_g + 0.5 w_g]}$. If this is not the case, we stop the episode as the agent has crashed into the wall. Each episode is terminated after a maximum of $100$ steps. The reward function is given by
\begin{align*}
r(\cvec{s}, \cvec{a})  = \exp\left( -0.6 \| \cvec{o} - [x\ y] \|_2 \right),
\end{align*}
where $\cvec{o}=[0\ \ {-}3]$, $\|\cdot\|_2$ is the L2-Norm. The agent is always initialized at state $\cvec{s}_0 = [0\ 0\ 3\ 0]$.
\\
For all RL algorithms, we use a discount factor of $\gamma = 0.95$ and represent policy and value function by networks using two hidden layers with $64$ neurons and tanh activations. For TRPO and PPO, we take $2048$ steps in the environment between policy updates.
\\
For TRPO we set the GAE parameter $\lambda = 0.99$, leaving all other parameters to their implementation defaults. 
\\
For PPO we use GAE parameter $\lambda = 0.99$, an entropy coefficient of $0$ and disable the clipping of the value function objective. The number of optimization epochs is set to $8$ and we use $32$ mini-batches. All other parameters are left to their implementation defaults.
\\
For SAC, we use an experience-buffer of $10000$ samples, starting learning after $500$ steps. We use the soft Q-Updates and update the policy every $5$ environment steps. All other parameters were left at their implementation defaults.
\\
For SPRL, we use $K_{\alpha}=40$, $K_{\text{OFFSET}} = 0$, $\zeta=2.0$ for the 3D- and $\zeta=1.5$ and 2D case. We use the same values for $\sigma_{\text{LB}}$ and $D_{\text{KL}_{\text{LB}}}$ as for SPDL (Table \ref{table:spdl-det}). Between updates of the episodic policy, we do $25$ policy rollouts and keep a buffer containing rollouts from the past $10$ iterations, resulting in $250$ samples for policy- and context distribution update. The linear policy over network weights is initialized to a zero-mean Gaussian with unit variance. We use polynomial features up to degree two to approximate the value function. For the allowed KL divergence, we observed best results when using $\epsilon = 0.5$ for the weight computation of the samples, but using a lower value of $\epsilon = 0.2$ when fitting the parametric policy to these samples. We suppose that the higher value of $\epsilon$ during weight computation counteracts the effect of the buffer containing policy samples from earlier iterations.
\\
Looking at Figure \ref{fig:point_mass_plot}, we can see that depending on the learning algorithm, ALP-GMM, GoalGAN and a random curriculum allowed to learn policies that sometimes are able to pass the gate. However, in other cases, the policies crashed the point-mass into the wall. Opposed to this, directly training on the target task led to policies that learned to steer the point-mass very close to the wall without crashing (which is unfortunately hard to see in the plot). Reinvestigating the above reward function, this explains the lower reward of GoalGAN compared to directly learning on the target task, as a crash prevents the agent from accumulating positive rewards over time. SPDL learned more reliable and directed policies across all learning algorithms.

\subsubsection{Ant Environment}

\begin{figure}[b!]
\centering
\begin{subfigure}{0.33\textwidth}
	\centering
	\includegraphics{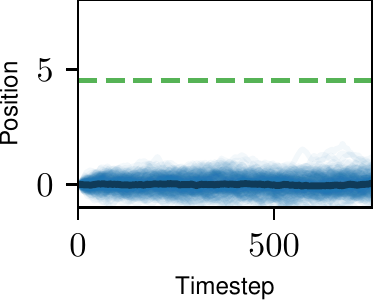}
	\caption{Default}
    \label{fig:doc1}
\end{subfigure}\begin{subfigure}{0.33\textwidth}
	\centering
	\includegraphics{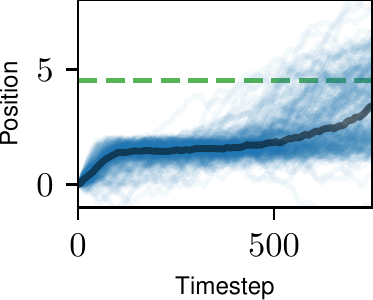}
	\caption{Random}
\end{subfigure}\begin{subfigure}{0.33\textwidth}
	\centering
	\includegraphics{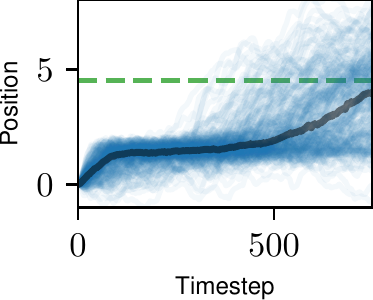}
	\caption{ALP-GMM}
\end{subfigure}
\begin{subfigure}{0.33\textwidth}
	\vspace{10pt}
	\centering
	\includegraphics{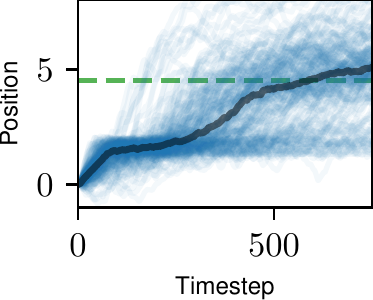}
	\caption{GoalGAN}
\end{subfigure}\hspace{10pt}\begin{subfigure}{0.33\textwidth}
	\vspace{10pt}
	\centering
	\includegraphics{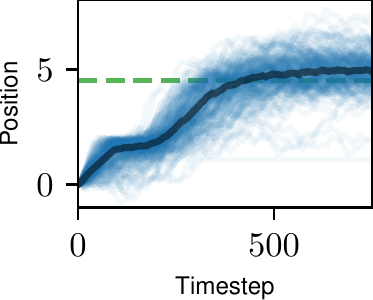}
	\caption{SPDL}
\end{subfigure}
\caption{Visualizations of the $x$-position during policy rollouts in the ant environment with policies learned using different curricula. The blue lines correspond to $200$ individual trajectories and the thick black line shows the median over these individual trajectories. The trajectories were generated from $20$ algorithms runs, were each final policy was used to generate $10$ trajectories.}
\label{fig:ant_plot}
\end{figure}

As mentioned in the main paper, we simulate the ant using the Isaac Gym simulator \citep{isaac-gym}. This allows to speed up training time by parallelizing the simulation of policy rollouts on the graphics card. Since the Stable-Baselines implementation of TRPO and SAC do not support the use of vectorized environments, it is hard to combine Isaac Gym with these algorithms. Because of this reason, we decided not to run experiments with TRPO and SAC in the ant environment.
\\
The state $\cvec{s} \in \mathbb{R}^{29}$ is defined to be the 3D-position of the ant's body, its angular and linear velocity as well as positions and velocities of the $8$ joints of the ant. An action $\cvec{a} \in \mathbb{R}^8$ is defined by the $8$ torques that are applied to the ant's joints.
\\
The context $\svec{c} = [p_g\ w_g] \in [-10, 10] \times [3, 13] \subset \mathbb{R}^2$ defines, just as in the point-mass environment, the position and width of the gate that the ant needs to pass.
\\
The reward function of the environment is computed based on the $x$-position of the ant's center of mass $c_x$ in the following way
\begin{align*}
r(\cvec{s}, \cvec{a}) = 1 + 5 \exp\left(-0.5 \min(0, c_x - 4.5)^2\right) - 0.3 \| \cvec{a} \|_2^2.
\end{align*}
The constant $1$ term was taken from the OpenAI Gym implementation to encourage the survival of the ant \citep{brockman2016openai}. Compared to the OpenAI Gym environment, we set the armature value of the joints from $1$ to $0$ and also decrease the maximum torque from $150\text{Nm}$ to $20\text{Nm}$, since the values from OpenAI Gym resulted in unrealistic movement behavior in combination with Isaac Gym. Nonetheless, these changes did not result in a qualitative change in the algorithm performances.
\\
With the wall being located at position $x{=}3$, the agent needs to pass it in order to obtain the full environment reward by ensuring that $c_x >= 4.5$.
\\
The policy and value function are represented by neural networks with two hidden layers of $64$ neurons each and $\tanh$ activation functions. We use a discount factor $\gamma = 0.995$ for all algorithms, which can be explained due to the long time horizons of $750$ steps. We take $81920$ steps in the environment between a policy update. This was significantly sped-up by the use of the Isaac Gym simulator, which allowed to simulate $40$ environments in parallel on a single GPU.
\\
For PPO, we use an entropy coefficient of $0$ and disable the clipping of the value function objective. All other parameters are left to their implementation defaults. We disable the entropy coefficient as we observed that for the ant environment, PPO still tends to keep around $10-15\%$ of its initial additive noise even during late iterations.  
\\
Investigating Figure \ref{fig:ant_plot}, we see that both SPDL and GoalGAN learn policies that allow to pass the gate. However, the policies learned with SPDL seem to be more reliable compared to the ones learned with GoalGAN. As mentioned in the main paper, ALP-GMM and a random curriculum also learn policies that navigate the ant towards the goal in order to pass it. However, the behavior is less directed and less reliable. Interestingly, directly learning on the target task results in a policy that tends to not move in order to avoid action penalties. Looking at the main paper, we see that this results in a similar reward compared to the inefficient policies learned with ALP-GMM and a random curriculum.

\subsubsection{Ball-Catching Environment}

In the final environment, the robot is controlled in joint space via the desired position for $5$ of the $7$ joints. We only control a subspace of all available joints, since it is not necessary for the robot to leave the ''catching'' plane (defined by $x=0$) that is intersected by each ball. The actions $\cvec{a} \in \mathbb{R}^5$ are defined as the displacement of the current desired joint position. The state $\cvec{s} \in \mathbb{R}^{21}$ consists of the positions and velocities of the controlled joints, their current desired positions, the current three-dimensional ball position and its linear velocity.

\begin{figure}[t]
\begin{subfigure}{\textwidth}
    \centering
    \includegraphics[height=1.5in]{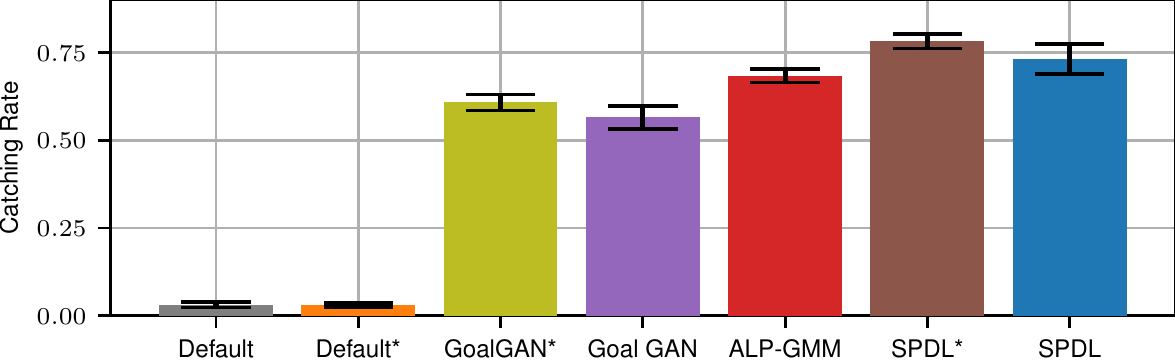}
    \caption{SAC}
    \label{fig:doc1}
    \end{subfigure}

    \bigskip
    \begin{subfigure}{\textwidth}
    \centering
    \includegraphics[height=1.5in]{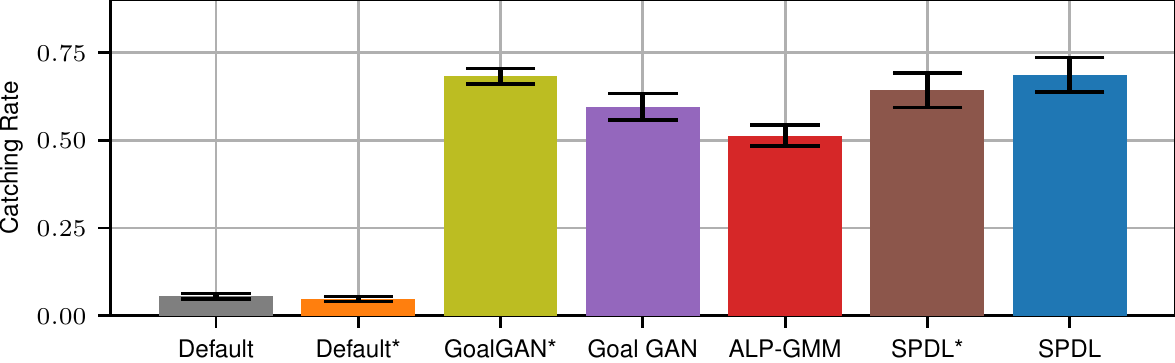}
    \caption{TRPO}
    \label{fig:doc2}
    \end{subfigure}

    \bigskip
    \begin{subfigure}{\textwidth}
    \centering
    \includegraphics[height=1.5in]{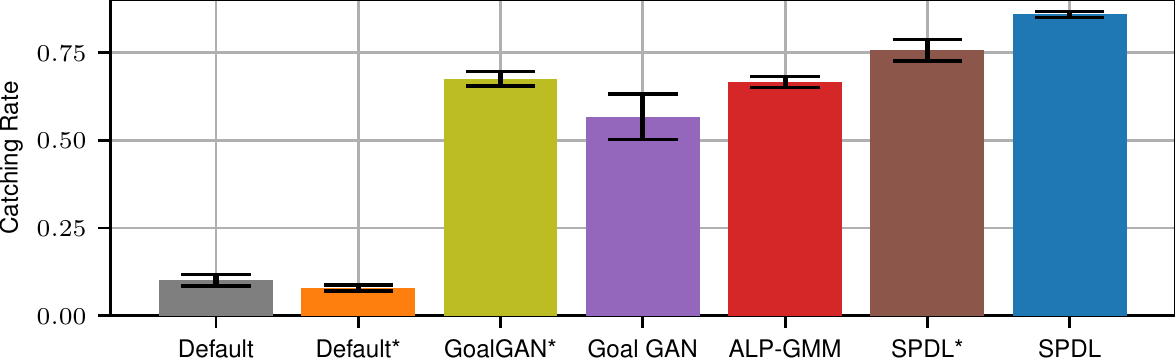}
    \caption{PPO}
    \label{fig:doc3}
    \end{subfigure}
\caption{Mean catching rate of the final policies learned with different curricula and RL algorithms on the ball catching environment. The mean is computed from $20$ algorithm runs with different seeds. For each run, the success rate is computed from $200$ ball-throws. The bars visualize the estimated standard error.}
\label{fig:ball_catching}
\end{figure}

As previously mentioned, the reward function is sparse,
\begin{align*}
r(\cvec{s}, \cvec{a}) = 0.275 - 0.005 \| \cvec{a} \|_2^2 + \begin{cases}
50 + 25 (\cvec{n}_s \cdot \cvec{v}_b)^{5},\ \text{if ball catched}\\
0,\ \text{else}
\end{cases},
\end{align*}
only giving a meaningful reward when catching the ball and otherwise just a slight penalty on the actions to avoid unnecessary movements. In the above definition, $\cvec{n}_s$ is a normal vector of the end effector surface and $\cvec{v}_b$ is the linear velocity of the ball. This additional term is used to encourage the robot to align its end effector with the curve of the ball. If the end effector is e.g. a net (as assumed for our experiment), the normal is  chosen such that aligning it with the ball maximizes the opening through which the ball can enter the net.
\\
The context $c = [\phi, r, d_x] \in [0.125 \pi, 0.5 \pi] \times [0.6, 1.1] \times [0.75, 4] \subset \mathbb{R}^3$ controls the target ball position in the catching plane, i.e.
\begin{align*}
\cvec{p}_{\text{des}} = [0\quad {-r \cos(\phi)}\quad {0.75 + r \sin(\phi)}].
\end{align*}
Furthermore, the context determines the distance in $x$-dimension from which the ball is thrown
\begin{align*}
\cvec{p}_{\text{init}} = [d_x\ d_y\ d_z],
\end{align*}
where $d_y \sim \mathcal{U}(-0.75, -0.65)$ and $d_z \sim \mathcal{U}(0.8, 1.8)$ and $\mathcal{U}$ represents the uniform distribution. The initial velocity is then computed using simple projectile motion formulas by requiring the ball to reach $\cvec{p}_{\text{des}}$ at time $t = 0.5 + 0.05 d_x$. As we can see, the context implicitly controls the initial state of the environment.
\\
The policy and value function networks for the RL algorithms have three hidden layers with $64$ neurons each and $\tanh$ activation functions. We use a discount factor of $\gamma = 0.995$. The policy updates in TRPO and PPO are done after $5000$ environment steps.
\\
For SAC, a replay buffer size of $100,000$ is used. Due to the sparsity of the reward, we increase the batch size to $512$. Learning with SAC starts after $1000$ environment steps. All other parameters are left to their implementation defaults.
\\
For TRPO we set the GAE parameter $\lambda = 0.95$, leaving all other parameters to their implementation defaults.
\\
For PPO we use a GAE parameter $\lambda = 0.95$, $10$ optimization epochs, $25$ mini-batches per epoch, an entropy coefficient of $0$ and disable the clipping of the value function objective. The remaining parameters are left to their implementation defaults.
\\
Figure \ref{fig:ball_catching} visualizes the catching success rates of the learned policies. As can be seen, the performance of the policies learned with the different RL algorithms achieve comparable catching performance. Interestingly, SAC performs comparable in terms of catching performance, although the average reward of the final policies learned with SAC is lower. This is to be credited to excessive movement and/or bad alignment of the end effector with the velocity vector of the ball.

\bibliography{lit}

\end{document}